\documentclass[10pt,twocolumn,letterpaper]{article}

\usepackage[pagenumbers]{cvpr} %

\usepackage{graphicx}
\usepackage{amsmath}
\usepackage{amssymb}
\usepackage{booktabs}

\newcommand{\degree}[1]{${#1}^o$}
\def\360{\degree{360}}
\usepackage{xfrac}

\usepackage{hyphenat}

\usepackage{booktabs}
\usepackage[dvipsnames, svgnames]{xcolor}
\usepackage{nicefrac}
\usepackage{colortbl}
\usepackage{pifont}
\newcommand{\cmark}{\ding{51}}%
\newcommand{\xmark}{\ding{55}}%
\usepackage{multirow}

\definecolor{Yellow}{rgb}{1.0, 1.0, 0.6}
\definecolor{Orange}{rgb}{1.0, 0.8, 0.6}
\definecolor{Red}{rgb}{1.0, 0.6, 0.6}
\definecolor{brightpink}{rgb}{1.0, 0.0, 0.5}
\definecolor{jade}{rgb}{0.0, 0.66, 0.42}
\definecolor{pastelbrown}{rgb}{0.51, 0.41, 0.33}

\newcommand{\first}[1]{\textbf{#1} \cellcolor{Red}}
\newcommand{\second}[1]{#1 \cellcolor{Orange}}
\newcommand{\third}[1]{#1 \cellcolor{Yellow}}

\usepackage[pagebackref,breaklinks,colorlinks]{hyperref}
\hypersetup{
    colorlinks = true,
    citecolor = {green},
    linkbordercolor = {green},
}

\usepackage[capitalize]{cleveref}
\crefname{section}{Sec.}{Secs.}
\Crefname{section}{Section}{Sections}
\Crefname{table}{Table}{Tables}
\crefname{table}{Tab.}{Tabs.}

\def\cvprPaperID{00000} %
\def\confName{CVPR}
\def\confYear{1970}

\begin{document}

\title{Monocular Spherical Depth Estimation with Explicitly Connected Weak Layout Cues}

\author{Nikolaos Zioulis\\
Centre for Research and Technology Hellas\\
Universidad Polit\'{e}cnica de Madrid\\
{\tt\small nzioulis@gmail.com}
\and
Federico Alvarez\\
Universidad Polit\'{e}cnica de Madrid\\
{\tt\small fag@gatv.ssr.upm.es}
\and
Dimitrios Zarpalas\qquad Petros Daras\\
Centre for Research and Technology Hellas\\
{\tt\small \{zarpalas,daras\}@iti.gr}
\vspace{10pt}
\\
\vspace{10pt}
\small\url{https://vcl3d.github.io/ExplicitLayoutDepth/}
}
\maketitle

\begin{abstract}
Spherical cameras capture scenes in a holistic manner and have been used for room layout estimation.
Recently, with the availability of appropriate datasets, there has also been progress in depth estimation from a single omnidirectional image.
While these two tasks are complementary, few works have been able to explore them in parallel to advance indoor geometric perception, and those that have done so either relied on synthetic data, or used small scale datasets, as few options are available that include both layout annotations and dense depth maps in real scenes.
This is partly due to the necessity of manual annotations for room layouts.
In this work, we move beyond this limitation and generate a \360 geometric vision (360V) dataset that includes multiple modalities, multi-view stereo data and automatically generated weak layout cues.
We also explore an explicit coupling between the two tasks to integrate them into a single-shot trained model.
We rely on depth-based layout reconstruction and layout-based depth attention, demonstrating increased performance across both tasks.
By using single \360 cameras to scan rooms, the opportunity for facile and quick building-scale 3D scanning arises.
\end{abstract}

\section{Introduction}
\label{sec:intro}
Geometry perception is a fundamental computer vision task, and a core technology for applications like Augmented Reality (AR), robotic navigation and 3D reconstruction.
It can be achieved using direct sensing (\textit{i.e.}~time-of-flight or LiDaR technology) or vision-based techniques (\textit{i.e.} multi-ocular stereo).
Recently, the increased performance of data-driven methods has enabled monocular geometry perception.
The applicability of monocular approaches is far superior to approaches that require specific sensors or multiple cameras, and despite the recent progress that modern machine (deep) learning has brought, depth estimation from monocular input remains a challenging problem \cite{bhoi2019monocular}.
This stems from the inherent ill-posedness of the task, the complexity of image formation, as well as the lack of large, high quality datasets.

Specifically for indoor scenes, a large body of work has focused on simple representations instead of dense pixel-wise geometry estimates \cite{https://doi.org/10.1111/cgf.14021}.
A coarse planar 3D reconstruction can be achieved by inferring the scene's structural layout which comprises walls, ceiling and floor, and is estimated by localising the T-junctions where the two walls and a horizontal plane intersect.
Still, this representation is quasi-counterfactual, as it ignores the scene's inner geometry and structure (\textit{i.e.}~the objects).
An important shortcoming though, is that since the corner (T-junction) localisation results are estimated on the projected images, the results are up-to-scale.
On the other hand, finer-grained depth estimation seeks to provide metric-scale measurements, showcasing the complementarity of these two tasks.

In this work, we focus on exploiting this complementarity in larger-scale than prior works, focusing on indoor scene depth estimation with a single omnidirectional\footnote{While the terms omnidirectional, spherical and \360 are interchangeable, we will be using \360 for the remainder of this document.} camera.
Recent advances in sensor miniaturization and consumer hardware open up the opportunity for facile \360 image captures.
Compared to traditional cameras, their \360 nature provides them a couple of advantages.
They require a sparser set of captures to cover a scene/room/building due to their holistic Field-of-View (FoV).
This is also very important for the downstream tasks as shown in \cite{zhang2014panocontext}.
Indeed, tasks like layout estimation are more suited to \360 images.
When using perspective images, the models need to extrapolate beyond their limited FoV to reason about the global scene structure, compared to \360 models that receive the entire scene as input.

A key problem that needs to be addressed in order to facilitate progress towards transforming low-cost \360 cameras to depth sensors is the availability of data.
For \360 depth, a number of datasets have been introduced recently, some synthetic, like Structured3D \cite{Structured3D} and Kujiale \cite{jin2020geometric}, others from real-world scans, like Stanford2D3D \cite{armeni2017joint} and Matterport3D \cite{chang2018matterport3d}, and others generated via synthesis from both types of 3D datasets.
For \360 layout, the aforementioned synthetic datasets also provide layout annotations, but for their real counterparts, there exists a significant size discrepancy between the layout and depth annotated samples.
For synthetic datasets it is straightforward to provide multiple annotated modalities, therefore offering joint layout and depth ground-truth data, but the same does not apply to real datasets.
Up to now, only small subsets of \360 datasets have been annotated with scene layouts. PanoContext \cite{zhang2014panocontext} annotated samples from the Sun360 \cite{xiao2012recognizing} dataset, LayoutNet \cite{zou2018layoutnet} used the Stanford2D3D dataset, and a panorama-based layout estimation study \cite{zou20193d} which annotated a sample of the Matterport3D dataset, offering the LayoutMP3D dataset \cite{wang2020layoutmp3d}.
In addition, the Realtor360 dataset \cite{yang2019dula} was eventually not made public due to licensing issues.
This is reasonable given the effort required to manually annotate numerous samples in high quality.

Our approach seeks to exploit the complementarity of layout and depth estimation, a direction that only a small body of work \cite{jin2020geometric,zeng2020joint} has explored up to now for panoramic inputs.
However, the unavailability of datasets with simultaneous depth and layout ground-truth has limited them to smaller scale data pools.
Here, we take a diverging direction and rely on weak layout labels by generating the corresponding dataset and design a dual task model that can exploit this weak layout information to improve depth estimation performance from a single panorama.
Our dataset and models can be found in our project page \href{https://vcl3d.github.io/ExplicitLayoutDepth/}{vcl3d.github.io/ExplicitLayoutDepth/}.

In summary, our contributions are the following:
\begin{itemize}
    \item We build on prior work and deliver a new benchmark for indoor \360 scene understanding.
    Our 360V dataset contains color, depth, surface orientation, structural semantics and weak layout cues in \360 multi-ocular stereo.
    \item We overcome some of the issues associated to automatic layout labelling via semantic segmentation masks, and improve the quality of the inferior bottom labels using the scene's geometry.
    \item We design our dual task model in a principled manner, integrating best practices for layout and depth estimation, and properly adapting for single-shot training using the weak layout cue annotations.
    \item We integrate explicit constraints between the layout and depth estimation tasks, using the metric scale depth measurements to reconstruct the floor part of the layout, and using the higher quality top layout to attend to the depth estimation task. Apart from the increased performance the layout cues offer, this coupling allows the model to reach a higher performance consensus in both tasks and outperforms models that only implicitly couple the two tasks.
\end{itemize}

The remainder of this document is structured as follows.
In Section~\ref{sec:related}, we initially review the state of the art for \360 datasets in Section~\ref{sec:datasets}, followed by the recent developments in depth and layout estimation from monocular panoramas in Sections~\ref{sec:depth} and \ref{sec:layout}, respectively.
This section concludes with a brief outlook of works focusing on depth estimation combined with layout estimation, either jointly or as a supporting task, to set the grounds for positioning our work\.
In Section~\ref{sec:dataset} we describe the process for generating our dataset, following with our model's design in Section~\ref{sec:method}.
Our results are presented and analysed in Section~\ref{sec:results}, and finally we conclude with a short discussion in Section~\ref{sec:discussion} about the potential of the 360V dataset and geometry estimating \360 cameras for indoor 3D modelling.

\section{Related Work}
\label{sec:related}
\subsection{Spherical Geometric Datasets}
\label{sec:datasets}
Datasets and thus, benchmarks, are the drivers of progress in our field, as even before the advent of modern data-driven methods, they facilitated the assessment of different techniques that incrementally advance the development of new technologies.
Compared to the availability of data for traditional cameras, \360 datasets are lacking mainly due to the relatively recent advances made in \360 imaging.

On the real end of the spectrum, one of the first largest scale \360 datasets was SUN360 \cite{xiao2012recognizing}, offering $67,583$ color panoramas spanning $80$ categories.
While it is no longer available, a small subset of indoor scenes ($\sim 500$) was annotated by PanoContext \cite{zhang2014panocontext} with layout corners.
Similarly, LayoutNet \cite{zou2018layoutnet} annotated $571$ color panoramas with layouts originating from the Stanford2D3D dataset \cite{armeni2017joint}.
The latter is a building-scale 3D dataset that offers structural semantic annotations, in the form of $1413$ color, depth and semantically annotated panoramas from $6$ different large-scale indoor environments.
The panoramas are generated after stitching the Matterport\footnote{\href{https://matterport.com/}{https://matterport.com/}} camera's perspective views, or, for the depth and semantics, after rendering the 3D model from pre-defined viewpoints and then stitching them.
Yet this approach comes with artifacts, the camera's perspective views leave the zenith and nadir empty, which remains empty (\textit{i.e.} filled with black).
For the renderings, in order to provide sufficient quality when stitching them into equirectangular images, high resolutions are required as bilinear interpolation cannot apply to depth or semantics without adding noise to the results.
However, when used during training, the images are downsampled, resulting in aliasing.

Similarly, the Matterport3D dataset \cite{chang2018matterport3d} was scanned with Matterport cameras and provides $90$ 3D buildings annotated with semantics, as well as the original camera's perspective views.
Generating the panoramas requires stitching them, as with Stanford2D3D, resulting in similar artifacts with blurry inpainted zeniths and nadirs.
To this date though, no \360 semantics are available.
Still, Matterport3D offers more fine-grained labels compared to Stanford2D3D, which are not always compatible.
A subset was annotated by a recent layout estimation survey \cite{zou20193d}, which resulted in LayoutMP3D, a panorama dataset with $2295$ panoramas annotated with depth and layouts.

On the other end of the spectrum, there exist synthetic datasets.
Recently, high quality, professional made 3D indoor scenes were ray-traced into \360 panoramas, resulting in the Kujiale \cite{jin2020geometric} and Structured3D \cite{Structured3D} datasets.
These offer multi-modal data (color, depth, normals), in a variety of lighting (raw, warm, cold) and furniture (empty, simple, full) settings.
Due to their synthetic nature, they are easily supplemented with layout annotations and even albedo maps, totalling $3550$ and $21835$ unique samples for each one respectively.

In the middle, there exist hybrid approaches that generate data by re-using 3D datasets.
OmniDepth \cite{zioulis2018omnidepth} leveraged the two aforementioned real-world scanned 3D datasets, Matterport3D and Stanford2D3D, as well as two synthetic 3D datasets, SunCG \cite{song2017semantic} and SceneNet \cite{handa2015scenenet}, to synthesize \360 color and depth pairs via ray-traced rendering, totalling $23524$ unique samples.
It was later extended in \cite{zioulis2019spherical,karakottas2019360} with vertical and horizontal stereo, as well as normal maps, offering $8680$ real and $9311$ synthetic samples\footnote{The synthetic samples are from the discontinued SunCG dataset.}.
Our approach is an extension of these works that offers additional stereo viewpoints, fixes lighting issues, and additionally generates consistent structure-based semantics and weak layout cues.

\subsection{Spherical Depth Estimation}
\label{sec:depth}
Geometry estimation from \360 images with traditional (\textit{i.e.}~non data-driven) methods was achieved via stereo \cite{kim20133d} or structure-from-motion \cite{huang20176}.
Following the first data-driven monocular depth estimation work \cite{eigen2014depth}, OmniDepth \cite{zioulis2018omnidepth} was the first data-driven method for monocular \360 depth estimation, trained with a generated \360 color and depth dataset to overcome the distinct lack of data. It employed supervised regression, showing that training directly using equirectangular images is beneficial.
In parallel, distortion\hyp{}aware filters \cite{tateno2018distortion} were used to apply perspective trained models to equirectangular images with reduced performance deviation.

Following the advances in self-supervised depth estimation \cite{zhou2017unsupervised}, the concept was applied to \360 videos in \cite{wang2018self}, where a small video dataset was generated using SunCG.
A cube-map representation was used with a pose consistency loss applied to restrict the poses estimates when applying the perspective self-supervised mode to each face.
Similarly, stereo-based principles have also been applied to \360 depth estimation in \cite{zioulis2019spherical}, with a trinocular dataset rendered to demonstrate the feasibility of horizontal stereo for \360 inputs, apart from the more frequently used vertical stereo \cite{wang2020360sd}.
Recently, Bi-Fuse \cite{wang2020bifuse} exploited both representations -- equirectangular and cube-map -- showcasing the increased performance arising from their fusion.
Our work focuses on monocular depth estimation from a single panorama, in a dual task setting, integrating layout cues in a way beneficial to both tasks, and using an adapted coarse-to-fine architecture.

\subsection{Spherical Layout Estimation}
\label{sec:layout}
Lately, layout estimation from \360 panoramas has received considerable attention.
The reader is referred to a survey about 3D reconstruction of structured indoor environments \cite{https://doi.org/10.1111/cgf.14021} for an extended review.
Some of these focus on using multiple panoramas to reconstruct planar approximations of interior spaces
\cite{pintore2016omnidirectional, pintore2018recovering, pintore20183d, pintore2019automatic}.
This further highlights the interplay between the coarse layout estimation task, that can be used to align captures between themselves, and the finer grained depth estimation task, which can additionally offer more structural details for these reconstructions.
Still, when seeking to apply data-driven methods for layout estimation, manual annotations are required, hindering progress.

With the pioneering work of PanoContext \cite{zhang2014panocontext}, the advantages of using \360 inputs were demonstrated for indoor geometry inference tasks.
Different variants of this traditional optimization approach focused on reformulating its optimization \cite{fukano2016room,yang2016efficient}.
Follow up works exploited the increased performance offered by early data-driven methods to replace traditional components of these approaches \cite{fernandez2018layouts,xu2017pano2cad,yang2018automatic}.
Naturally, end-to-end models also emerged, like PanoRoom \cite{fernandez2018panoroom} and LayoutNet \cite{zou2018layoutnet} that densely approximate the layout corners as spatial probabilities, and then extract their location via maximal activation detection.
This way, a post-processing step is required to ensure the Manhattan alignment of the estimations.

Lately, more elaborate deep models like DuLa-Net \cite{yang2019dula} and HorizonNet \cite{sun2019horizonnet} presented superior results by exploiting the nature of \360 images when projected in various ways, or via new parameterizations of the estimated layout, respectively.
Nevertheless, all these works, used the SUN360 and Stanford2D3D subsets, totalling about $\sim 1000$ annotated samples.
A single exception is DuLa-Net that introduced the -- now unavailable -- Realtor360 dataset.

Out of these samples only the Stanford2D3D ones also had finer-grained geometry (\textit{i.e.}~depth) ground-truth data.
The LayoutMP3D dataset \cite{zou20193d} offers a higher percentage of non-cuboid rooms compared to previous datasets that are limited in this aspect.
All these approaches require high quality annotations, which translates to larger-scale but synthetic datasets, or smaller scale, manually annotated real-world acquired datasets.
In this work, we automatically generate weak layout cues, offering a dataset a magnitude larger than prior work that also offers multi-modal geometric annotations.

\subsection{Joint Layout \& Depth}
\label{sec:joint}
A drawback of current layout estimation approaches is that their estimates are up-to-scale.
Typically, a single measurement (\textit{i.e.}camera-to-floor) is required to lift them to metric\hyp{}scale 3D when relying on the Manhattan assumption.
An exception is DuLa-Net which also regresses the room's height to ensure metric-scale measurements.
Nonetheless, the interplay between layout and depth estimation is apparent and has recently been considered in \cite{jin2020geometric} and \cite{zeng2020joint}.
In the former work, the correlation between these two tasks is considered, and layout estimation is indirectly used as an attention mechanism to separate the foreground from the background, and as a cycle consistency, considering that it should be possible to infer the room's layout from the predicted depth image.
In the latter work, a virtual layout only depth map is integrated in a coarse-to-fine learning framework, to simultaneously predict the layout and depth of a scene.

For both of these works though, layout information is implicitly handled within the network, and in complex training regimes involving multiple sub-models that are trained progressively.
Another issue is the availability of datasets, \cite{jin2020geometric} uses the synthetic Kujiale dataset, also offering results on the layout subset of Stanford2D3D, while \cite{zeng2020joint} only presents results for depth and layout estimation in the latter.
This is because no other datasets with co-present layout and depth annotations were available.
Even more so, for real-world acquired data, only a subset of Matterport3D and Stanford2D3D are currently available.
In this work, we generate a larger dataset that offers depth and layout annotations from Matterport3D and Stanford2D3D.
To do so, we move beyond manual annotations and instead automatically annotate weak layout cues, and use them to improve a depth estimation model by integrating them explicitly into the model, which is designed to overcome the disadvantages of the weakly annotated data.

\section{360V Indoors Dataset}
\label{sec:dataset}
\begin{figure*}[!htbp]
    \begin{tabular}{ccccc}
    \includegraphics[width=0.19\linewidth]{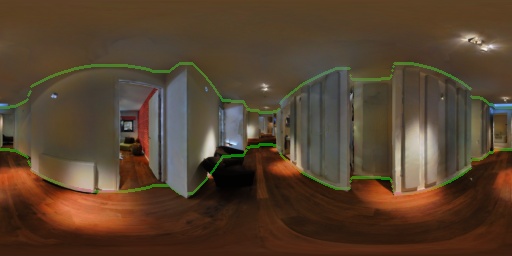} &
    \includegraphics[width=0.19\linewidth]{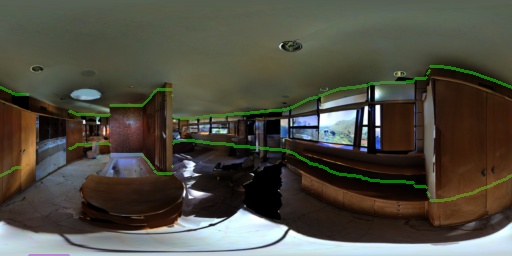}
    &
    \includegraphics[width=0.19\linewidth]{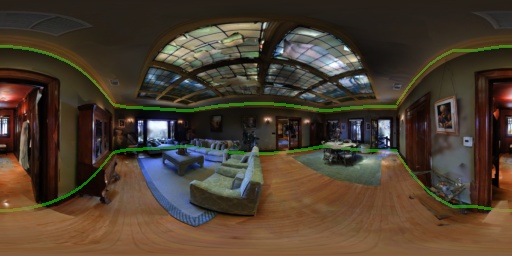}
    &
    \includegraphics[width=0.19\linewidth]{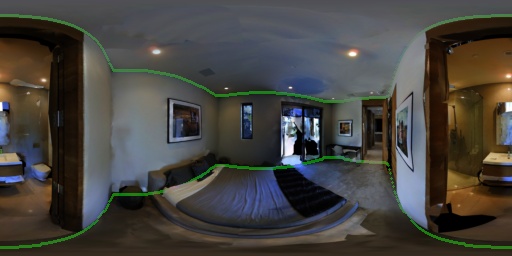}
    &
    \includegraphics[width=0.19\linewidth]{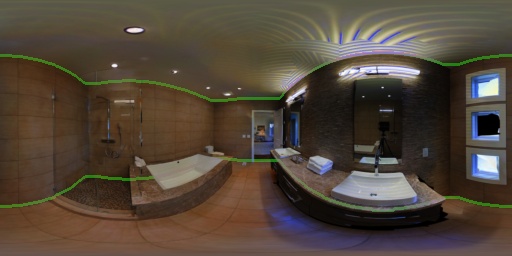}\\
    
    \includegraphics[width=0.19\linewidth]{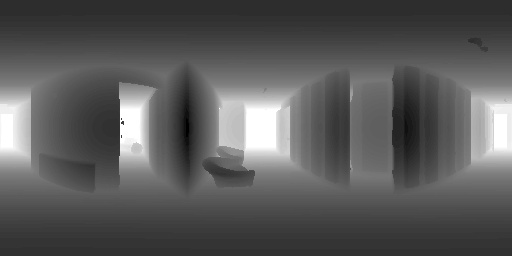}
    &
    \includegraphics[width=0.19\linewidth]{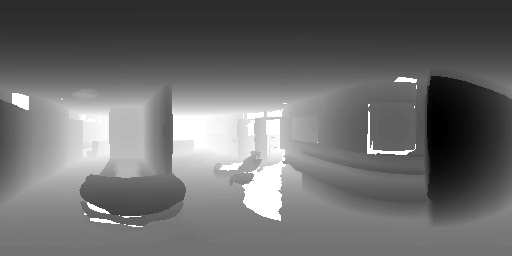}
    &
    \includegraphics[width=0.19\linewidth]{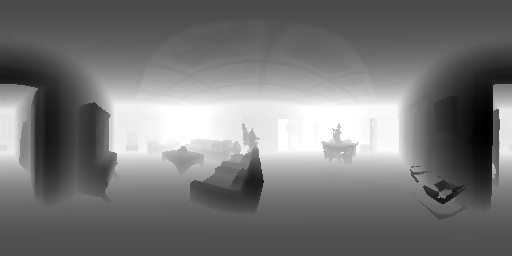}
    &
    \includegraphics[width=0.19\linewidth]{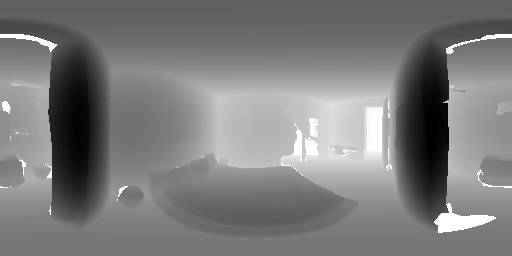}
    &
    \includegraphics[width=0.19\linewidth]{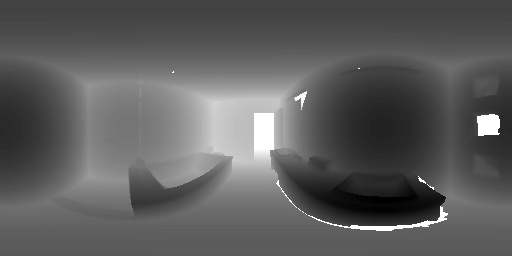}\\

    \includegraphics[width=0.19\linewidth]{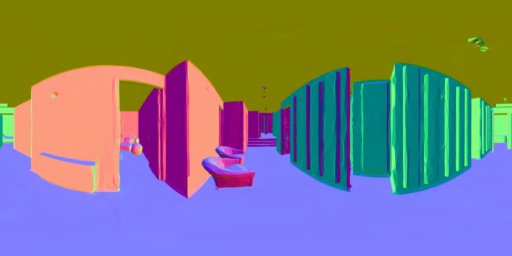}
    &
    \includegraphics[width=0.19\linewidth]{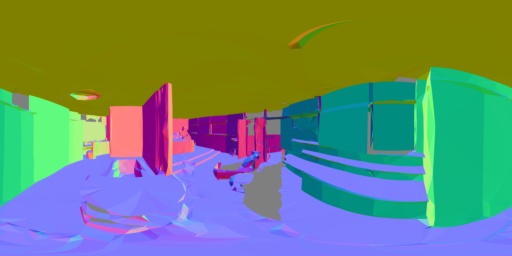}
    &
    \includegraphics[width=0.19\linewidth]{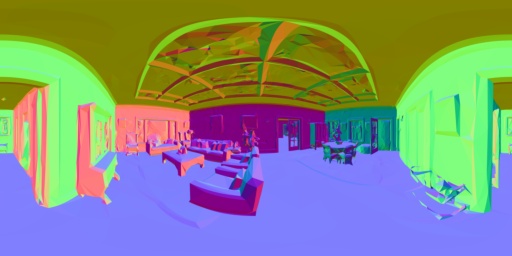}
    &
    \includegraphics[width=0.19\linewidth]{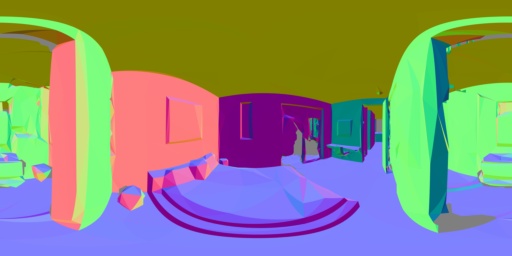}
    &
    \includegraphics[width=0.19\linewidth]{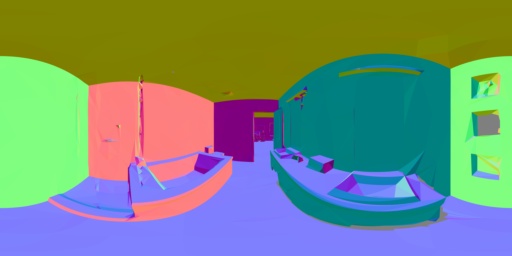}\\
    
    \includegraphics[width=0.19\linewidth]{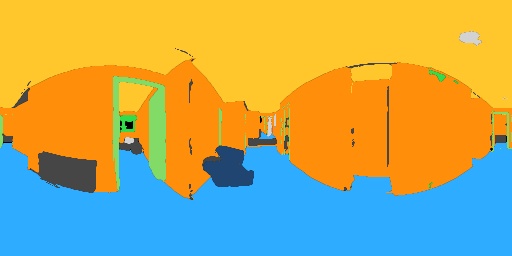}
    &
    \includegraphics[width=0.19\linewidth]{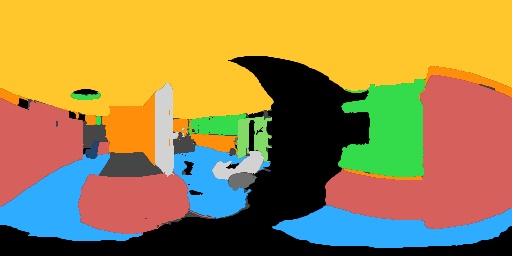}
    &
    \includegraphics[width=0.19\linewidth]{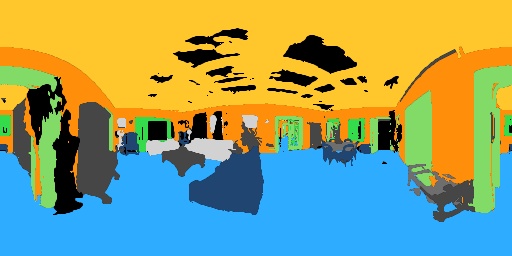}
    &
    \includegraphics[width=0.19\linewidth]{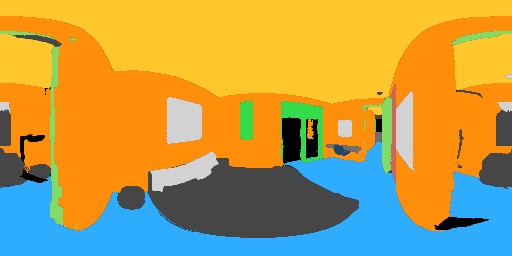}
    &
    \includegraphics[width=0.19\linewidth]{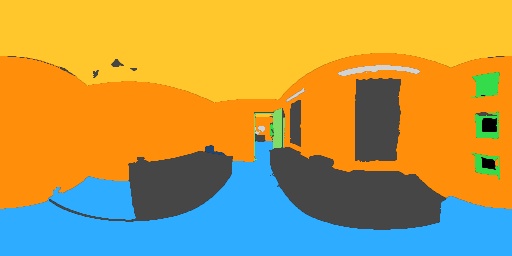}\\

    \includegraphics[width=0.19\linewidth]{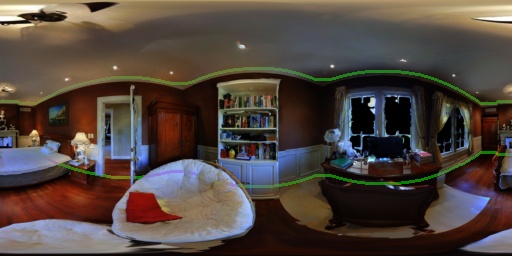}
    &
    \includegraphics[width=0.19\linewidth]{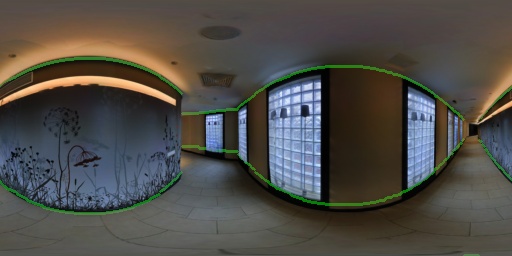}
    &
    \includegraphics[width=0.19\linewidth]{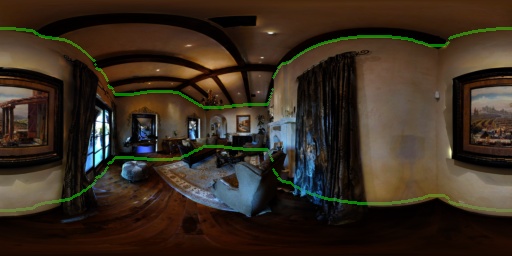}
    &
    \includegraphics[width=0.19\linewidth]{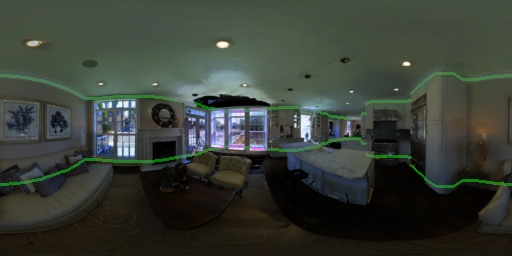}
    &
    \includegraphics[width=0.19\linewidth]{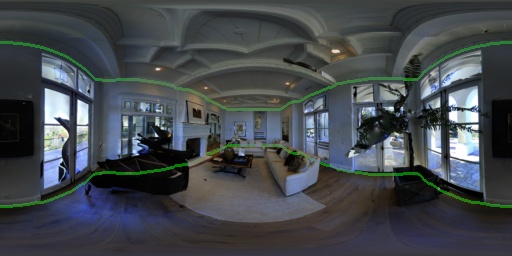}\\

    \includegraphics[width=0.19\linewidth]{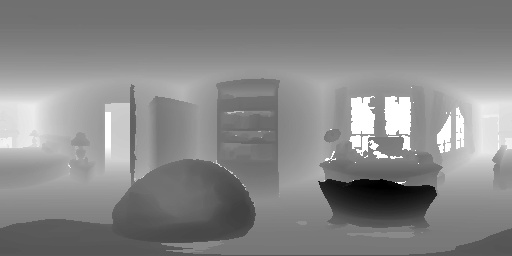}
    &
    \includegraphics[width=0.19\linewidth]{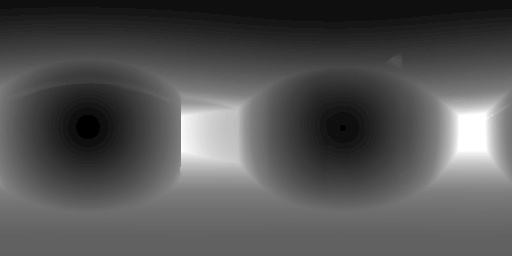}
    &
    \includegraphics[width=0.19\linewidth]{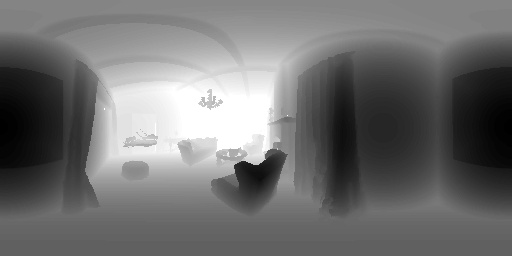}
    &
    \includegraphics[width=0.19\linewidth]{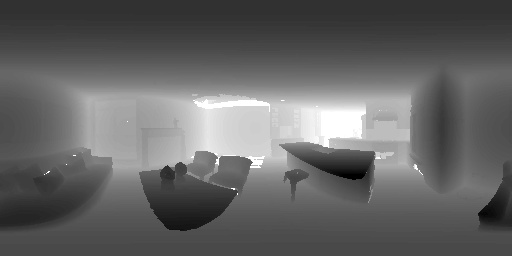}
    &
    \includegraphics[width=0.19\linewidth]{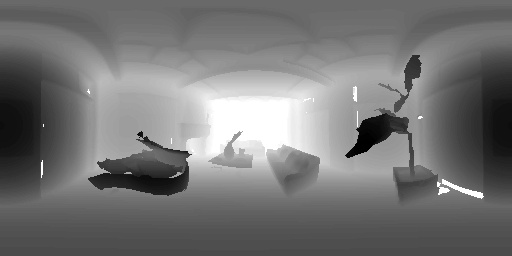}\\
    
    \includegraphics[width=0.19\linewidth]{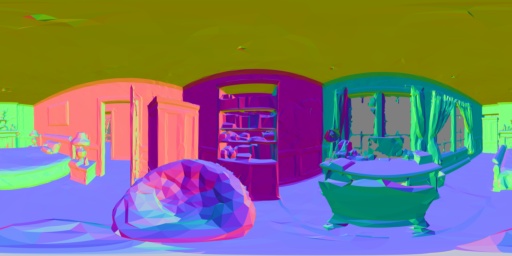}
    &
    \includegraphics[width=0.19\linewidth]{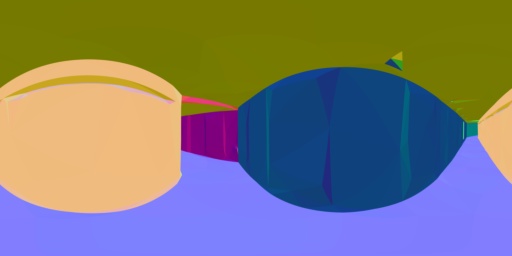}
    &
    \includegraphics[width=0.19\linewidth]{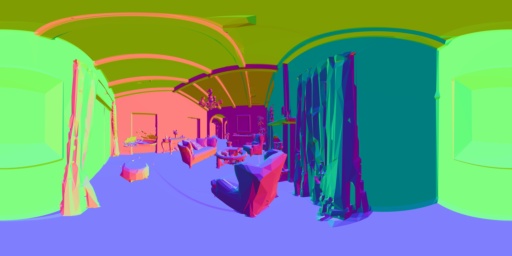}
    &
    \includegraphics[width=0.19\linewidth]{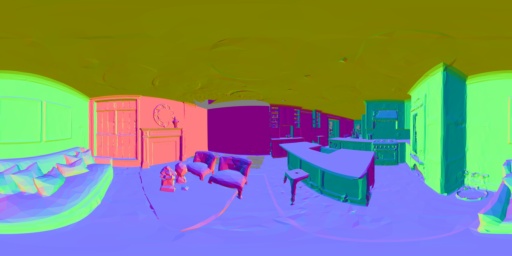}
    &
    \includegraphics[width=0.19\linewidth]{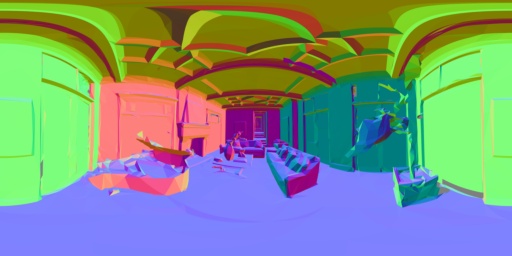}\\

    \includegraphics[width=0.19\linewidth]{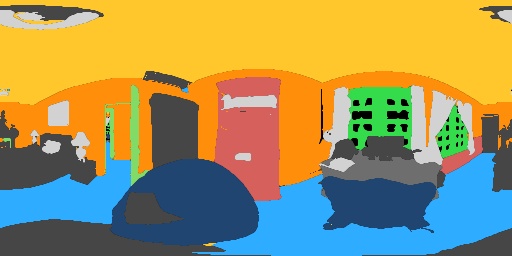}
    &
    \includegraphics[width=0.19\linewidth]{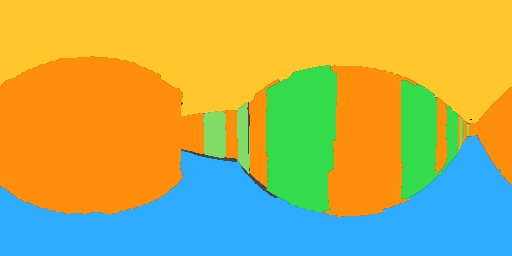}
    &
    \includegraphics[width=0.19\linewidth]{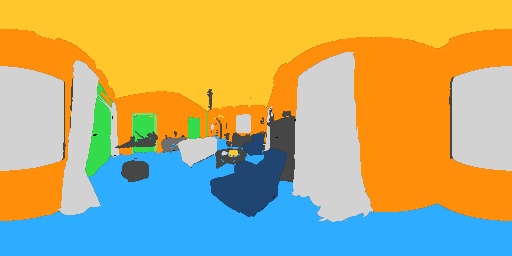}
    &
    \includegraphics[width=0.19\linewidth]{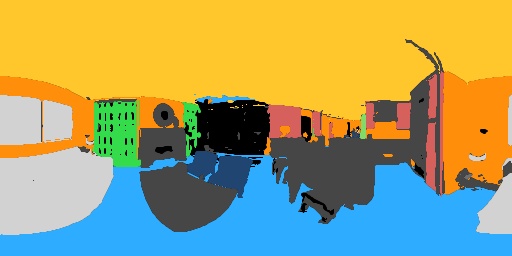}
    &
    \includegraphics[width=0.19\linewidth]{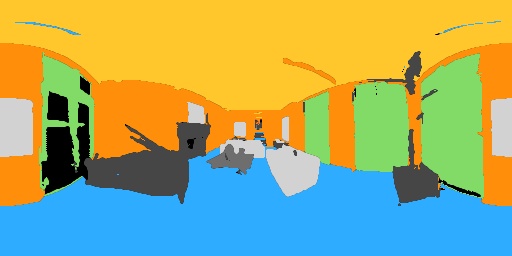}\\
    
    \multicolumn{5}{c}{\includegraphics[width=0.96\linewidth]{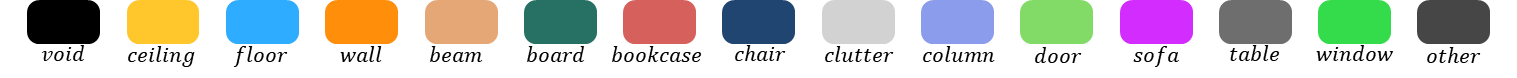}}
    
    \end{tabular}
    \caption{
        Random samples from the 360V dataset. 
        For each sample, the automatically annotated weak layout cues are depicted on the color image in rows (a) and (e), which is accompanied by a high quality depth map in rows (b) and (f) with, a surface orientation map in rows (c) and (g) with, and a semantic label in rows (d) and (h) with map that contains the main scene structural elements as illustrated in row (i) depicting the colored labels legend, at the bottom of the figure.
    }
\label{fig:dataset}
\end{figure*}

In this section we describe our introduced dataset for \360 vision in indoor scenes, the \textit{360V Indoors Dataset}.
We build on prior work, specifically the generated dataset of \cite{zioulis2018omnidepth,zioulis2019spherical} but improve and expand it with additional samples and modalities.
In particular, we improve the color samples by correcting the shortcoming of the previous versions \cite{zioulis2018omnidepth,zioulis2019spherical}, which was the addition of extra lighting in the rendered scenes.
Additionally, we complement it with an additional stereo viewpoint as well as semantic and weak layout annotations.
The following sections describe our methodology in detail, while Figure~\ref{fig:dataset} depicts a set of random samples and the provided annotations.

\subsection{Preliminaries \& Notation}
\label{sec:preliminaries}
In this section we briefly introduce the notation and conventions that are used throughout the paper.
Eq.(\ref{eq:spherical_cartesian_coords}) presents the spherical coordinate system that we use, with $\phi \in [0, 2\pi]$, $\theta \in [0, \pi]$, and $y$ being the vertical axis.

\small
\begin{gather}
\label{eq:spherical_cartesian_coords}
    \left[\begin{matrix}
        r \\
        \phi \\
        \theta
    \end{matrix}\right]
    = \left[\begin{matrix}
        \sqrt{x^2 + y^2 + z^2} \\
        \tan^{-1}(\nicefrac{x}{z}) \\
        \cos^{-1}(\nicefrac{y}{r})
    \end{matrix}\right], \quad
    \left[\begin{matrix}
        x \\
        y \\
        z
    \end{matrix}\right]
    = \left[\begin{matrix}
        r\sin\phi\sin\theta \\
        r\cos\theta               \\
        r\cos\phi\sin\theta
    \end{matrix}\right].
\end{gather}
\normalsize

Panorama images are projected into an equirectangular grid, where two domains are defined, the pixel domain $\Omega: [0, W] \times [0, H]$, with $W$ and $H$ being the image's width and height, respectively, where each pixel $\mathbf{p} = (u, v) \in \Omega$ is defined with its discrete horizontal and vertical coordinates $u, v$.
The second domain $\mathcal{A}: [0, 2\pi] \times [0, \pi]$, is the angular domain where each angular coordinate $\boldsymbol{\rho} = (\phi, \theta) \in \mathcal{A}$ is defined with its continuous longitude/azimuth and latitude/elevation coordinates $\phi, \theta$, respectively.
Given that the mapping between pixel and angular coordinates is linear, we flexibly transition between the two domains.

We consider a set of scalar or vector valued signals defined on the aforementioned domains: \textbf{i)} color images $\mathbf{C} \in \mathbb{R}^3$ that contain trichromatic color values $c \in [0, 255]$, \textbf{ii)} depth maps $\mathbf{D} \in \mathbb{R}$ where each pixel value corresponds to the distance/radius $r \in \mathbb{R}$ as defined Eq.(\ref{eq:spherical_cartesian_coords}), \textbf{iii)} normal maps $\mathbf{N} \in \mathbb{R}^3$ where each pixel corresponds to a normalized surface orientation $\mathbf{n} = (n_x, n_y, n_z), n_i \in [-1, 1]$, and \textbf{iv)} semantic label maps $\mathbf{L} \in \mathbb{N}$ where each pixel contains a class assignment label $l \in [1, L]$, with $L$ being the total number of semantic classes.
All these panorama signals are functions of pixel or angular coordinates, but this notation has been omitted for brevity.

\subsection{Generation via Ray-tracing 3D scenes}
\label{sec:dataset_raytracing}
\begin{figure*}[!htbp]
    \centering
    \includegraphics[width=\linewidth]{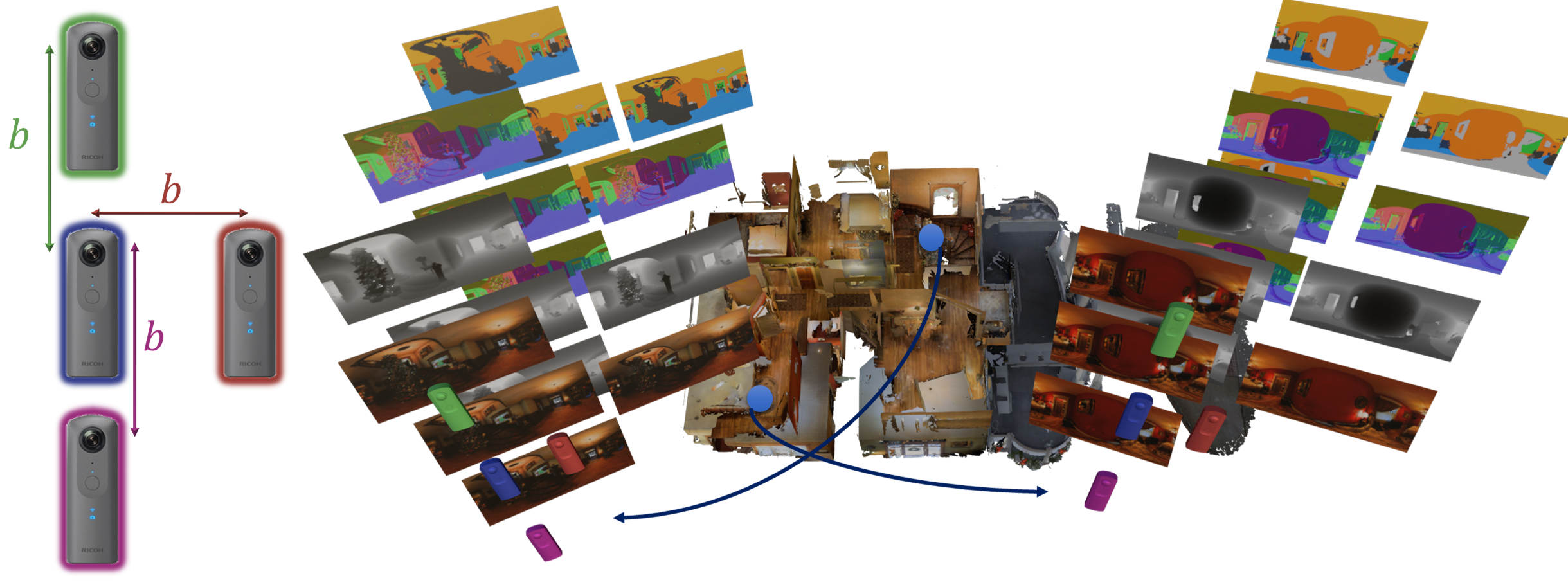}
    
    \caption{
        Our synthesis-based data generation approach.
        We place a virtual \360 camera rig (left) that comprises a trinocular vertical stereo setup (\textcolor{magenta}{down}, \textcolor{blue}{center}, and \textcolor{ForestGreen}{up}), and simultaneously, a binocular horizontal stereo pair (\textcolor{blue}{left-center} and \textcolor{red}{right}), within pre-defined positions (right) inside a 3D scanned building (Section~\ref{sec:dataset_raytracing}).
        A consistent baseline $b$ is used for all camera displacements.
        We render multi-modal data from each viewpoint (right), to generate annotations for each color image, either geometric (depth and surface orientation), or semantic (Section~\ref{sec:dataset_semantic}), and additionally post-process the data to extract weak layout cues (Section~\ref{sec:dataset_layout})
    }
\label{fig:rendering}
\end{figure*}

\begin{figure}[!htbp]
    \centering
        
    \subfloat{\includegraphics[width=0.5\linewidth]{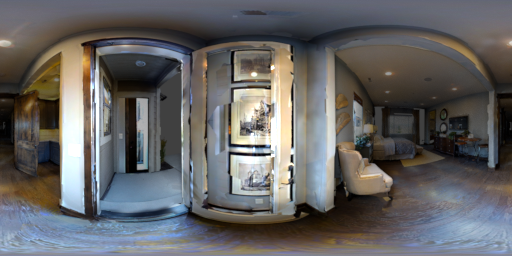}}
    \hfill
    \subfloat{\includegraphics[width=0.5\linewidth]{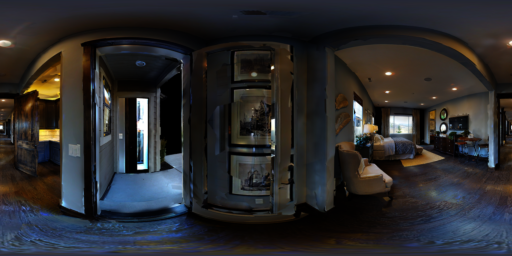}}
    \hfill
    \vspace{-0.3 cm}
    \subfloat{\includegraphics[width=0.5\linewidth]{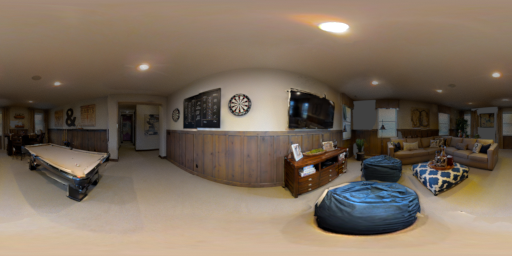}}
    \hfill
    \subfloat{\includegraphics[width=0.5\linewidth]{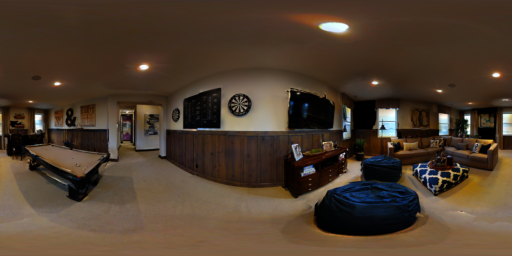}}
    \hfill
    \vspace{-0.3 cm}
    \subfloat{\includegraphics[width=0.5\linewidth]{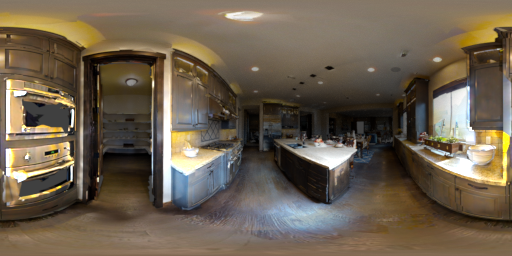}}
    \hfill
    \subfloat{\includegraphics[width=0.5\linewidth]{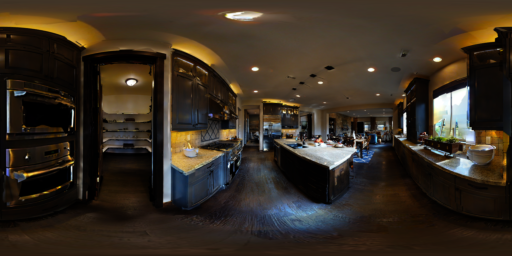}}
    \hfill
    \vspace{-0.3 cm}
    \subfloat{\includegraphics[width=0.5\linewidth]{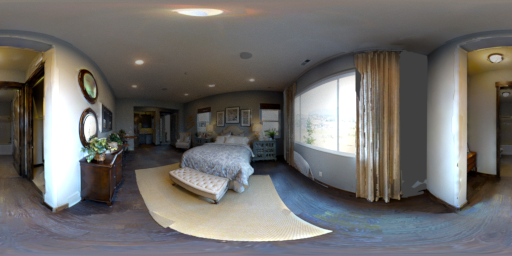}}
    \hfill
    \subfloat{\includegraphics[width=0.5\linewidth]{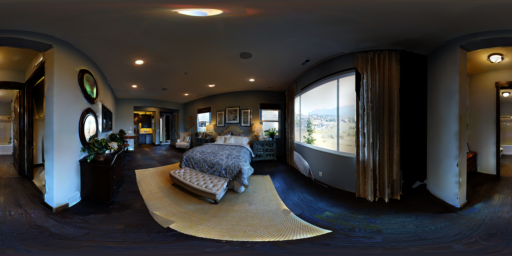}}
    \hfill
    \vspace{-0.3 cm}
    \subfloat{\includegraphics[width=0.5\linewidth]{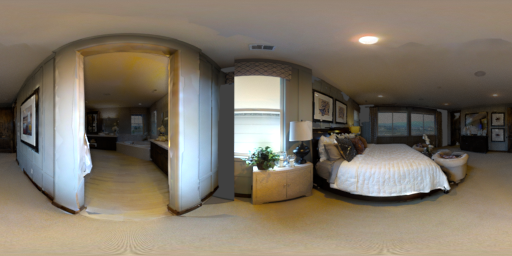}}
    \hfill
    \subfloat{\includegraphics[width=0.5\linewidth]{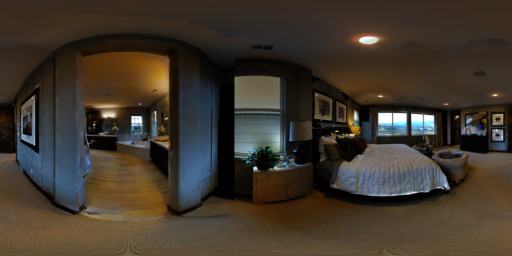}}
    \hfill

    \caption{
        Adding light sources during ray-traced generation from pre-scanned buildings introduces a significant depth bias into the dataset.
        On the left, the samples from OmniDepth \cite{zioulis2018omnidepth} where this bias is clearly depicted as close up surfaces are bright, while farther away surfaces are dark.
        On the right, the corresponding 360V samples where the original texture with pre-baked lighting was sampled and gamma corrected.
        The scenes are realistically lit allowing models to focus on capturing the context instead of relating pixel intensities to depth measurements.
    }
\label{fig:lighting}
\end{figure}

Since the manual acquisition of panoramas is tedious, and the annotation of dense geometric information, largely impossible, we generate our \360 dataset via synthesis.
We re-use the results of large-scale 3D building datasets like Matterport3D and Stanford2D3D to render multi-modal \360 data.
These datasets were acquired using the Matterport camera and software, which stitches a set of captures together to generate a 3D model of interior spaces.
The scan positions of the Matterport camera within these 3D building-scale reconstructions are available, which allows us to position virtual viewpoints in appropriate predetermined physical viewpoints to generate our panoramas via synthesis (\textit{i.e.}~rendering).
We rely on high performance and quality ray-tracing as our synthesis process, using Blender\footnote{\href{https://www.blender.org/}{https://www.blender.org/}} and the Cycles\footnote{\href{https://www.cycles-renderer.org/}{https://www.cycles-renderer.org/}} ray-tracer to directly output \360 panoramas.

The ray-tracing approach can exploit the advantages of computational synthesis procedures to generate multimodal outputs, and to also emplace extra virtual viewpoints in the scene.
Capitalizing on this, we perform one data generation pass for each scan position $\mathbf{t} \in \mathbb{R}^3$, adding extra virtual viewpoints in two stereo configurations.
We place virtual viewpoints at positions $\mathbf{t}^c, \mathbf{t}^d, \mathbf{t}^u, \mathbf{t}^r$, with $c, d, u, r$ being \textit{center}, \textit{down}, \textit{up} and \textit{right} respectively.
This offers a trinocular vertical stereo configuration (\textit{c}-\textit{d}-\textit{u}), which is the most typical one in the literature \cite{li2008binocular,wang2020360sd,kim20133d}, as well as a horizontal stereo configuration (\textit{c}-\textit{r}), which is a more complex case \cite{zioulis2019spherical}.
We keep a consistent baseline of $b = 0.26m$ between all placements with respect to the central virtual viewpoint that is placed at the scan position.
This corresponds to a ray convergence distance of $8m$, and additionally offers a variation of camera-to-floor/ceiling distance, as most datasets are captured at an approximate $1.6m$ distance from the floor using standard tripods \cite{xiao2012recognizing,chang2018matterport3d,armeni2017joint}, or, for the synthetic case, rendered from the center of the room \cite{zioulis2018omnidepth,jin2020geometric}.
A visualization of the virtual camera setup and the data generation process can be found in Figure~\ref{fig:rendering}.

For each rendering pass we use a $512 \times 256$ resolution and $512$ samples per ray and output a color image $\mathbf{C}$, a depth map $\mathbf{D}$, a world aligned normal map $\mathbf{N}$ (our viewpoints are not rotated, so this surface orientation aligns with the local coordinate system as well), and semantic labels $\mathbf{L}$ (to be described in Section~\ref{sec:dataset_semantic}).
This results in a total of $12,213$ unique viewpoints (only counting the central viewpoints) with high quality depth, surface, semantic and weak layout cue annotations (see Section~\ref{sec:dataset_layout}).
While OmniDepth \cite{zioulis2018omnidepth} offered horizontally rotated viewpoints as well, this only adds redundant data as circular shift augmentations can easily create the necessary rotational variety.
Contrary to the OmniDepth renderings that added a light source at the camera position, we use a custom shader and output raw texture sampled values, preserving the photo-realism of the data.
We additionally perform gamma correction on the texture samples as again, random gamma augmentations (and other global color augmentations) can adjust the color space on demand.
This allows for an expanded range of simultaneous color augmentations without reaching saturation levels, as the images are stored in low dynamic range to comply with most cameras' capabilities.
More importantly though, the aforementioned light source removal improves the quality of the dataset as, besides highly saturated images, this also added an important bias to the dataset.

Indeed, light strength attenuation resulted in distant geometry to be lowly lit, while close geometry to be more saturated.
This was an important cue for the data-driven methods to exploit, which unfortunately departed from photo-realism and was improbable in real-world scenarios.
To demonstrate this point, we extract the Pearson Correlation Coefficient (PCC) between the lightness of the color images, after converting them to the Lab color space, and the inverse depth of the depth images.
Table~\ref{tab:lighting} presents the color bias PCC of the OmniDepth dataset and our 360V dataset, which shows a strong positive correlation of lighting and inverse depth for OmniDepth, and a low positive correlation for 360V.
It also demonstrates that data-driven models will exploit this bias to improve performance, but at the cost of generalization, as we also provide results for two \360 depth models trained on our dataset and the biased OmniDepth.
The large performance boost that both methods exhibit, shows how data-driven methods can, and will, exploit this bias. Further, Figure~\ref{fig:lighting} illustrates this important difference qualitatively with a set of corresponding samples from the two datasets.
Realistic lighting is a very important factor for synthetically generated datasets \cite{zhang2020study}, and despite the fact that 360V is not synthetic \footnote{The scanned 3D models that are rendered were ``measured'' using cameras via the 3D reconstruction of real-world scenes.}, it is generated via synthesis, and thus, the necessary attention to this detail ensures the quality of the data and its suitability for data-driven methods.

\begin{table*}[!htbp]
\centering
\caption{Artificial lighting bias for generated renders from pre-scanned buildings.
The OmniDepth dataset exhibits significant correlation between the luminance and the inverse depth, which is detrimental for models trained on that dataset.
Instead, 360V is lowly correlated, offering more photorealistic scenes.
The performance deviation between two depth estimation models trained on each dataset showcases the fact that data-driven models can and will exploit this bias.}
\label{tab:lighting}
\begin{tabular}{@{}lc|cc|cc@{}}
\toprule
\multirow{2}{*}{} &
  \multirow{2}{*}{\begin{tabular}[c]{@{}c@{}}Luminance \&\\ Inv. Depth PCC\end{tabular}} &
  \multicolumn{2}{c|}{RectNet \cite{zioulis2018omnidepth}} &
  \multicolumn{2}{c}{BiFuse \cite{wang2020bifuse}} \\ \cmidrule(l){3-6} 
 &
   &
  \multicolumn{1}{c|}{RMSE $\downarrow$} &
  $\delta_1 \uparrow$ &
  \multicolumn{1}{c|}{RMSE $\downarrow$} &
  $\delta_1 \uparrow$ \\ \midrule
OmniDepth \cite{zioulis2018omnidepth} &
  \textbf{0.6374} &
  \textbf{0.3588} &
  \textbf{91.21\%} &
  \textbf{0.3465} &
  \textbf{93.56\%} \\
360V (Ours) &
  0.1429 &
  0.4408 &
  82.54\% &
  0.4220 &
  87.55\% \\ \bottomrule
\end{tabular}
\end{table*}

\subsection{Material Mattes for Semantic Labelling}
\label{sec:dataset_semantic}
\begin{figure}[!htbp]
    \centering
    
    \subfloat{\includegraphics[width=0.49\linewidth]{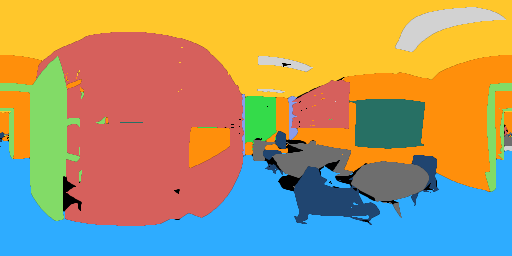}}
    \hfill
    \subfloat{\includegraphics[width=0.49\linewidth]{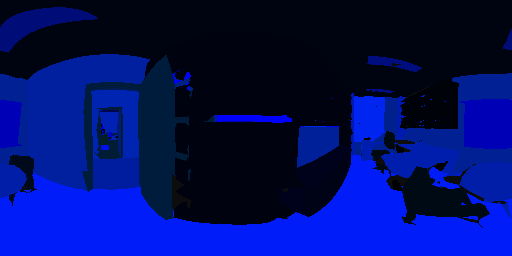}}
    \vspace{-0.3cm}
    \subfloat{\includegraphics[width=0.49\linewidth]{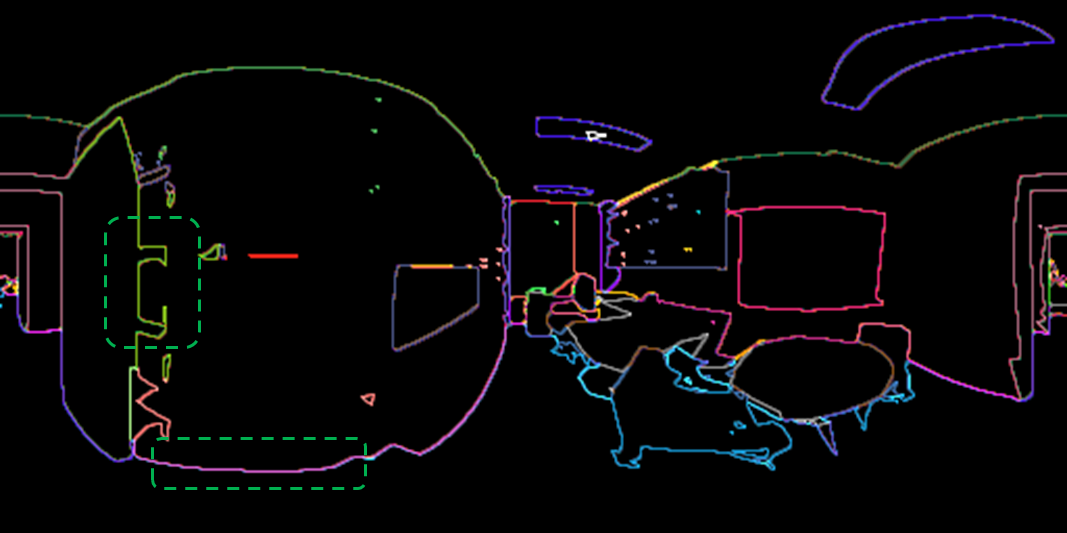}}
    \hfill
    \subfloat{\includegraphics[width=0.49\linewidth]{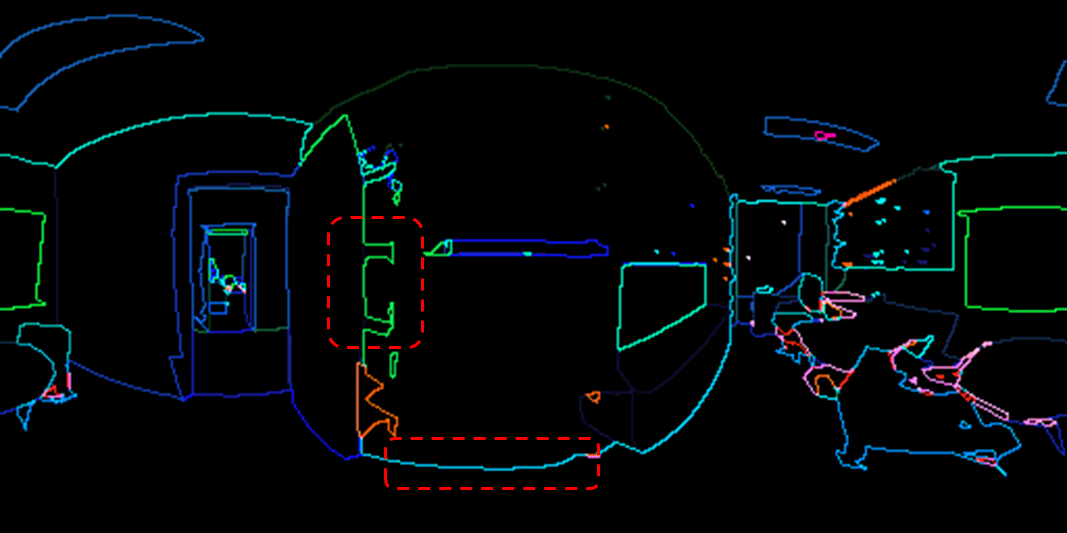}}
    \vspace{-0.3cm}
    \subfloat{\includegraphics[width=0.49\linewidth]{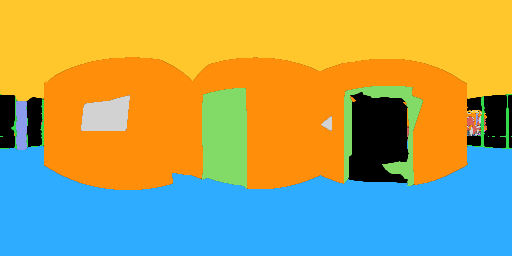}}
    \hfill
    \subfloat{\includegraphics[width=0.49\linewidth]{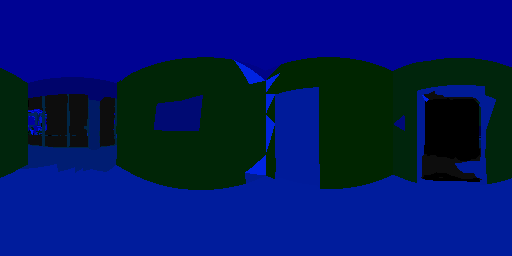}}
    \vspace{-0.3cm}
    \subfloat{\includegraphics[width=0.49\linewidth]{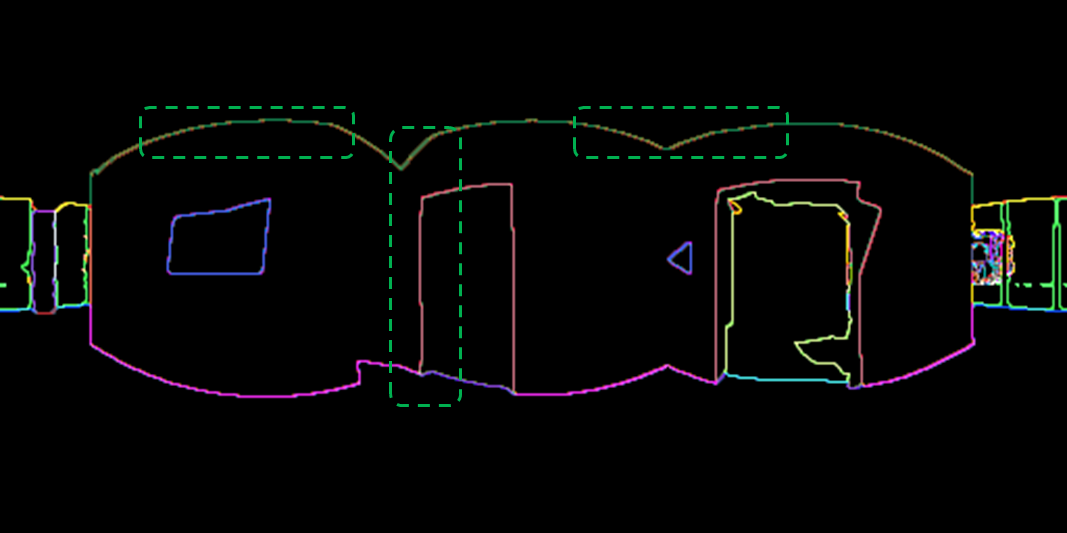}}
    \hfill
    \subfloat{\includegraphics[width=0.49\linewidth]{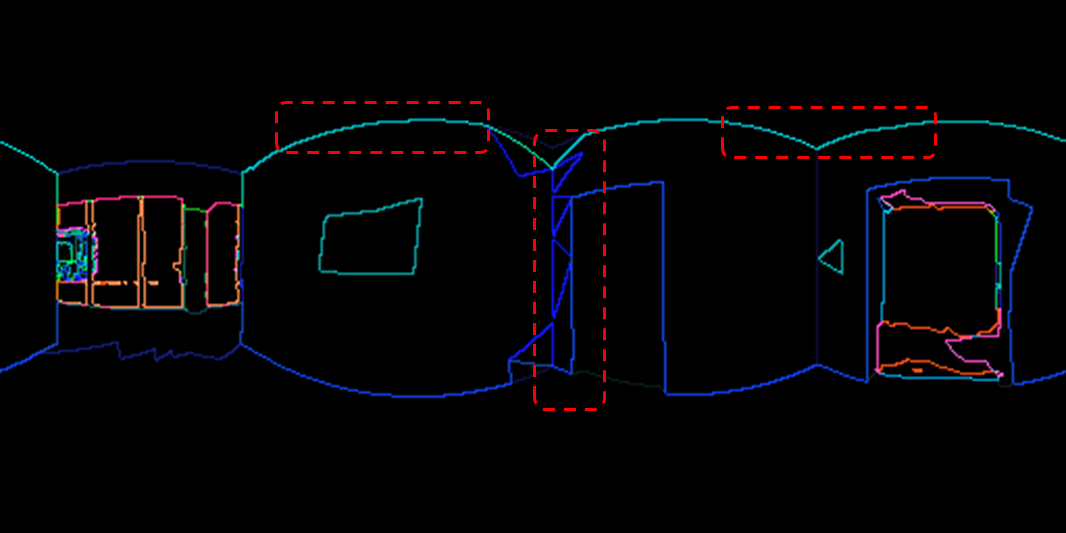}}
    \vspace{-0.1cm}\\
    \rule[0.5ex]{8cm}{0.5pt}\\
    \vspace{-0.3cm}
    \subfloat{\includegraphics[width=0.49\linewidth]{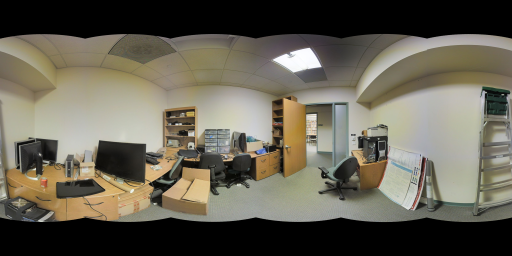}}
    \hfill
    \subfloat{\includegraphics[width=0.49\linewidth]{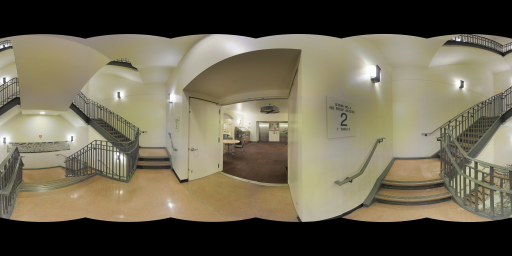}}
    \vspace{-0.3cm}
    \subfloat{\includegraphics[width=0.49\linewidth]{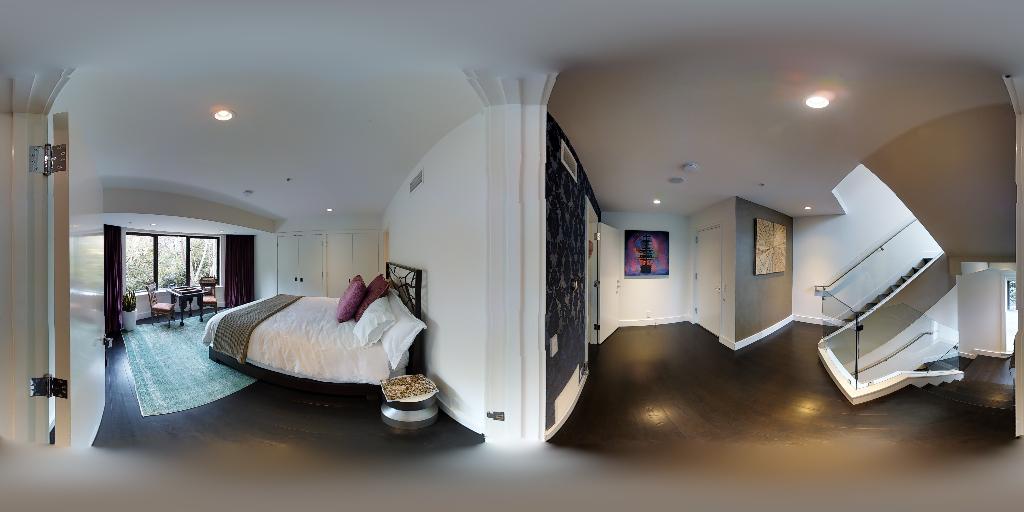}}
    \hfill
    \subfloat{\includegraphics[width=0.49\linewidth]{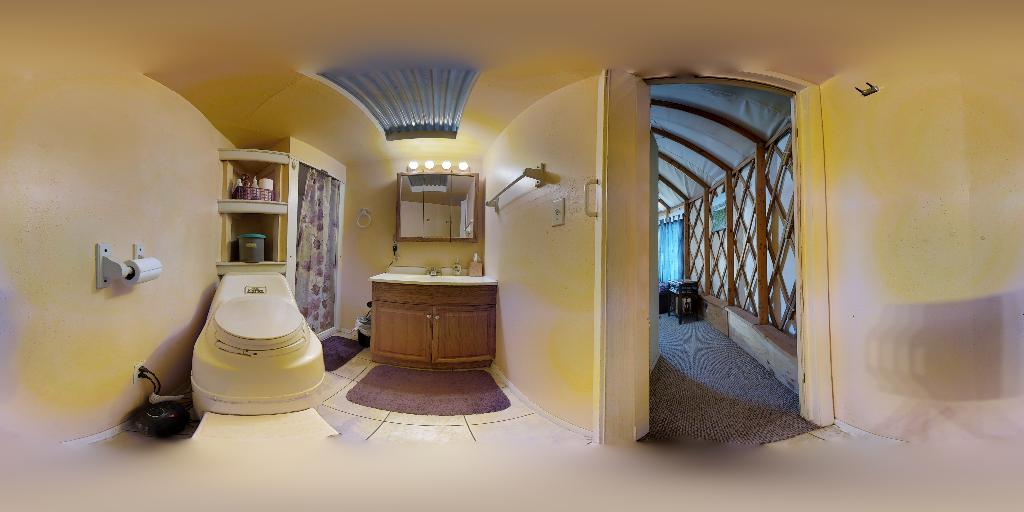}}
    \hfill
    \vspace{-0.3cm}
    \subfloat{\includegraphics[width=0.49\linewidth]{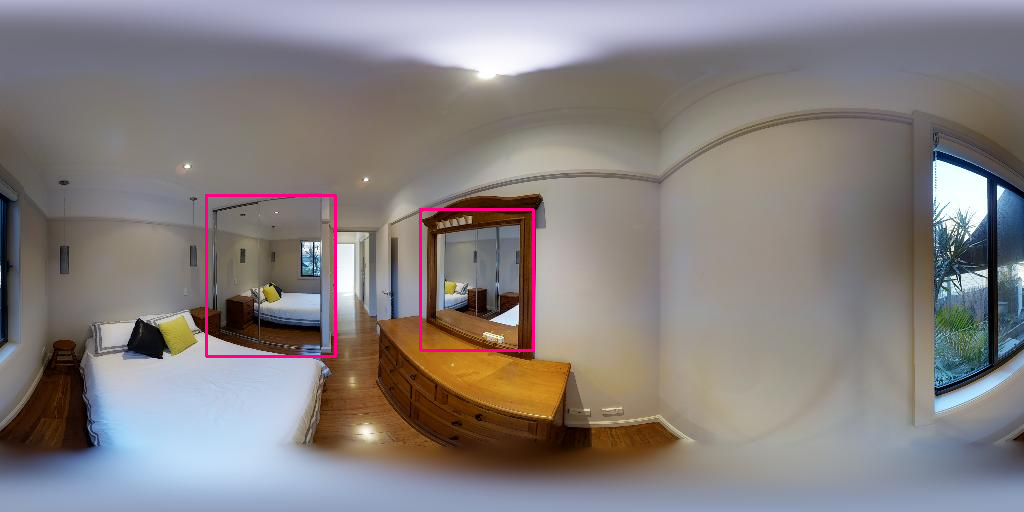}}
    \hfill
    \subfloat{\includegraphics[width=0.49\linewidth]{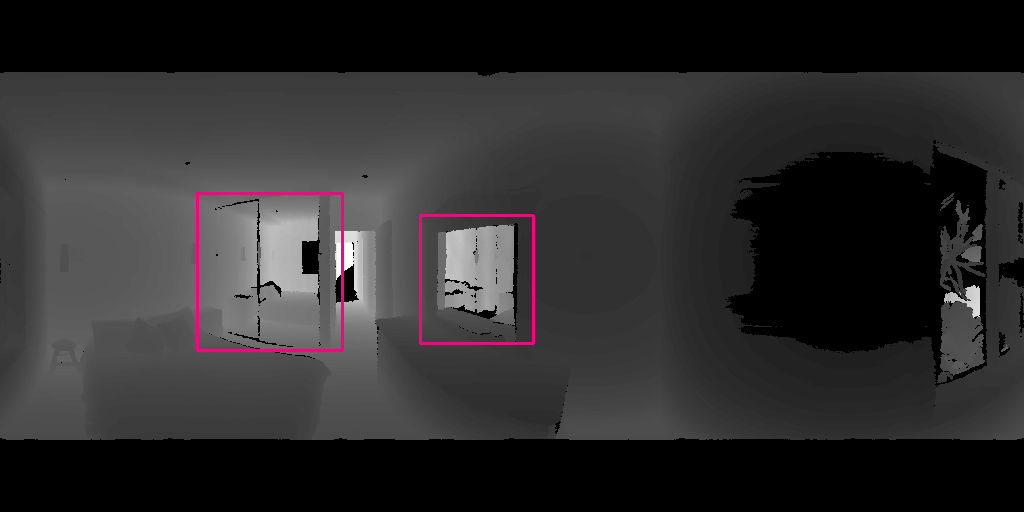}}
    \hfill
    \vspace{-0.3cm}
    \subfloat{\includegraphics[width=0.49\linewidth]{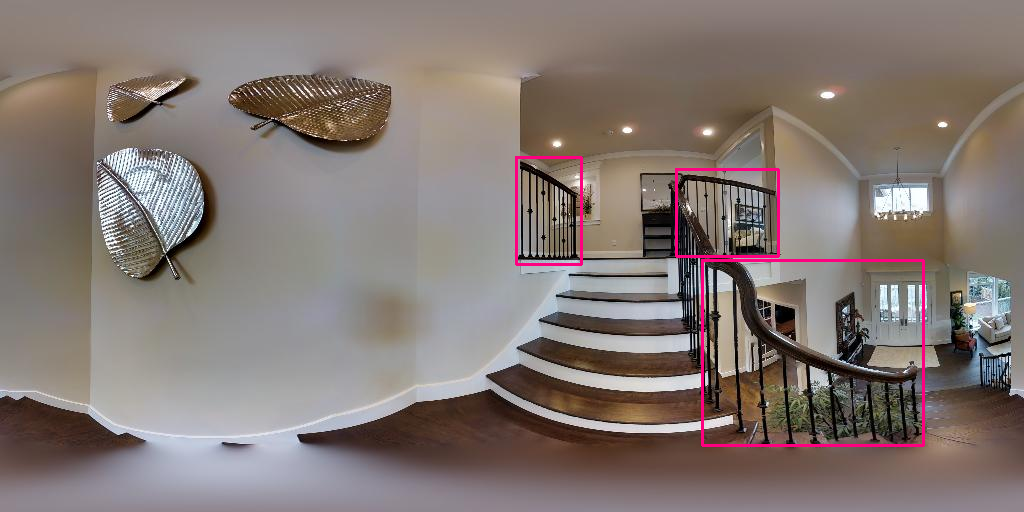}}
    \hfill
    \subfloat{\includegraphics[width=0.49\linewidth]{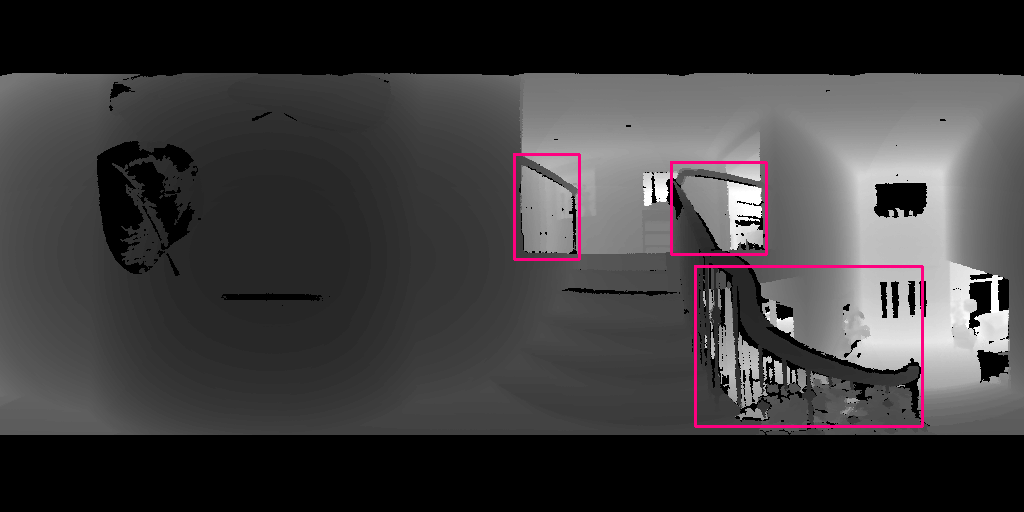}}
    \hfill
    
    \caption{
       \textbf{Top}: Qualitative differences between the rasterization and stitching-based semantic labels, and the 360V ray-casted ones at the corresponding $512 \times 256$ resolution.
       Aliasing and z-fighting artifacts manifest on the original Stanford2D3D samples (right), which are mitigated using our rendering approach (left), as depicted on the original colored semantic maps, as well as their annotated edge maps.
       The images are misaligned due to the horizontal shift induced by 360V's rendering of world-aligned viewpoints. \textbf{Bottom}: Different type of artifacts when stitching perspective camera views. The upper $2$ rows show the color cases (black pole regions and blurry inpaints), where the images are corrupted in unnatural ways, with important content being distorted (\textit{i.e.}~ceiling beams on the bottom left and stitching artifacts manifesting on the bottom right).
       The following $2$ rows showcase counterfactual depth artifacts arising from stereo-based ground truth estimation.
    }
\label{fig:semantic}
\end{figure}

Both Matterport3D and Stanford2D3D offer semantic annotations on the 3D building meshes which are acquired via scanning.
To generate high quality semantic labels at the lower resolutions that data-driven models are applied to, we integrate a modern material matte technique \cite{friedman2015fully} into our ray-casting pipeline.
This is a high quality matte technique that separates different materials into rendered mask layers.
As a result, to use it we first convert the per face labels into specific material indices that correspond to dummy materials that we generate for each label, and use another rendering pass to output our semantic labels.
Even though labels cannot be anti-aliased, using a high sampling ray-tracing rendering technique, we partly address artifacts that arise from the coarse scans.
The reconstructed meshes, albeit high-resolution (millions of vertices), span entire buildings.
This comes at a loss of fidelity, resulting in mesh artifacts that do not necessarily align with the original color images. 
This is not the case for our generated samples as they are rendered from the same meshes, instead of labelled on the meshes and associated with the camera-acquired samples, a problem that is also evident in the counterfactual nature of stereo depth maps as presented at the bottom of Figure~\ref{fig:semantic}.

While an approach similar to that used in \cite{armeni2017joint} would generally produce good enough labels, it would also suffer from higher frequencies of these artifacts. 
The semantic labels offered by Stanford2D3D are generated by stitching multiple perspective label renders.
There are a number of issues with this approach.
First, warping the perspective images into the equirectangular panorama necessitates sub-pixel sampling, which cannot be applied to semantic labels.
Instead, nearest neighbor sampling is used, which introduces noise, and to address that, high resolution renders are stitched into very high resolution (\textit{i.e.}~$4096 \times 2048$) panoramas.
But using these in a data-driven model typically requires downscaling, which again introduces nearest neighbor sampling artifacts.
Finally, traditional rasterization approaches suffer from z-fighting when aggregating multiple contributions into a single pixel.
Instead, ray-tracing aggregates all samples, which in our case corresponds to majority voting, a technique that reduces noise.
We qualitatively demonstrate the differences of our ray-traced semantic labels compared to those offered by Stanford2D3D at the top part of Figure~\ref{fig:semantic}.
At the same time, its bottom part showcases the problems incurred when relying on stitching perspective color images and using coupled stereo-based depth images for training data-driven models.

Since our focus lies on structured, indoor geometric understanding, we map their labels to the coarser set used by the NYUDepth dataset \cite{silberman2012indoor}, and then select the semantic labels that correspond to important structural elements of indoor scenes, instead of finer-grained furniture or object labels.
In total we offer $15$ labels focusing on the main structural elements and larger objects within indoor scenes, clustering finer grained ones into a couple of labels.

\subsection{Weak Layout Cues}
\label{sec:dataset_layout}
Given the semantic labels of each panorama, a subset of these includes the \textit{wall}, \textit{ceiling}, \textit{floor} labels which correspond to the scene's layout information.
Layout annotations usually involve the junctions/corners as these can reconstruct the layout in its entirety, assuming a Manhattan scene alignment.
However, annotating each single corner is a tedious task, and sometimes the scenes are far more complex, including extruding or interior spaces, making annotations ambiguous.
The semantic labels though correspond to the actual, complex layout, where complexity refers to non-Manhattan alignment or the scene's structure (interior spaces, multiple floors, scenes with stairs, unusual layouts, etc.).
But automatically extracting the layout from semantic labels needs to overcome a number of challenges, namely, the low quality annotations and foreground labels.

Our key observation here is that while the floor-to-wall boundaries are very frequently occluded with foreground objects, making the underlying layout edge invisible, it is the opposite of that for the ceiling-to-wall boundaries.
Even though occluding doors, structural beams, curtains, and ceiling mounted lights still manifest, they do at a much lower rate.
Another important issue is the lack of, or the low quality of the 3D annotations.
In the remainder of this section, we present a principled approach to largely handle all these issues and generate weak layout cues.

\subsubsection{Layout Boundary Detection}
\label{sec:layout_edge}
The rendered semantic labels can be partitioned into $4$ classes, with the $3$ being the layout classes, and the remainder class being the ``not layout'' class.
More specifically, doors, columns and beams are mapped to walls, and all other classes are marked as ``not layout''.
The resulting layout segmentation maps contain artifacts manifesting as holes due to objects placed on the floor, ceiling and walls.
Our goal is to seamlessly remove the invalid class to be able to apply a straightforward edge detection algorithm to identify the layout boundary.
To that end, we employ a standard segmentation post-processing technique, namely the conditional random fields (CRF) \cite{krahenbuhl2011efficient}.
We assign a confident (corresponding to $75\%$ probability) unary potential to the known layout classes and an uncertain (evenly distributed probability across all $4$ classes) unary potential to the ``not layout'' class.
We then formulate the following dense fully-connected CRF over the unary label distribution potential, and a bilateral pairwise potential defined over the surface orientation (normal) map:

\begin{equation}
\label{eq:crf}
    E_{CRF}(\mathbf{p}) = \sum_{\mathbf{p} \in \Omega} \psi_{unary}(\mathbf{p}) + \sum_{\mathbf{p} \in \Omega}\sum_{\substack{\mathbf{q} \in \Omega, \\ \mathbf{q} \neq \mathbf{p}}} \psi_{pairwise}(\mathbf{p}, \mathbf{q}).
\end{equation}

The bilateral pairwise potential includes a spatial term defined over the pixel domain, and a feature distance using each pixel's normal.
The rationale behind this choice is that we are labelling planar surfaces and the surface's direction at each location is highly correlated to the label type.
Minimizing Eq.\eqref{eq:crf} fills the holes and provides a cleaner layout segmentation map that can be used to extract edges corresponding to the layout boundary.
Exploiting the filled layout map, we follow a greedy approach and extract the first vertical edge in both directions, \textit{i.e.}~top-to-bottom, and bottom-to-top.

The predicted edges can be noisy which is a result of the coarse mesh-based segmentation annotation, and thus, we apply median filtering and a median absolute deviation \cite{leys2013detecting} outlier rejection strategy.
The more important issue that needs to be resolved though, is the inferior quality of the bottom (floor-wall) layout boundary, compared to that of the top (ceiling-wall) boundary.
To address this, we discard the extracted bottom boundary and instead reconstruct it from the top one.
The intermediate results from this step are depicted in Figure~\ref{fig:layout_reconstruction_qualitative}.

\subsubsection{Layout Boundary Completion}
\label{sec:layout_completion}

\begin{figure}[!htbp]
    \centering
        
    \subfloat{\includegraphics[height=0.35\linewidth]{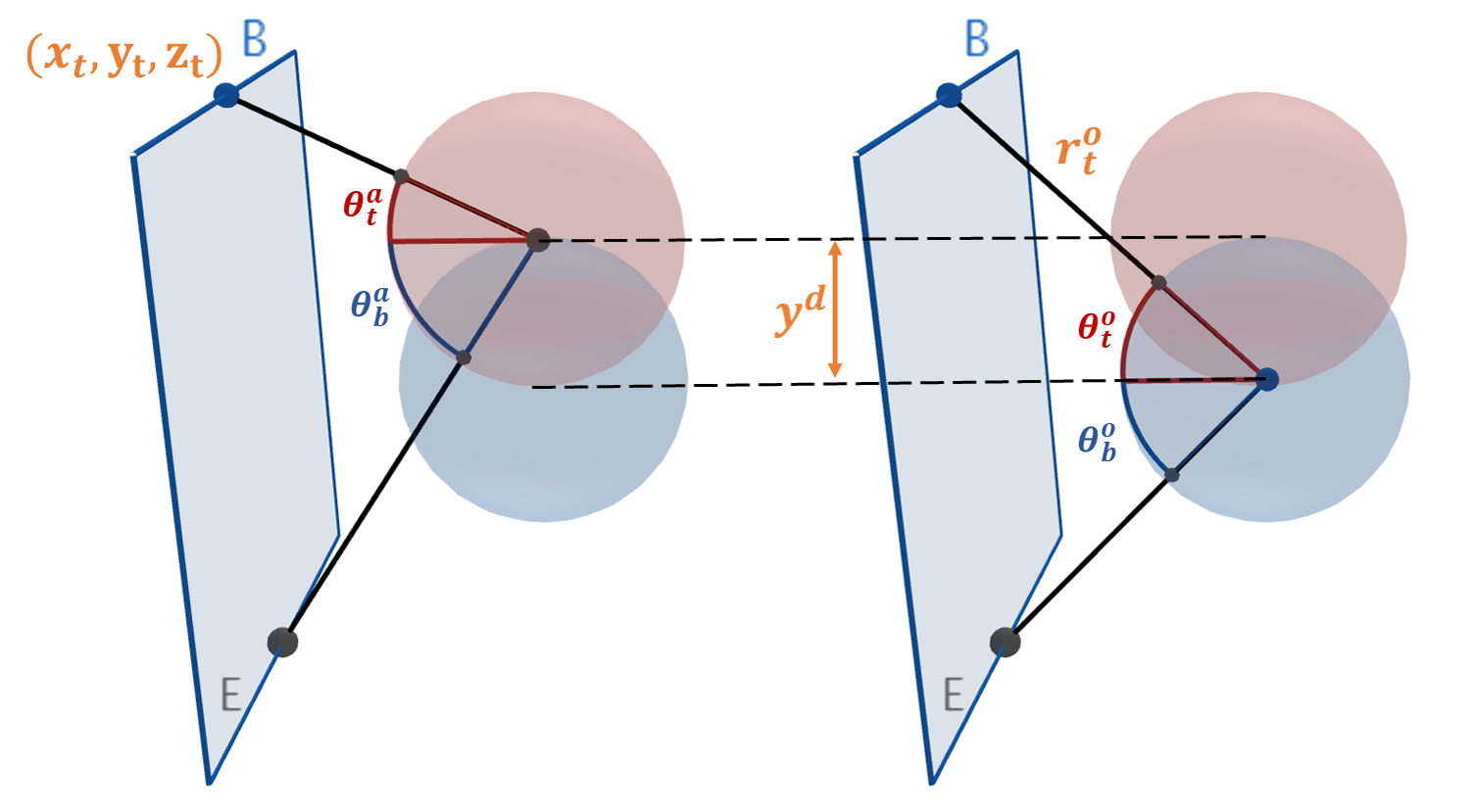}}
    \hfill
    \subfloat{\includegraphics[width=0.3\linewidth]{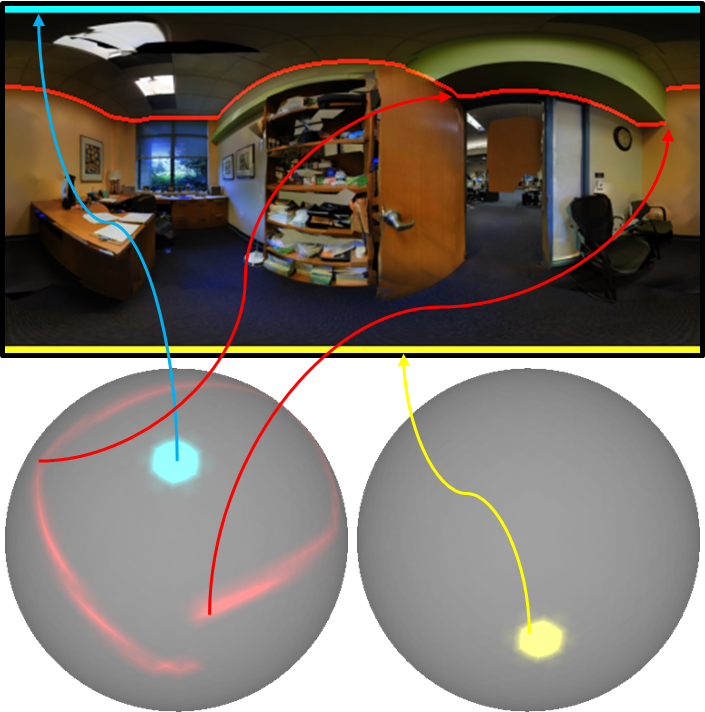}}
    \hfill

    \caption{
        The bottom layout reconstruction approach.
        Considering the known actual boundary at a specific meridian \textcolor{red}{$\theta^a_t$}, and the world position of the corresponding point $(x_t, y_t, z_t)$ (left), we can reconstruct the bottom layout boundary position \textcolor{blue}{$\theta_b^a$} in the panorama.
        Assuming a Manhattan aligned scene we need to estimate the vertical translation \textcolor{orange}{$y^d$} that will translate the current origin to the original vertical mid-point of the scene (middle).
        In that position, the distance of the bottom and top boundary, and by extension their latitudes \textcolor{red}{$\theta_t^o$}, \textcolor{blue}{$\theta^o_b$} with respect to the horizon are equal.
        As their latitudinal displacement is equal to, we only need to estimate $\gamma = \textcolor{blue}{\theta^o_t} - \textcolor{blue}{\theta^a_t}$, meaning only $\theta_b^o$ to reconstruct the bottom latitude at the current position $\textcolor{blue}{\theta_b^a} = \textcolor{blue}{\theta_b^o} + \gamma = \gamma - \textcolor{blue}{\theta_t^o}$.
        To estimate \textcolor{orange}{$y^d$} we need to sample the panorama depth at the zenith and nadir to find the true centroid and calculate its difference from the current ceiling position (zenith sampling).
        This process is differentiable to both the top layout latitude, the depth values at these latitudes, and the depth values at the poles (right).
    }
\label{fig:layout_reconstruction}
\end{figure}
\begin{figure}[!htbp]
    \centering
        
    \subfloat{\includegraphics[width=0.49\linewidth]{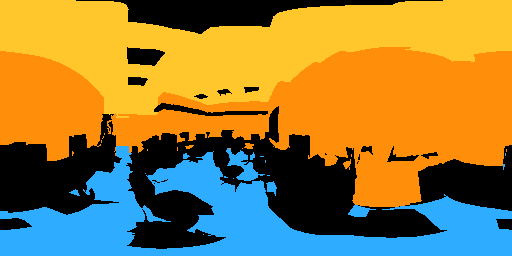}}
    \hfill
    \subfloat{\includegraphics[width=0.49\linewidth]{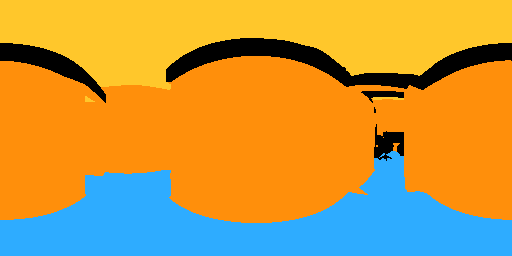}}
    \hfill
    \vspace{-0.3cm}
    \subfloat{\includegraphics[width=0.49\linewidth]{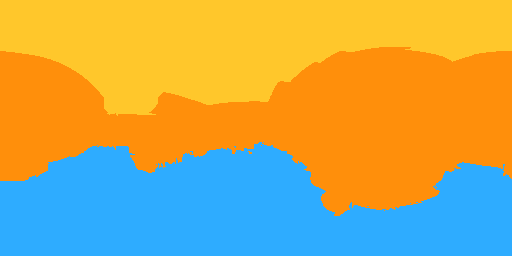}}
    \hfill
    \subfloat{\includegraphics[width=0.49\linewidth]{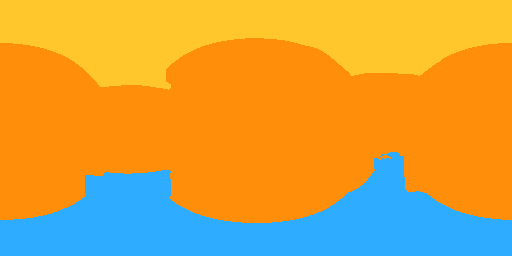}}
    \hfill
    \vspace{-0.3cm}
    \subfloat{\includegraphics[width=0.49\linewidth]{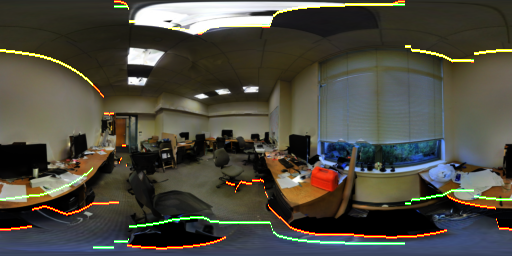}}
    \hfill
    \subfloat{\includegraphics[width=0.49\linewidth]{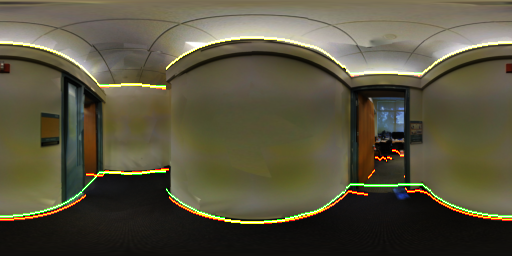}}
    \hfill
    \vspace{-0.3cm}
    \subfloat{\includegraphics[width=0.49\linewidth]{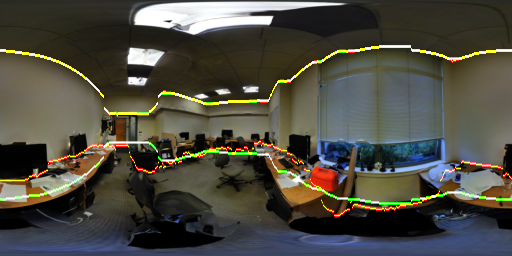}}
    \hfill
    \subfloat{\includegraphics[width=0.49\linewidth]{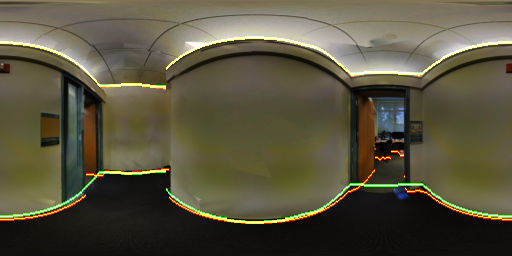}}
    \vspace{-0.3cm}
    \subfloat{\includegraphics[width=0.49\linewidth]{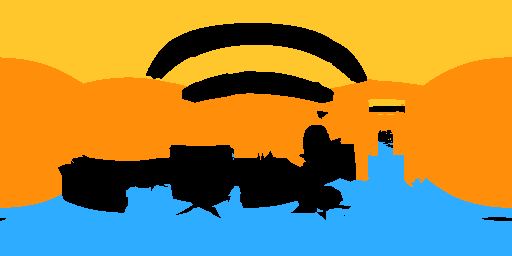}}
    \hfill
    \subfloat{\includegraphics[width=0.49\linewidth]{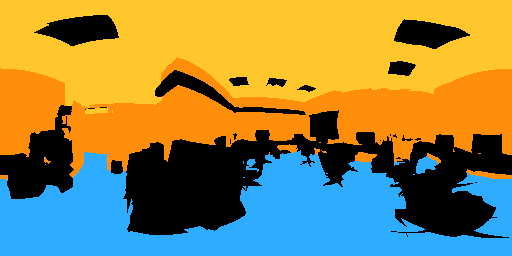}}
    \hfill
    \vspace{-0.3cm}
    \subfloat{\includegraphics[width=0.49\linewidth]{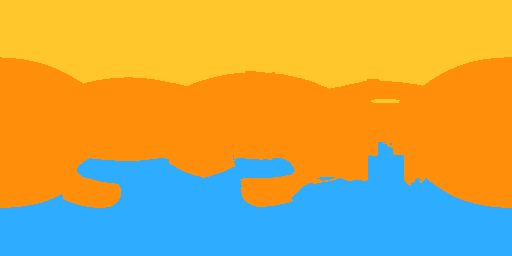}}
    \hfill
    \subfloat{\includegraphics[width=0.49\linewidth]{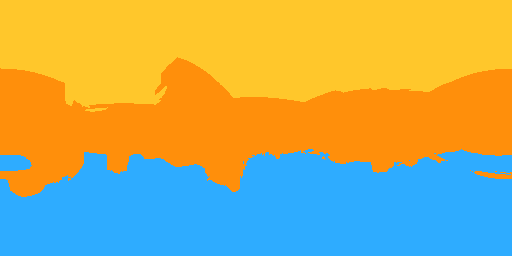}}
    \hfill
    \vspace{-0.3cm}
    \subfloat{\includegraphics[width=0.49\linewidth]{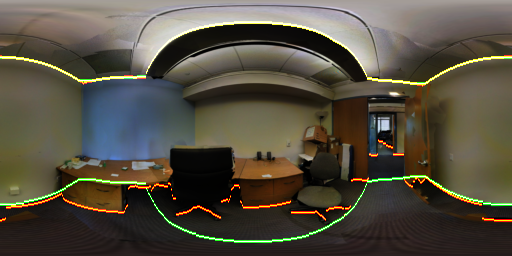}}
    \hfill
    \subfloat{\includegraphics[width=0.49\linewidth]{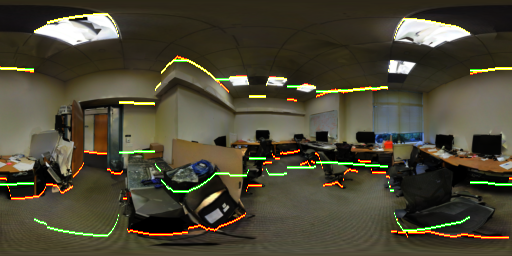}}
    \vspace{-0.3cm}
    \subfloat{\includegraphics[width=0.49\linewidth]{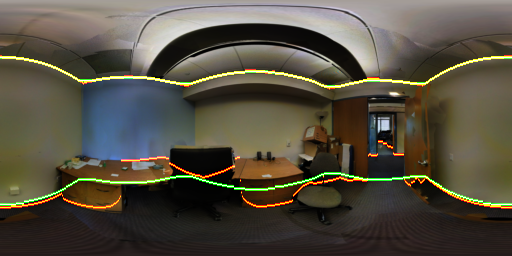}}
    \hfill
    \subfloat{\includegraphics[width=0.49\linewidth]{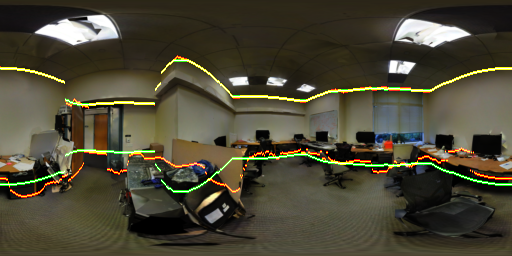}}
    \hfill
    \caption{
        Final and intermediate results from the weak layout cue calculation process for $4$ scenes in a top to bottom presentation.
        Initially, the input layout segmentation map is presented, with \textcolor{Dandelion}{gold}, \textcolor{orange}{orange}, \textcolor{cyan}{cyan}, and \textcolor{black}{black} the colors for the \textit{ceiling}, \textit{walls}, \textit{floor}, and the \textit{foreground} classes respectively.
        Following vertically, the processed CRF result that fills the foreground using the normal map guidance.
        In this representation we apply greedy vertical edge detection, median filtering and MAD outlier rejection.
        Finally, the last two rows for each example present the original detected boundaries with \textcolor{orange}{orange}, as well as the filtered top and reconstructed bottom boundaries in  \textcolor{ForestGreen}{green}, with the overlap on the top resulting in \textcolor{yellow}{yellow}, for the two above layout segmentation maps respectively.
        The qualitative difference between the original top and bottom boundaries is clearly depicted, as well as the CRF optimization gains.
    }
\label{fig:layout_reconstruction_qualitative}
\end{figure}

Under the Manhattan world assumption, the ceiling/floor are parallel and horizontally aligned, and the walls are perpendicular to them ($y$ axis aligned, \textit{i.e.} vertically oriented).
As a result, the walls' top and bottom boundaries, when viewed from a world aligned \360 camera, positioned at the centroid of the scene, project to mirrored latitude coordinates with respect to the equator.
As presented in \cite{zioulis2019spherical} the spherical coordinate partial derivatives are:

\begin{equation}
    \begin{aligned}
    \label{eq:spherical_derivs}
                \left[\begin{matrix}
                    \partial{r}     \\
                    \partial{\phi}  \\
                    \textcolor{teal}{\partial{\theta}}\\
                \end{matrix}\right]
                =\left[\begin{matrix}
                    \sin\phi\sin\theta & \cos\theta & \cos\phi\sin\theta \\
                    \frac{\cos\phi}{r\sin\theta} & 0 & \frac{-\sin\phi}{r\sin\theta}                   \\ \frac{\sin\phi\cos\theta}{r} & \textcolor{teal}{\frac{-\sin\theta}{r}} & \frac{\cos\phi\cos\theta}{r}
                \end{matrix}\right]
                \left[\begin{matrix}
                    \partial{x} \\
                    \textcolor{teal}{\partial{y}} \\
                    \partial{z}
                \end{matrix}\right].
    \end{aligned}
\end{equation}

\noindent From these, the latitudinal partial derivative for purely vertical Cartesian displacement is 
\begin{equation}
\label{eq:disparity}
    \textcolor{teal}{\partial{\theta} = \frac{-\sin\theta}{r} \partial{y}}
\end{equation}

\noindent Therefore, according to Eq.(\ref{eq:disparity}), for purely vertical displacements along the $y$ axis, the latitude displacement of the wall top and bottom edges are equal, as they share the same radius when the camera resides at the vertical centroid, and because of sine's trigonometric reflection at $\nicefrac{\pi}{2}$.
Consequently, calculating the top latitude angular displacement $\gamma$ between the current top wall latitude and that of a viewpoint at the vertical center of the room, allows for the calculation of the true latitude of the bottom wall edge as:
\begin{equation}
\label{eq:latitude_disparity}
    \gamma = \theta_t^o - \theta_t^a = \theta_b^a - \theta_b^o.
\end{equation}
To reconstruct the bottom boundary, we need to estimate $\theta_t^o$ which is the top wall's latitude at the ceiling and floor mid-point viewpoint.
Each rendered ground-truth depth map $\mathbf{D}$ can be transformed to a structured point cloud $\mathbf{V}(\mathbf{p}) \in \mathbb{R}^3, \mathbf{p} \in \Omega$, defined in the image domain using Eq.\eqref{eq:spherical_cartesian_coords}.
By sampling the scene's zenith and nadir, we estimate $\bar{y}_b$ and $\bar{y}_t$, which is the average vertical position of the ceiling and floor, from which we calculate the scene's height $h = \bar{y}_t - \bar{y}_b$.
The vertical displacement can be found by $y^d = \bar{y}_t - \bar{y}_m$, after estimating the vertical mid-point $\bar{y}_m = 0.5 h$.
Using that we extract the distance at the mid-point, and subsequently the latitude at the origin:
\begin{align}
    r^o_t &= \sqrt{x_t^2 + (y_t + y^d)^2 + z^2} \\
    \theta_t^o &= \cos^{-1}\frac{y_t + y^d}{r_t^o}.
\end{align}
This way we can reconstruct the bottom wall edge from the top wall edge, its distance, and the distances of the zenith and nadir.
This provides a more robust estimation of the bottom wall edge, as it overcomes the aforementioned issues related to the segmentation map extracted bottom boundary, completing the layout boundary information, as presented in Figure~\ref{fig:layout_reconstruction_qualitative}.

\begin{table*}[!htbp]
\centering
\caption{Comparison with other panorama datasets. The 360V dataset is the only real world domain dataset to offer $\ge 10.000$ samples with layout information, albeit at the form of weak cues and not corners. In addition, apart from not suffering from stitching artifacts, it is the only dataset offering stereo viewpoints.}
\label{tab:dataset}
\begin{tabular}{@{}lccccccccc@{}}
\toprule
 &
  Domain &
  Panorama &
  \rotatebox[]{60}{Color} &
  \rotatebox[]{60}{Depth} &
  \rotatebox[]{60}{Normal} &
  \rotatebox[]{60}{Semantic} &
  \rotatebox[]{60}{Layout} &
  \rotatebox[]{60}{Stereo} &
  \rotatebox[]{60}{Samples} \\ \midrule
SUN360 \cite{xiao2012recognizing} &
  Real &
  Equirect. &
  {\color[HTML]{9AFF99} \cmark} &
  {\color[HTML]{FD6864} \xmark} &
  {\color[HTML]{FD6864} \xmark} &
  {\color[HTML]{FD6864} \xmark} &
  {\color[HTML]{9AFF99} \cmark} $(511)$ &
  {\color[HTML]{FD6864} \xmark} &
  $67583$ \\
Stanford2D3D \cite{armeni2017joint} &
  Real &
  Stitched &
  {\color[HTML]{9AFF99} \cmark} &
  {\color[HTML]{9AFF99} \cmark} &
  {\color[HTML]{9AFF99} \cmark} &
  {\color[HTML]{9AFF99} \cmark} &
  {\color[HTML]{9AFF99} \cmark} $(550)$ &
  {\color[HTML]{FD6864} \xmark} &
  $1413$ \\
Matterport3D \cite{chang2018matterport3d} &
  Real &
  Stitched &
  {\color[HTML]{9AFF99} \cmark} &
  {\color[HTML]{9AFF99} \cmark} &
  {\color[HTML]{FD6864} \xmark} &
  {\color[HTML]{FD6864} \xmark} &
  {\color[HTML]{9AFF99} \cmark} $(2295)$ &
  {\color[HTML]{FD6864} \xmark} &
  $10800$ \\
Kujiale \cite{jin2020geometric} &
  Synthetic &
  Equirect. &
  {\color[HTML]{9AFF99} \cmark} &
  {\color[HTML]{9AFF99} \cmark} &
  {\color[HTML]{FD6864} \xmark} &
  {\color[HTML]{FD6864} \xmark} &
  {\color[HTML]{9AFF99} \cmark} &
  {\color[HTML]{FD6864} \xmark} &
  $3550$ \\
Structured3D \cite{Structured3D} &
  Synthetic &
  Equirect. &
  {\color[HTML]{9AFF99} \cmark} &
  {\color[HTML]{9AFF99} \cmark} &
  {\color[HTML]{9AFF99} \cmark} &
  {\color[HTML]{9AFF99} \cmark} &
  {\color[HTML]{9AFF99} \cmark} &
  {\color[HTML]{FD6864} \xmark} &
  $21835$ \\
360V (Ours) &
  Scanned &
  Equirect. &
  {\color[HTML]{9AFF99} \cmark} &
  {\color[HTML]{9AFF99} \cmark} &
  {\color[HTML]{9AFF99} \cmark} &
  {\color[HTML]{9AFF99} \cmark} &
  {\color[HTML]{9AFF99} \cmark} (cues) &
  {\color[HTML]{9AFF99} \cmark} (4) &
  $12213$ \\ \bottomrule
\end{tabular}
\end{table*}

We refer to our automatically calculated layout annotations as weak, as they still rely on the quality of the semantic annotations, and the availability of depth measurements, which both contain holes, and, also, despite the CRF optimization and filtering, noise is not fully eliminated.
In addition, compared to traditional layout annotations that start from the layout junctions/corners, we only provide the (sometimes noisy) boundaries, offering only cues about the true layout.
An important differentiation from traditional layout annotations, is that due to the availability of the \textit{down} and \textit{up} generated virtual viewpoints, our dataset is not biased towards central viewpoints, and instead offers viewpoints skewed towards the floor or the ceiling too.
Finally, not all scenes provide the necessary information for extracting the top layout (missing semantic annotations), or reconstructing the bottom one (depth holes).
These cases are identified, resulting in a layout validity mask which is used to disregard the ground-truth during training.

Concluding, our dataset contains $4$ stereo color viewpoints, with matching depth, semantic and normal maps, as well as weak layout cues.
As Table~\ref{tab:dataset} shows, it is the only real-world domain (\textit{i.e.}~scanned) dataset to offer such a large number of multi-modal annotations combined with layout information.

\section{Explicitly Connected Layout \& Depth}
\label{sec:method}
\begin{figure*}[!htbp]
    \centering
    \includegraphics[width=\linewidth]{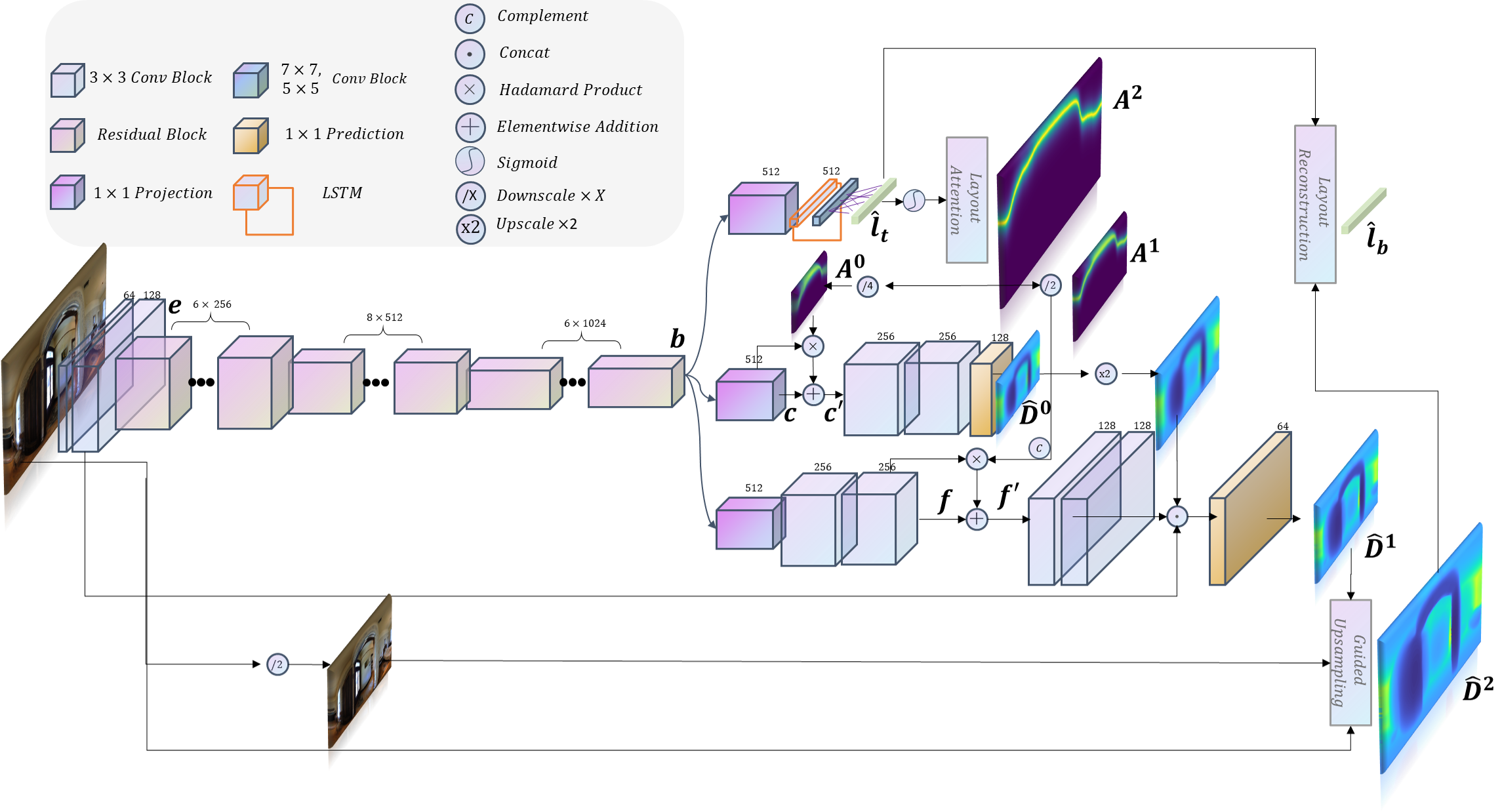}
    
    \caption{
        An illustration of our model's network architecture as described in Section~\ref{sec:architecture}.
        The legend on the top left presents the different components used in the figure.
        A shared encoder on the left encodes a latent representation $\mathbf{b}$.
        This gets fed into $4$ different branches, namely, layout decoder prediction branch that predicts the top layout boundary $\mathbf{\hat{l}}_t$, and the coarse and finer depth decoder branches, that predict the depth at two different scales, $\mathbf{\hat{D}}^0$ and $\mathbf{\hat{D}}^1$ respectively.
        Depth predictions and layout attention maps $\mathbf{A}^s$ propagate from one scale $s$ to the next.
        The attention maps are used in a residual attention mechanism to boost the necessary depth features in the coarse ($\mathbf{c}$) and finer ($\mathbf{f}$) branches, with the complement attention applied to the latter.
        The finest depth scale $\mathbf{\hat{D}}^2$ is predicted by a learnable guided upsampling filter, and is used to reconstruct the bottom layout boundary $\mathbf{\hat{l}}_b$.
    }
\label{fig:model}
\end{figure*}

Our goal is to exploit the larger-scale availability of layout and depth annotated data to improve the performance of panorama-based monocular depth estimation.
The key intuition is that both tasks are complementary indoor geometric understanding tasks.
While prior works \cite{jin2020geometric,zeng2020joint} have only implicitly associated the two tasks, they also used high quality layout annotations.
Instead, we rely on weak layout annotations, and additionally explicitly couple the two tasks.
In this section we will describe the details of our approach, and more specifically, the dual task architecture (Section~\ref{sec:architecture}), the explicit integration of the two tasks (Section~\ref{sec:explicit}), and our supervision scheme (Section~\ref{sec:supervision}).

\subsection{Network Architecture}
\label{sec:architecture}

Our network uses a shared encoder and two different decoding branches, one for each task\, with Figure~\ref{fig:model} presenting its network architectural overview in a schematic manner and will be used as a reference in the following detailed description..
We use efficient spherically padded convolutions \cite{zioulis2021single} to address the boundary discontinuity, with ReLU activations \cite{nair2010rectified} and batch normalization \cite{ioffe2015batch} throughout the model.
On the left of Figure~\ref{fig:model}, we present the shared encoder which is a ResNet \cite{he2016deep}  with pre-activated residual blocks \cite{he2016identity} that encodes a shared latent representation $\mathbf{b}$ for both tasks.
The residual blocks are preceded by a two convolutional block stem module for early feature extraction that generates features $\mathbf{e}$ which get fed into the residual units.
presents the network architecture's overview in a schematic manner and will be described in the following paragraphs.

From the middle to the right, four decoder branches follow, with the top being the layout decoder, which is inspired by HorizonNet \cite{sun2019horizonnet}.
We represent the boundary using the normalized height from the horizon, which is the panorama equator, and leverage recurrent layers to efficiently capture the global structure, as demonstrated by HorizonNet.
After a single convolution block feature extraction, we perform height pooling, squeezing the resulting features into a vector capturing the boundary's features, in contrast to HorizonNet which uses multi-scale features from its encoder.
This gets fed into a 2 layer bidirectional LSTM \cite{hochreiter1997long}, followed by a linear prediction layer with a \textit{sigmoid} activation function \cite{han1995influence}.
The LSTM and linear layer cascade ensures the effective capturing of the global context when estimating the scene's encoded top layout boundary.
Notably, compared to HorizonNet which predicts both the top and bottom layout boundaries, we only predict the top one, which results in a single vector prediction.

The vertically following three branches jointly represent a multi-headed depth decoder which is designed to process information at different scales in different branches, with each lower scale prediction used in the subsequent scale branch. 
The rationale behind this design decision is that each scale requires different features, and should improve upon the results of the former by adding more detail.
The first depth branch focuses on the coarsest resolution, and first projects the encoded bottleneck features, applies bilinear upsampling and uses two convolution blocks followed by a prediction layer.
Prior to the upsampling, the layout-based attention map (see Section~\ref{sec:explicit}) is applied on these low spatial resolution features that are used to predict the coarse depth map.

The second depth branch captures a finer resolution by again projecting the encoded bottleneck features, followed by a cascade of upsampling and convolution blocks, before predicting the refined depth map from the incoming features concatenation with the coarse scale prediction and the early stem features.
Prior to the second upsampling, a complement attention is applied using the layout attention map to the features predicting the medium scale depth map.
These first two depth branches follow a \textit{project}-\textit{upsample}-\textit{convolve} processing cascade, separating learning across different scales.
While the first depth branch only uses spatially squeezed information, the second branch spatially upscales one octave\footnote{Defined as a scaling of the spatial dimensions by a power of $2$.\cite{lindeberg2013scale,chen2019drop}} up from the coarse branch, and re-uses the coarse prediction, as well as the early encoded features capturing higher frequency information.
Finally, the last -- fine resolution -- depth branch contains minimal learnable parameters and instead applies a learnable guided depth map upsampling layer \cite{wu2018fast} using the input color image and the middle resolution predicted depth map from the second branch.

\subsection{Layout-based Attention \& Depth-based Layout Reconstruction}
\label{sec:explicit}

While Section~\ref{sec:architecture} outlines the model's architecture, in this section we present the building blocks connecting the two tasks in a complementary manner.
Prior works that jointly consider the layout and depth estimation tasks in parallel, only implicitly capture their the inter-task dependencies using exclusively learnable parameters.

\textbf{Layout-based Attention.}
Even though layout estimation is partly counterfactual with respect to the observed scene, it offers important information regarding the scene's scale.
As explained in Section~\ref{sec:dataset_layout} the ceiling typically provides cleaner information compared to the floor, whose nature is more counterfactual due to the heavier presence of objects.
Consequently, we use the top layout estimations in an attentive manner in our model.
As each layout prediction $\mathbf{l}_t(\phi) \in \mathbf{R}^{W}$ represents the normalized height, it is converted to angular coordinates $\mathbf{l}_t^a = \frac{\pi}{2} (\mathbf{1} - \mathbf{l}_t)$.
Using this vectorized top boundary representation that spans the entire image width $W$, or otherwise, the complete azimuth range on the sphere, we create a layout-based attention map by differentiably reconstructing a latitude Gaussian for each meridian:
\begin{equation}
\label{eq:meridian_gaussian}
    \mathbf{A}(\phi, \theta) = \frac{1}{\sigma\sqrt{2\pi}} exp\big({\frac{-h(\theta,\mathbf{l}_t(\phi))}{2\sigma^2}}\big),
\end{equation}
where $h(\theta_1, \theta_2)$ is the haversine (or otherwise great circle) distance, adapted for angular coordinates of equal longitude and a unit radius:
\begin{equation}
\label{eq:harvesine}
    h(\theta_1, \theta_2) = 2\sin^{-1}\sin(\frac{\theta_2 - \theta_1}{2}).
\end{equation}

This way, we reconstruct an attention map around the predicted top layout boundary and use it to attend to the first two (coarse and medium scale) depth decoder branches.
Initially, as the layout offers an important hint about the scene's global scale and coarse relative depth, we use the attention map $\mathbf{A}$ in the depth decoder's coarse prediction branch.

After projecting the bottleneck features $\mathbf{b}$ to the coarse branch's entry features $\mathbf{c}$, similar to \cite{jin2020geometric}, we use a residual attention mechanism $\mathbf{c}' = \mathbf{c} + (\mathbf{c}  \cdot \mathcal{D}(\mathbf{A}))$, where $\mathcal{D}$ is a spatial downsampling operation to align the resolution of the feature map and the attention map.
This boosts the features spatially associated to the top layout boundary region, allowing the model to better reason about the global spatial context.
Inspired by \cite{tao2020hierarchical}, we additionally use the layout reconstructed attention map in the subsequent branch used to predict the next octave depth map.
However, re-using the same attention map would only boost the features around the same spatial area, which, given our hierarchical multi-scale prediction architecture, are already provided in the concatenated coarse scale depth map.
Instead, we use the complement attention in this medium scale prediction $\mathbf{f}' = \mathbf{f} + (\mathbf{f} \cdot (\mathbf{1} - \mathcal{D}(\mathbf{A})))$, which also additionally receives the skip connection features coming from the encoder, boosting the features that were previously diminished in the coarse scale's residual attention.

Under this scheme, the finer scale prediction layer receives the coarse scale predicted depth map, the early encoder detail preserving features $\mathbf{e}$, and the boosted, non\hyp{}coarse representation encoded by the network, to improve upon the coarse scale estimation.
The finest scale prediction, uses learnable guided filter upsampling, that contains minimal learnable parameters. Therefore, this medium scale prediction -- between the coarse and finest -- is the branch mostly responsible for separating the foreground from the background.
This generally provides the overview of our architecture and inter-twined attention mechanism, with the coarse branch being attended to the scene's scale and relative depth, and the finer -- medium scale -- branch receiving higher frequency information and being attended to it, in order to focus on predicting the foreground depth.
Since our layout cues are weak, the attention map may suffer from similar artifacts. To reduce their effect, we perform a two-pass spherically padded Gaussian blur on the reconstructed attention maps.

\textbf{Depth-based Layout Reconstruction.}
As presented in Section~\ref{sec:layout_completion}, reconstructing the bottom layout boundary is a function of the predicted top layout boundary latitude $\mathbf{l}_t(\phi)$, the ceiling and floor heights $\bar{y}_t$ and $\bar{y}_b$ respectively, which are also a function of the predicted depth, and the depth values at the top boundary $r_t(\phi) = \mathcal{S}(\mathbf{D}, \mathbf{l}_t) \in \mathbb{R}^W$, where $\mathcal{S}$ is a sampling operation across the panoramas meridians, sampling the depth map $\mathbf{D}$ at the latitudes given by $\mathbf{l}_t$.
As a result, the reconstruction of the bottom boundary is differentiable with respect to both the predicted top boundary, as well as certain specific areas of the predicted depth map.
For these, similar to when reconstructing the annotations from the ground-truth depth map, we sample a number of panorama rows $k$ from the estimated Cartesian coordinate $\mathbf{V}$ at the top (zenith) and bottom (nadir) to extract the average heights.
In addition, when training, we extract the mean depth value across each meridian within a window $w$ around the predicted layout latitude.
Therefore, supervising the bottom boundary under this reconstruction process, backpropagates to both the layout task (predicted top boundary locations), as well as the depth estimation task (predicted depth values at the zenith, nadir, and top layout regions).

\subsection{Multi-scale Supervision}
\label{sec:supervision}

We supervise both tasks, with the estimated depth maps being supervised in three scales $s \in \{0, 1, 2\}$ from coarse to fine, resulting in the following combined loss function:
\begin{equation}
\label{eq:total_loss}
    \mathcal{L} = \sum_s \mathbf{W}^s \mathbf{M}(\mathbf{D}^s) \mathcal{L}_D^s + \lambda_L \mathcal{L}_L.
\end{equation}
In this and following loss functions, we omit the pixel-based indexing and averaging over the image or boundary domain for brevity.
The mask $\mathbf{M}$ is a combined validity and depth range mask that ignores invalid pixels, and those lying out of the trained depth range, while the weights $\mathbf{W}^s$ are the spherical weights used in \cite{zioulis2019spherical} for each corresponding scale.

\textbf{Layout Supervision.}
For the layout loss we use the haversine distance as given in Eq.(\ref{eq:harvesine}):
\begin{equation}
\label{eq:layout_loss}
    \mathcal{L}_L = \mathbf{l}_t^m \cdot h(\mathbf{\hat{l}}_t, \mathbf{l}_t) + \mathbf{l}_b^m  \cdot h(\mathbf{\hat{l}}_b, \mathbf{l}_b),
\end{equation}
with $\mathbf{l}^m$ denoting the layout validity mask, preventing backpropagation from invalid layout boundaries.
The haversine distance implicitly clips the gradient due to its periodicity.
This is important as compared to other regression losses, it prevents the model from destabilizing during the early phases of training, where the depth predictions can greatly vary and saturate the layout branch which uses a \textit{sigmoid} activation. 
Compared to other works that jointly train layout and depth models, this allows the single-shot training of our model.
In contrast, both \cite{jin2020geometric} and \cite{zeng2020joint} progressively train their models, necessitating multiple incremental disjoint trains as each stage is added.

\textbf{Depth Supervision.}
The depth loss is a combination of different objectives that aim to address the complexity of the task:
\begin{equation}
\label{eq:depth_loss}
    \mathcal{L}_D^s = \lambda_{L1}^s \mathcal{L}_{log}(\hat{\mathbf{D}}^s, \mathbf{D}^s) + \lambda_{V}^s \mathcal{L}_{V}(\hat{\mathbf{D}}^s, \mathbf{D}^s) + \lambda^s_{S} \mathcal{L}_{S}(\hat{\mathbf{D}}^s, \mathbf{D}^s).
\end{equation}
The first term, $\mathcal{L}_{log}$, is the logarithm of the L1 depth error as presented in \cite{hu2019revisiting} that balances the loss with respect to both nearby and far away depths.
The second term, $\mathcal{L}_{V}$, is the virtual normal error as presented in \cite{yin2019enforcing} that captures longer distance dependencies between the predicted depth maps, it is thus, oriented towards preserving the global relative depth within the scene.
The third term, $\mathcal{L}_{S}$, is the surface loss, which corresponds to the cosine distance between the ground-truth and predicted surface orientation.
The normal maps are calculated from the depth maps after lifting them to their Cartesian coordinates $\mathbf{V}$ and extracting local surface information via the cross product of the vertical and horizontal central finite differences.
This loss preserves local smoothness, which is a reasonable prior to enforce as depth maps generally follow a strong piece-wise smooth prior \cite{huang2000statistics}, which is even more pronounced in our indoor scene specific context.

\section{Results}
\label{sec:results}
\subsection{Implementation Details}
\label{sec:implementation}
The input to our model is a \360 panorama of a $512 \times 256$ resolution, which is also the resolution of the output depth map.
We initialize our model's convolution and linear layer parameters using \cite{he2015delving}, and our 2-layer LSTM using \cite{glorot2010understanding}.
For the LSTM we also use a $50\%$ dropout. 
We use the default parameterized Adam \cite{kingma2014adam} optimizer with no weight decay, a learning rate of $0.0002$ and a batch size of $8$.
Since Blender's coordinate system is different than the one introduced in Section~\ref{sec:preliminaries}, we adapt all equations using the appropriate trigonometric reflections.
For the CRF we use a spatial standard deviation of $\sigma_{2D} = 7.0$ for the unary term, a $\sigma_{2D} = 35.0$ for the spatial standard deviation, and a $\sigma_{3D} = 0.2$ for the surface orientation bilateral term of the global normal bilateral term.
We optimize for $5$ iterations using the Potts compatibility model.

We use a fixed seed across all experiments for all random number generators.
Our models are implemented using PyTorch \cite{paszke2017automatic}, and PyTorch-Lightning \cite{falcon2019pytorch}.
Our loss weights are $\lambda_{L1} = [0.15, 0.1, 0.05]$ for coarse-to-fine scales respectively, $\lambda_V = [0.1, 0.1, 0.05]$, $\lambda_{S} = [0.1, 0.1, 0.05]$, and $\lambda_L = 0.05$.
Both the loss weights and CRF parameters were selected via a heuristic search with an empirically defined parameter space.
For the former we used a low epoch count (\textit{i.e.}~$5$) training scheme on M3D, while for the latter we relied on the manual inspection of selected samples.
For the virtual normal loss, we use the default parameters as presented in \cite{yin2019enforcing} and a $15\%$ sampling ratio.
The angular standard deviation used in Eq.\eqref{eq:meridian_gaussian} when reconstructing the layout-derived attention map corresponds to $9.5^\circ$, and the subsequent low-pass filtering uses a kernel size of $5$ and a spatial standard deviation of $1.0$ pixel.
For the boundary reconstruction we set the number of sampled rows for the heights and boundary at $k = w = 3$.
When training we use random circular shift and flip augmentations with a $75\%$ and $50\%$ probability respectively for the color and depth panoramas and the layout boundaries, as well as random gamma, brightness and contrast augmentations for the color images, with a $80\%$ probability of applying them both simultaneously.
We additionally employ random erasure augmentations \cite{fernandez2020corners} with a $50\%$ probability.

\subsection{Metrics}
\label{sec:metrics}
For the quantitative assessment of our approach, we use the standard metrics in the literature for depth estimation \cite{eigen2014depth} (without median scaling), and the RMSE for the layout boundaries.
We additionally define two performance indicators for the two tasks that aggregate metrics into a single quantity:
\begin{gather}
    I_D = (1 - \nicefrac{\delta_1}{100}) \times \text{RMSE}, \label{eq:depth_indicator} \\
    I_L = \text{RMSE}_{top} \times \text{RMSE}_{bottom} \times  1000. \label{eq:layout_indicator}
\end{gather}
We split the top and bottom layout RMSEs as one is directly inferred, while the other is also reconstructed from the predicted depth.
We use these to select the best performing models as well as comparing the inter-task performance. 

\subsection{Performance Analysis}
\label{sec:performance}
\begin{table*}[!htbp]
\centering
\caption{Results on the Matterport3D dataset. Bold red, orange and yellow denote the best, second-best and third-best performances, while the arrows next to each metric show the direction of better performance.}
\label{tab:m3d_sota}
\begin{tabular}{lccccccc}
\hline
 &
  \multicolumn{1}{l}{AbsRel $\downarrow$} &
  \multicolumn{1}{l}{SqRel $\downarrow$} &
  \multicolumn{1}{l}{RMSE $\downarrow$} &
  \multicolumn{1}{l}{RMSLE $\downarrow$} &
  \multicolumn{1}{l}{$\delta^1 \uparrow$} &
  \multicolumn{1}{l}{$\delta^2 \uparrow$} &
  \multicolumn{1}{l}{$\delta^3 \uparrow$} \\ \hline
ELD (Ours)                          & \first{0.1136}  & \first{0.0707}  & \first{0.4066}  & \first{0.0743}  & \first{88.46\%}   & \first{97.10\%}   & \first{98.91\%}   \\
BiFuse \cite{wang2020bifuse}        & \second{0.1193} & \second{0.0861} & \second{0.4220} & \third{0.0946}  & \second{87.55 \%} & \second{96.94 \%} & \second{98.78 \%} \\
UResNet \cite{zioulis2018omnidepth} & 0.1482          & 0.1022          & 0.4611          & 0.0901          & 79.72\%           & 94.51\%           & 98.10\%           \\
RectNet \cite{zioulis2018omnidepth} & \third{0.1351}  & \third{0.0890}  & \third{0.4408}  & \second{0.0833} & \third{82.54\%}   & \third{95.67\%}   & \third{98.49\%}   \\ \hline
\end{tabular}
\end{table*}
\begin{table*}[!htbp]
\centering
\caption{Results on the Stanford2D3D and Kujiale datasets. Bold red, orange and yellow denote the best, second-best and third-best performance respectively, while the arrows next to each metric show the direction of better performance.}
\label{tab:s2d3d_kuj_sota}
\begin{tabular}{lcccccccc}
\hline
 &
   &
  AbsRel $\downarrow$ &
  SqRel $\downarrow$ &
  RMSE $\downarrow$ &
  RMSLE $\downarrow$ &
  $\delta^1 \uparrow$ &
  $\delta^2 \uparrow$ &
  $\delta^3 \uparrow$ \\ \hline
\multicolumn{1}{c}{\multirow{4}{*}{Stanford2D3D}} &
  ELD (Ours) &
  \textbf{\first{0.1077}} &
  \second{0.0600} &
  \second{0.3923} &
  \second{0.0724} &
  \textbf{\first{87.54\%}} &
  \second{96.92\%} &
  \third{99.07\%} \\
\multicolumn{1}{c}{} &
  BiFuse \cite{wang2020bifuse} &
  \second{0.1109} &
  \textbf{\first{0.0563}} &
  \textbf{\first{0.3868}} &
  \third{0.0759} &
  \second{8616\%} &
  \third{96.65\%} &
  \second{99.19\%} \\
\multicolumn{1}{c}{} &
  \cite{zeng2020joint} &
  0.1377 &
  \third{0.0970} &
  0.4334 &
  0.0847 &
  80.53\% &
  94.04\% &
  97.67\% \\
\multicolumn{1}{c}{} &
  \cite{jin2020geometric} &
  \third{0.1180} &
  N/A &
  \third{0.4210} &
  \textbf{\first{0.0530}} &
  \third{85.10\%} &
  \textbf{\first{97.20\%}} &
  \textbf{\first{99.30\%}} \\ \hline
\multirow{3}{*}{Kujiale} &
  ELD (Ours) &
  \textbf{\first{0.0833}} &
  \textbf{\first{0.0403}} &
  \textbf{\first{0.2085}} &
  \second{0.0577} &
  \textbf{\first{93.98\%}} &
  \textbf{\first{98.05\%}} &
  \second{98.96\%} \\
 &
  BiFuse \cite{wang2020bifuse} &
  \second{0.0962} &
  \second{0.0495} &
  \second{0.2223} &
  0.0720 &
  \second{92.68\%} &
  97.79\% &
  98.86\% \\
 &
  \cite{jin2020geometric} &
  0.1030 &
  N/A &
  0.6660 &
  \textbf{\first{0.0410}} &
  91.20\% &
  \second{97.80\%} &
  \textbf{\first{99.00\%}} \\ \hline
\end{tabular}
\end{table*}

\begin{figure*}[!htbp]
\begin{tabular}{ccccc}
 \includegraphics[width=0.19\linewidth]{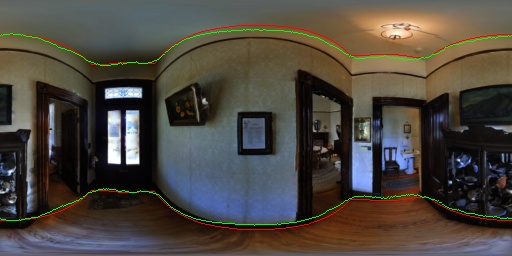} & \includegraphics[width=0.19\linewidth]{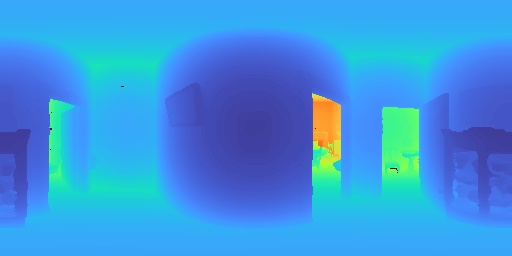}  & \includegraphics[width=0.19\linewidth]{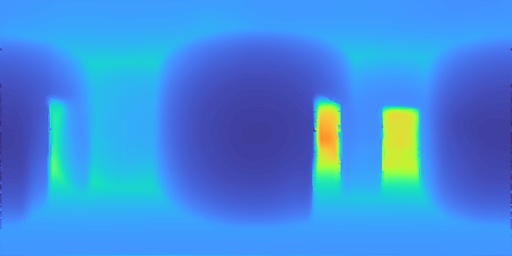} & \includegraphics[width=0.19\linewidth]{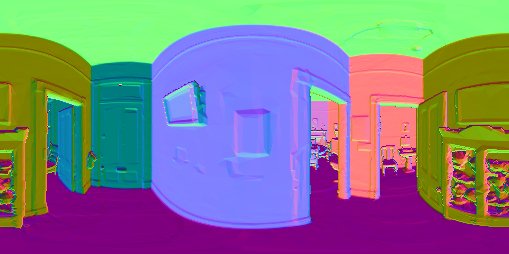} & \includegraphics[width=0.19\linewidth]{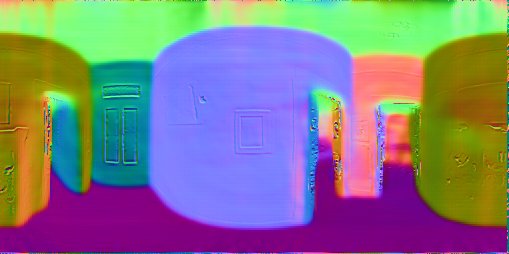} \\
 \includegraphics[width=0.19\linewidth]{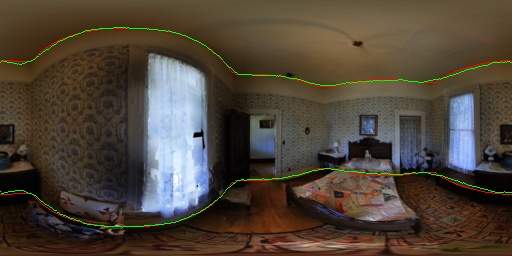} & \includegraphics[width=0.19\linewidth]{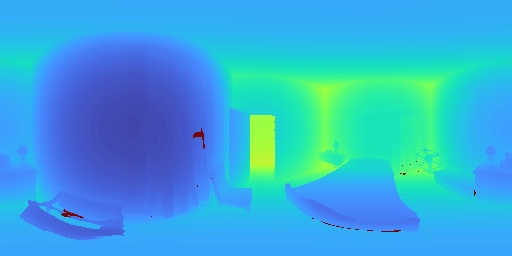} & \includegraphics[width=0.19\linewidth]{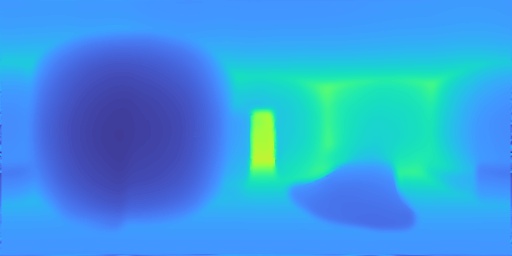} & \includegraphics[width=0.19\linewidth]{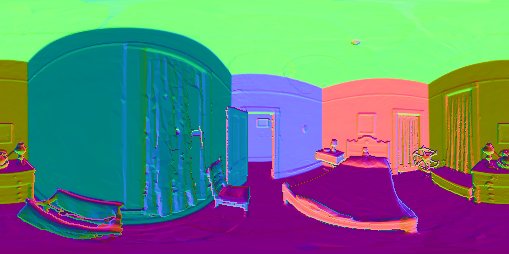} & \includegraphics[width=0.19\linewidth]{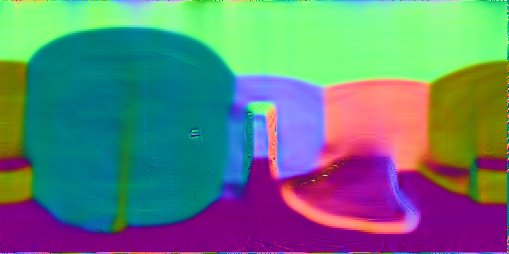} \\
\includegraphics[width=0.19\linewidth]{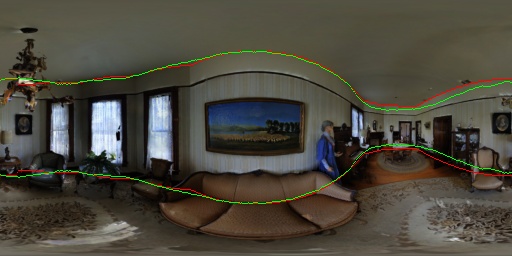} & \includegraphics[width=0.19\linewidth]{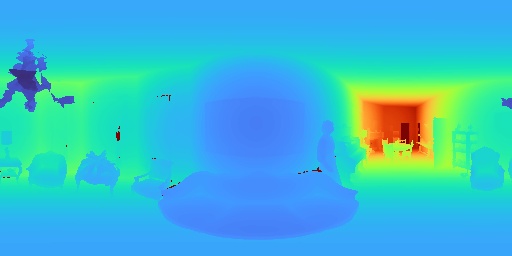} & \includegraphics[width=0.19\linewidth]{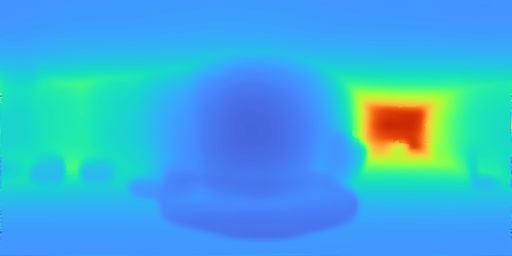} & \includegraphics[width=0.19\linewidth]{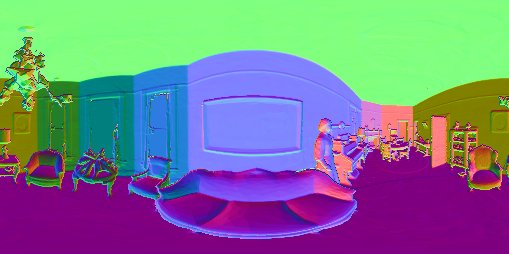} & \includegraphics[width=0.19\linewidth]{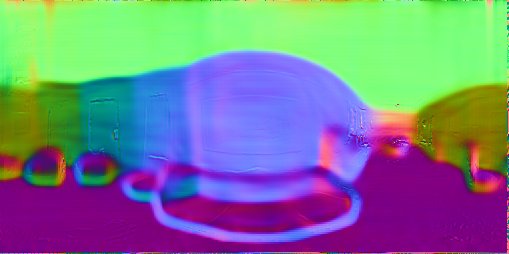} \\
\includegraphics[width=0.19\linewidth]{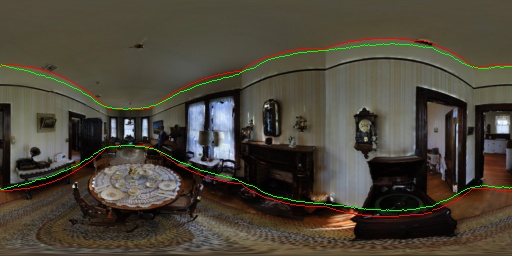} & \includegraphics[width=0.19\linewidth]{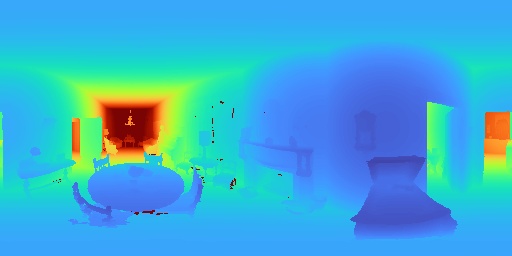} & \includegraphics[width=0.19\linewidth]{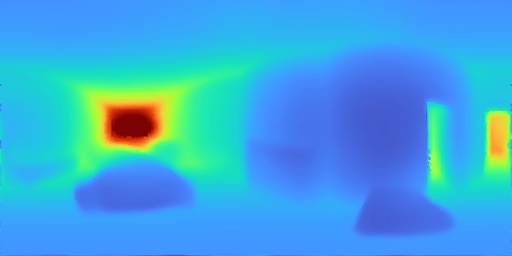} & \includegraphics[width=0.19\linewidth]{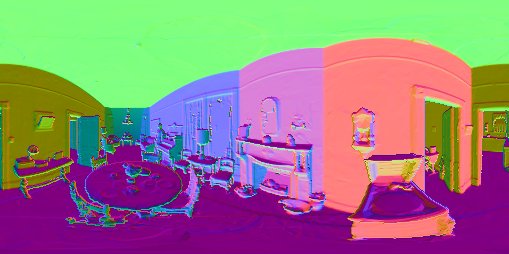} & \includegraphics[width=0.19\linewidth]{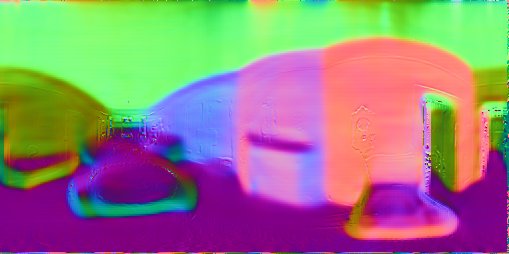} \\
\includegraphics[width=0.19\linewidth]{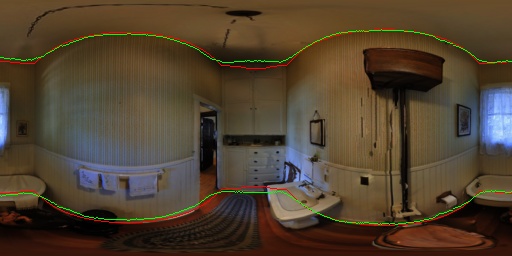} & \includegraphics[width=0.19\linewidth]{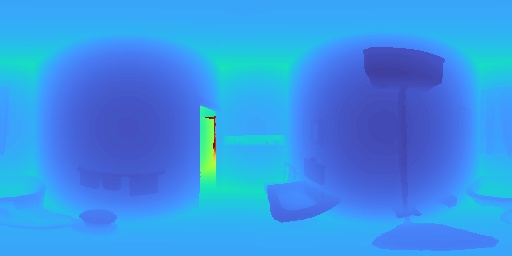} & \includegraphics[width=0.19\linewidth]{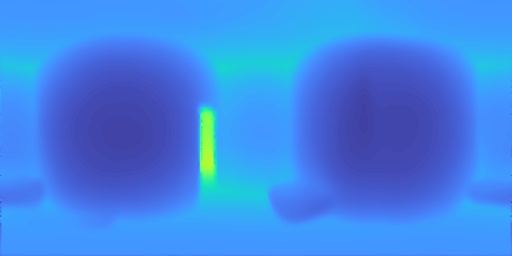} & \includegraphics[width=0.19\linewidth]{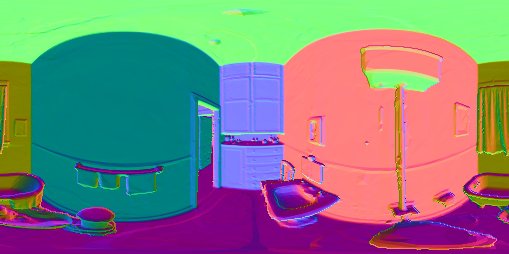} & \includegraphics[width=0.19\linewidth]{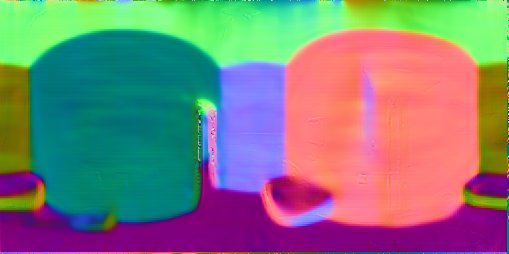} \\
\includegraphics[width=0.19\linewidth]{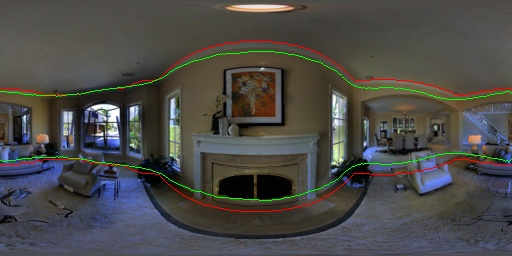} & \includegraphics[width=0.19\linewidth]{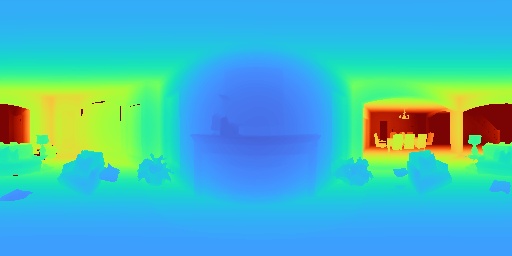} & \includegraphics[width=0.19\linewidth]{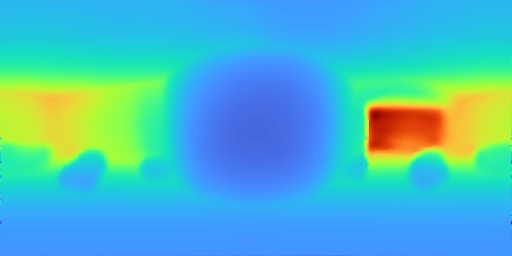} & \includegraphics[width=0.19\linewidth]{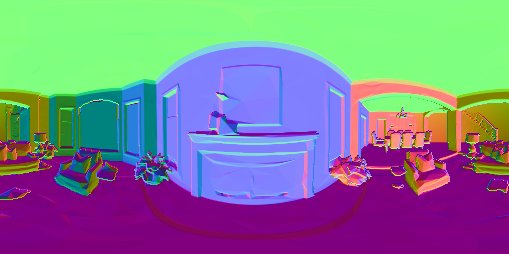} & \includegraphics[width=0.19\linewidth]{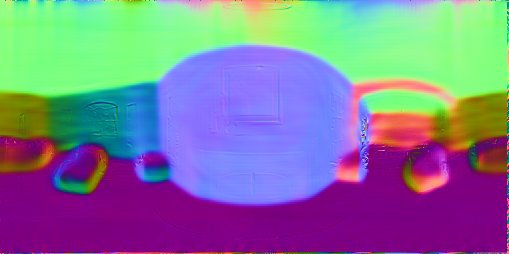} \\
\includegraphics[width=0.19\linewidth]{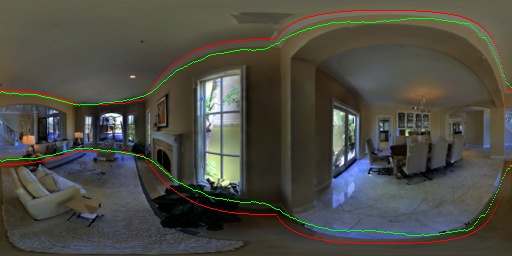} & \includegraphics[width=0.19\linewidth]{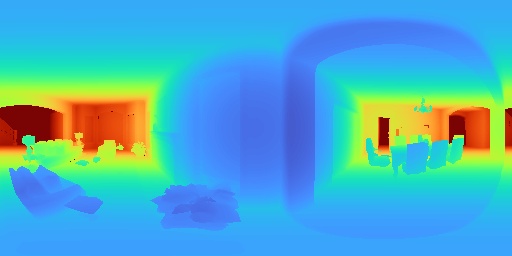} & \includegraphics[width=0.19\linewidth]{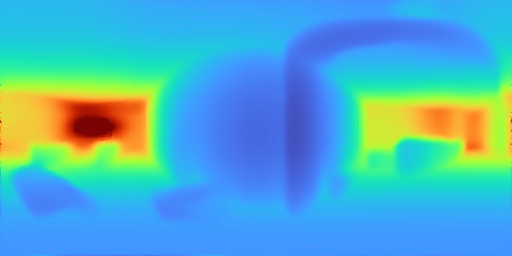} & \includegraphics[width=0.19\linewidth]{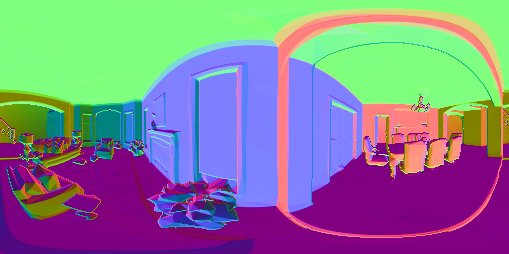} & \includegraphics[width=0.19\linewidth]{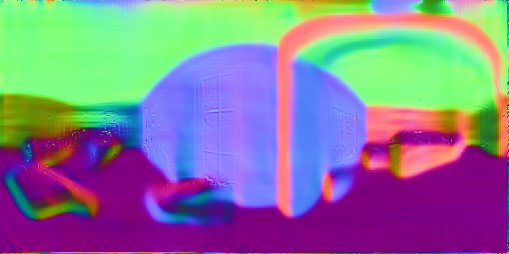} \\
\includegraphics[width=0.19\linewidth]{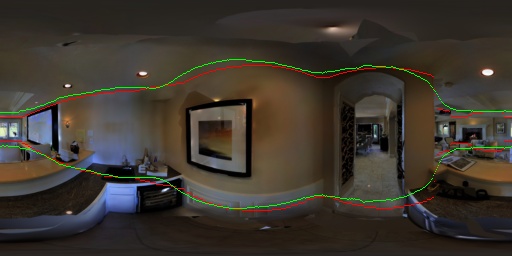} & \includegraphics[width=0.19\linewidth]{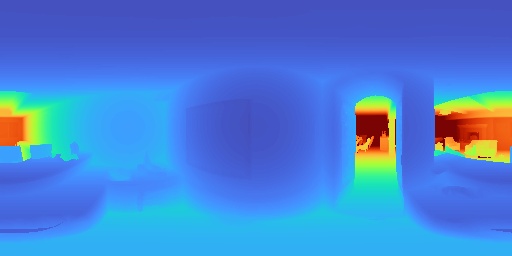} & \includegraphics[width=0.19\linewidth]{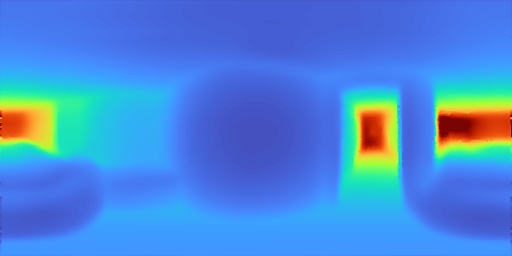} & \includegraphics[width=0.19\linewidth]{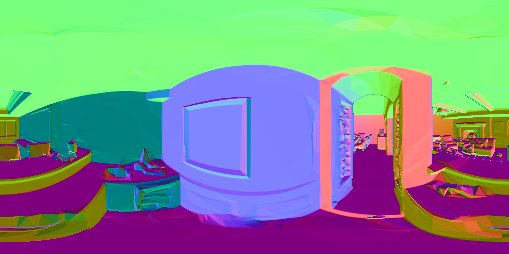} & \includegraphics[width=0.19\linewidth]{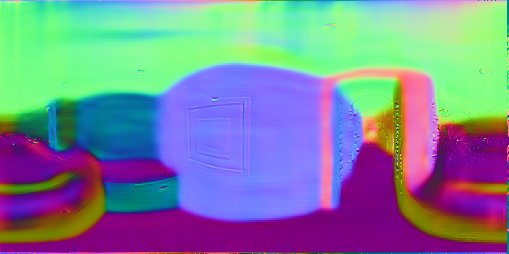} \\
\includegraphics[width=0.19\linewidth]{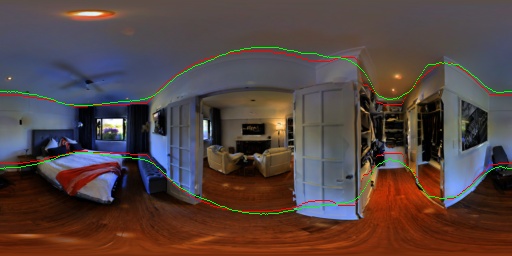} & \includegraphics[width=0.19\linewidth]{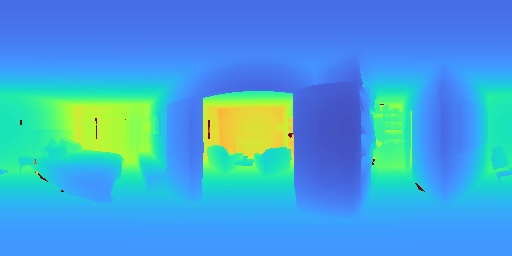} & \includegraphics[width=0.19\linewidth]{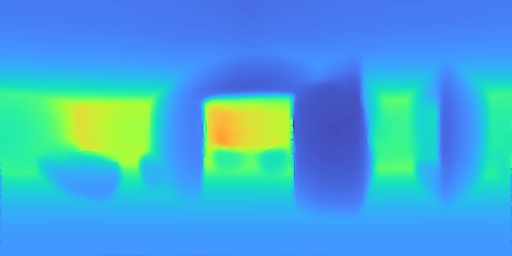} & \includegraphics[width=0.19\linewidth]{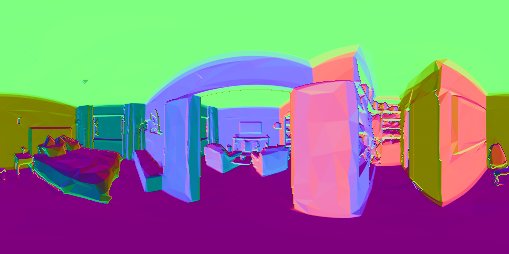} & \includegraphics[width=0.19\linewidth]{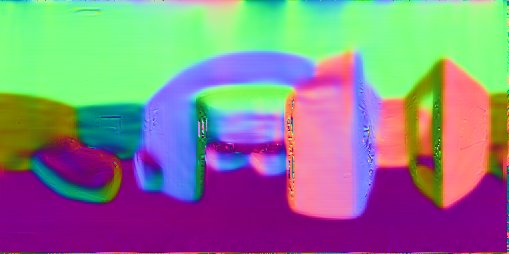} \\
\includegraphics[width=0.19\linewidth]{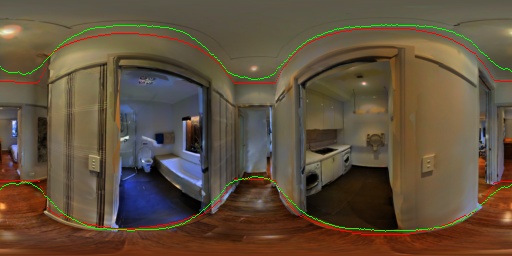} & \includegraphics[width=0.19\linewidth]{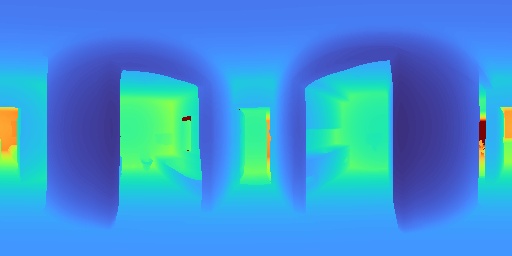} & \includegraphics[width=0.19\linewidth]{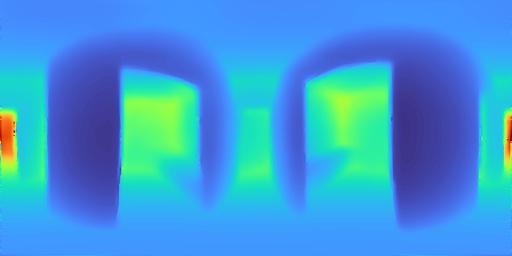} & \includegraphics[width=0.19\linewidth]{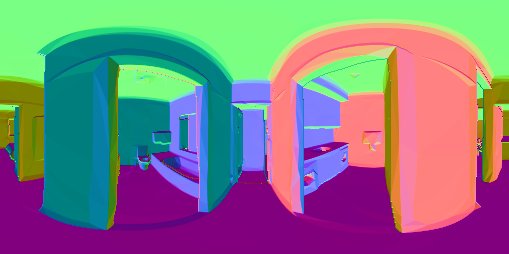} & \includegraphics[width=0.19\linewidth]{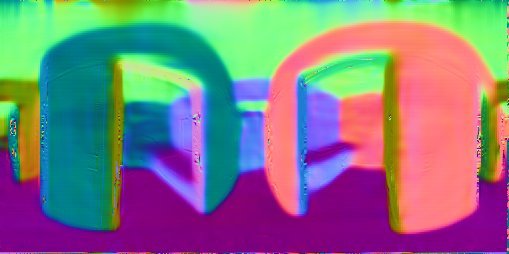} \\
\includegraphics[width=0.19\linewidth]{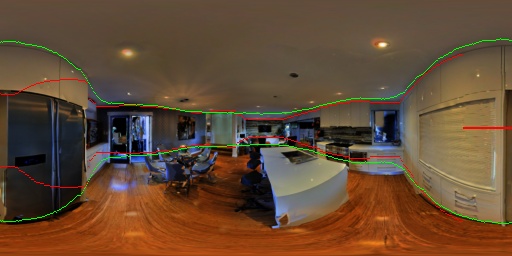} & \includegraphics[width=0.19\linewidth]{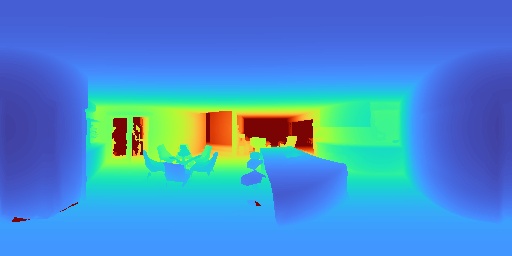} & \includegraphics[width=0.19\linewidth]{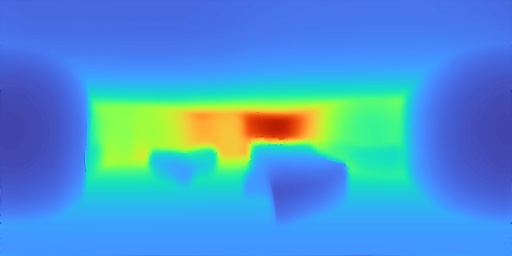} & \includegraphics[width=0.19\linewidth]{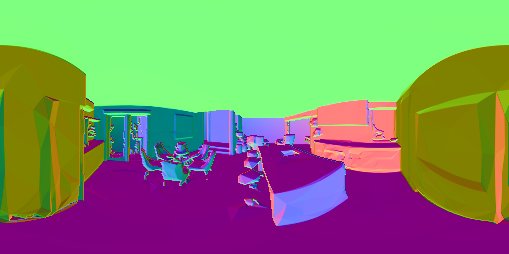} & \includegraphics[width=0.19\linewidth]{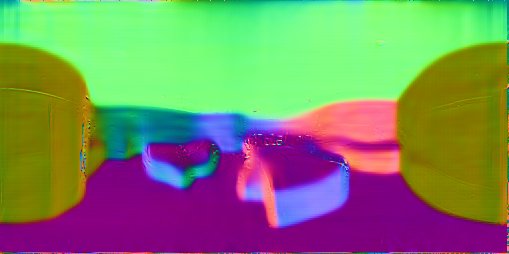}
\end{tabular}
\caption{
        Qualitative results on the Matterport3D test set. From left to right: color image with ground truth (\textcolor{red}{red}) and predicted (\textcolor{green}{green}) layout cues, ground truth depth map, predicted depth map, ground truth normal map, predicted normal map.
    }
\label{fig:m3d}
\end{figure*}

\begin{figure}[!htbp]
    \centering
      \captionsetup[subfigure]{position=top,labelformat=empty}
  
    \begin{subfigure}[b]{0.3\linewidth}\includegraphics[width=\hsize]{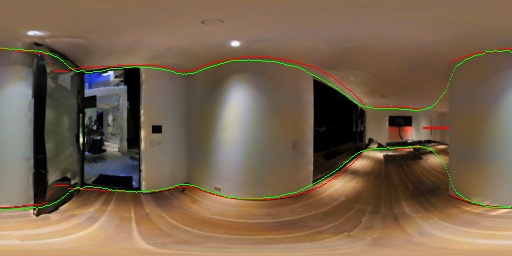}\end{subfigure}\begin{subfigure}[b]{0.3\linewidth}\includegraphics[width=\hsize]{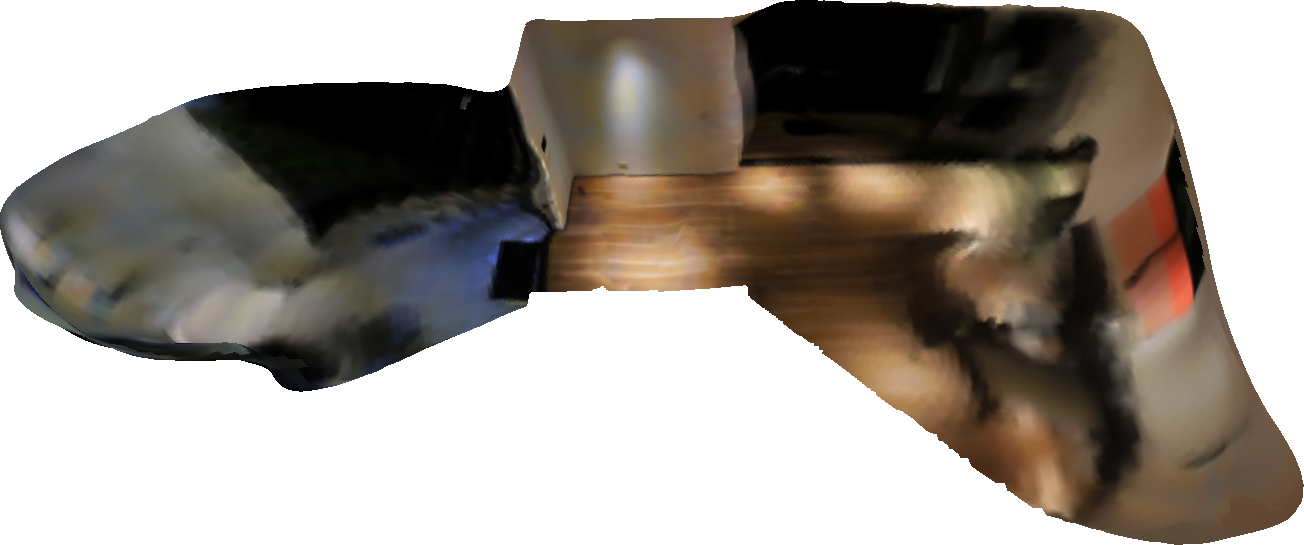}\end{subfigure}\begin{subfigure}[b]{0.3\linewidth}\includegraphics[width=\hsize]{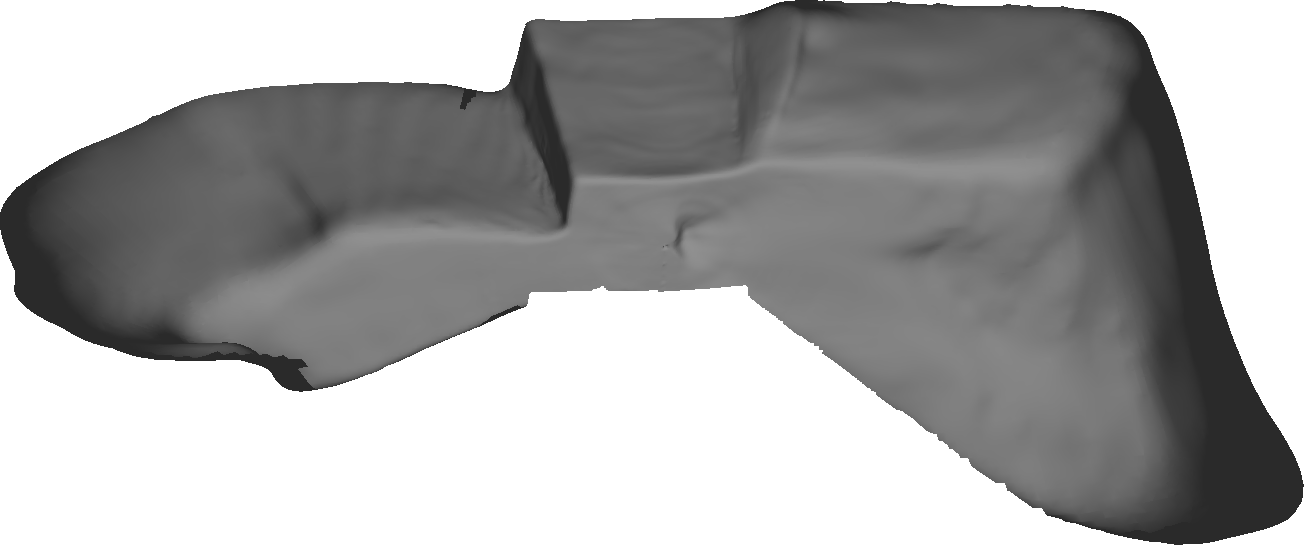}\end{subfigure}
    
    \begin{subfigure}[b]{0.3\linewidth}\includegraphics[width=\hsize]{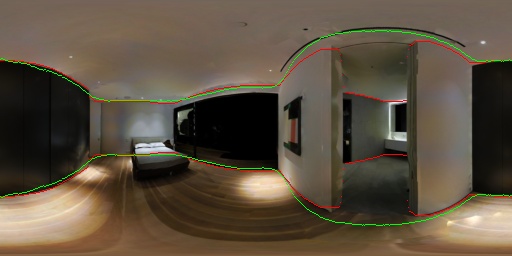}\end{subfigure}
    \begin{subfigure}[b]{0.3\linewidth}\includegraphics[width=\hsize]{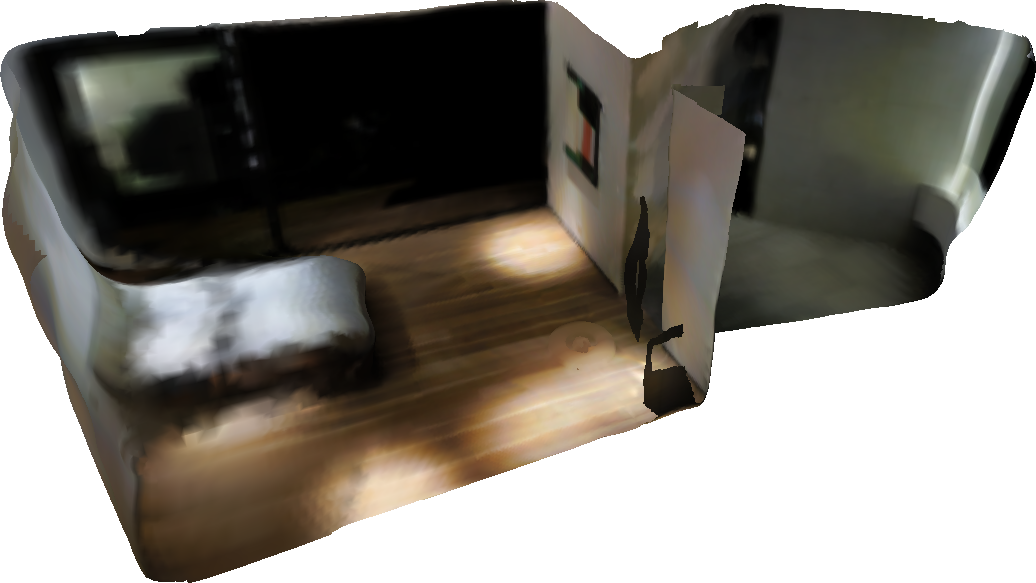}\end{subfigure}
    \begin{subfigure}[b]{0.3\linewidth}\includegraphics[width=\hsize]{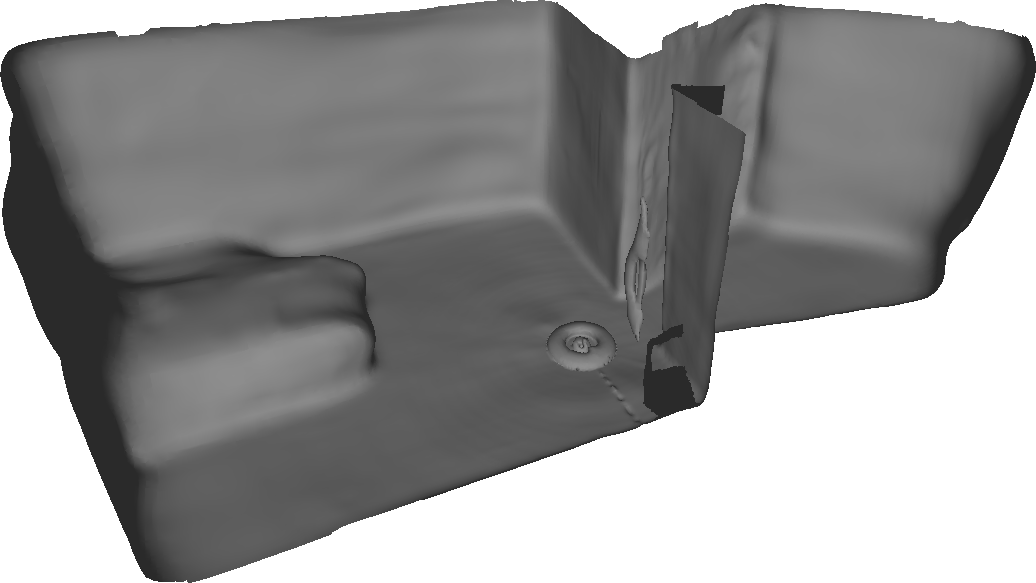}\end{subfigure}
    
    \begin{subfigure}[b]{0.3\linewidth}\includegraphics[width=\hsize]{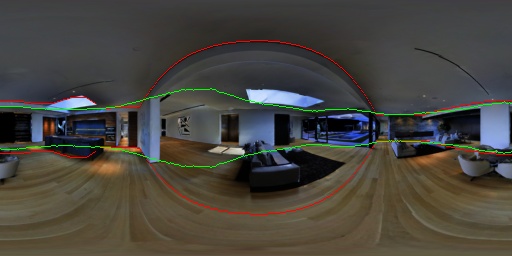}\end{subfigure}
    \begin{subfigure}[b]{0.3\linewidth}\includegraphics[width=\hsize,height=0.7\hsize]{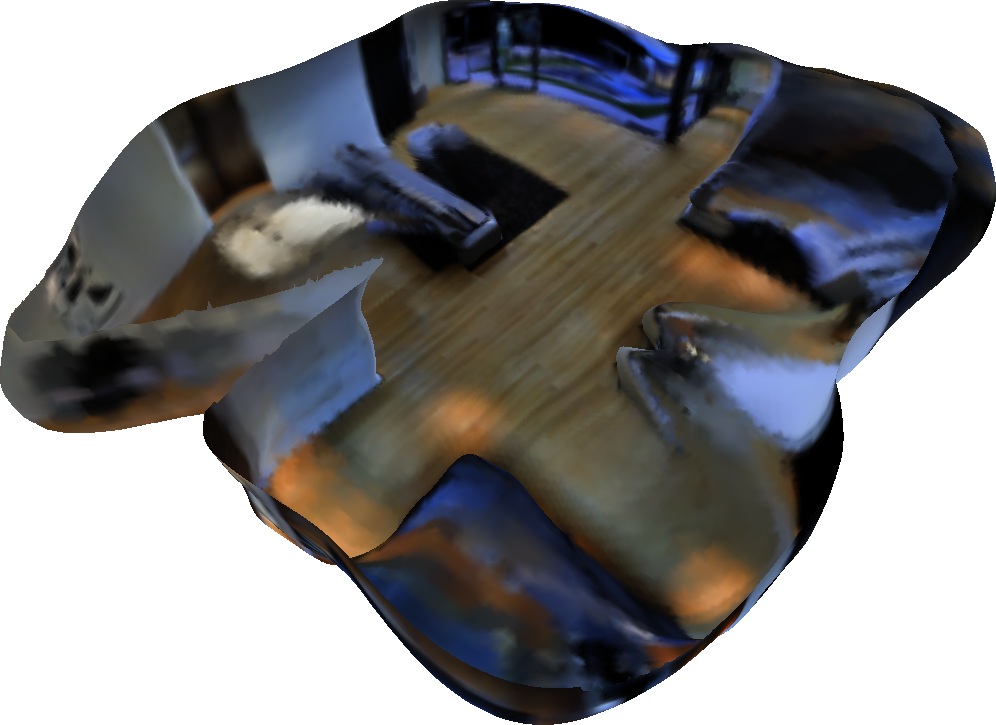}\end{subfigure}
    \begin{subfigure}[b]{0.3\linewidth}\includegraphics[width=\hsize,height=0.7\hsize]{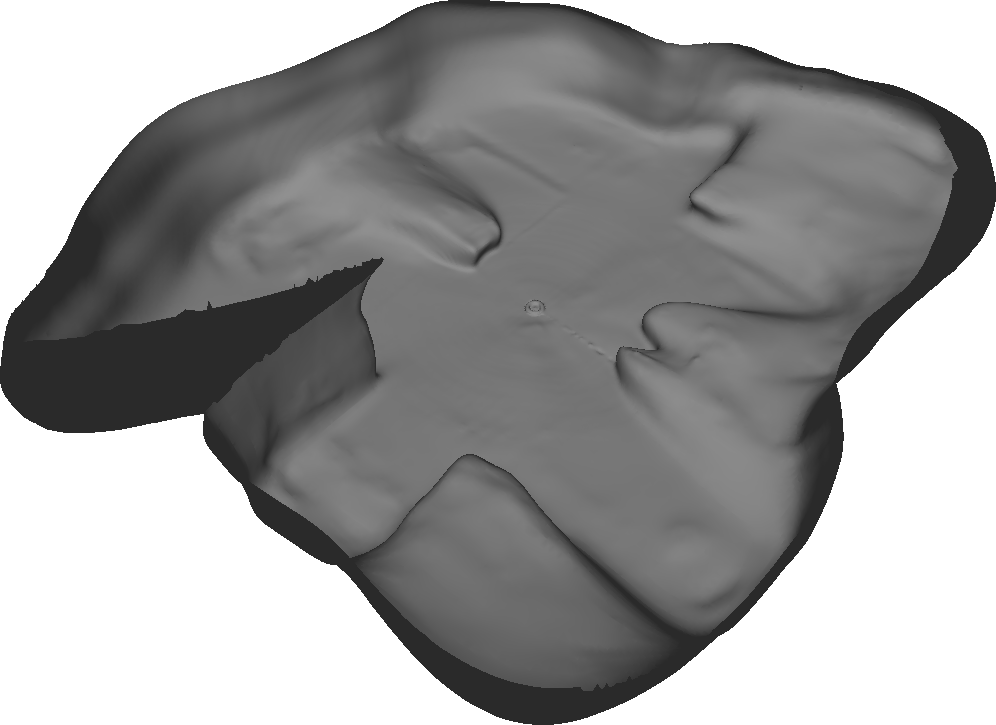}\end{subfigure}
    
    \begin{subfigure}[t]{0.3\linewidth}\includegraphics[width=\hsize]{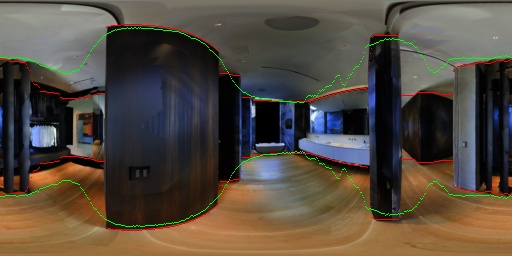}\end{subfigure}
    \begin{subfigure}[t]{0.3\linewidth}\includegraphics[angle=90,width=\hsize,height=0.7\hsize]{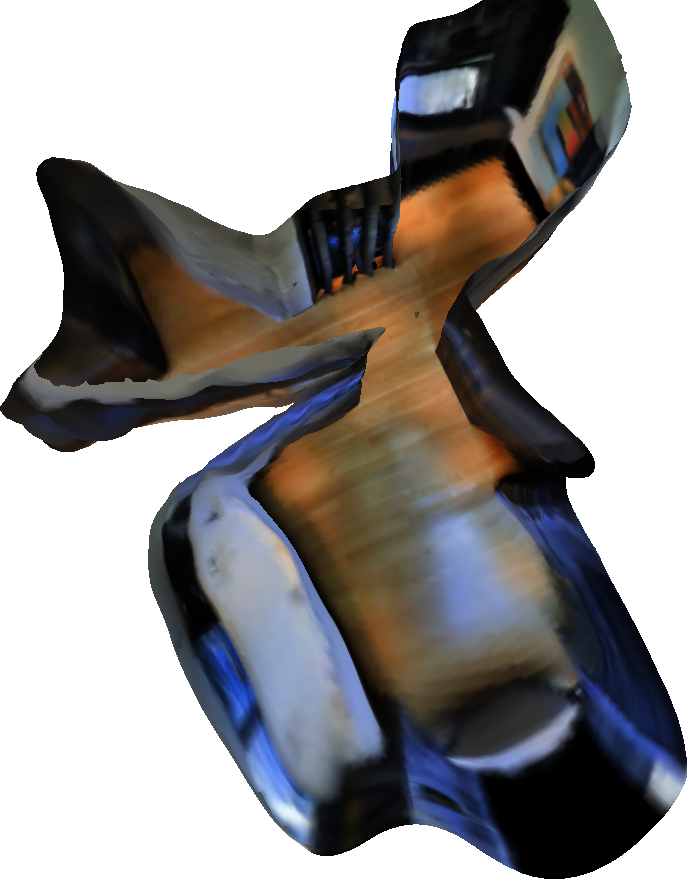}\end{subfigure}
    \begin{subfigure}[t]{0.3\linewidth}\includegraphics[angle=90,width=\hsize,height=0.7\hsize]{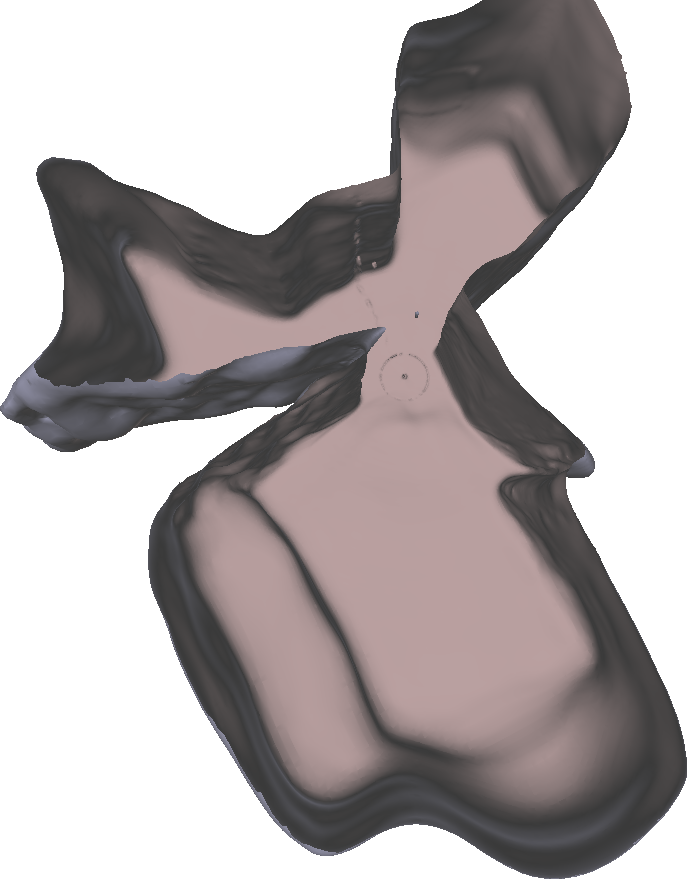}\end{subfigure}

    \caption{
        Qualitative results on the Matterport3D test set showcasing Screened Poisson 3D surface reconstructions \cite{kazhdan2013screened} of the resulting point clouds.
    }
\label{fig:m3d_3d}
\end{figure}

We split our dataset in two, taking into account the fact that it consists of both Matterport3D and Stanford2D3D.
We use the official splits from each dataset (for Stanford2D3D we use fold\#1).
For the Matterport3D part, we train for $40$ epochs, while for the smaller Stanford2D3D part we train for $100$ epochs.
In addition, we also use the Kujiale dataset introduced in \cite{jin2020geometric} which, contrary to our dataset, offers high quality layout annotations in the form of the scene's junctions locations.
To integrate them into our learning framework, we reconstruct the boundary from the corners, with the major difference being the quality of the boundaries compared to our automatically generated weak boundaries.
We train for $60$ epochs on Kujiale.

Table~\ref{tab:m3d_sota} presents the results of our explicitly connected layout and depth (ELD) approach when trained and evaluated on our dataset's Matterpor3D part, and a comparison with the two models presented in the pioneering \360 depth estimation work of \cite{zioulis2018omnidepth}, as well as the BiFuse \cite{wang2020bifuse} model, all trained on our dataset.
As also presented in \cite{zioulis2018omnidepth} the RectNet model outperforms the UResNet one, but our approach provides better quantitative results across all metrics, with BiFuse being closer in performance to our model.
Table~\ref{tab:s2d3d_kuj_sota} presents results on the Stanford2D3D and Kujiale datasets and a comparison to BiFuse, and \cite{jin2020geometric} which also leverages layout estimation, but as a regularizer and prior, when learning to estimate depth from a single color panorama.
Further, for Stanford2D3D we also provide a comparison against \cite{zeng2020joint}, which is another work using implicit layout integration into a depth estimation model.
Given the missing hyperparameter information, we use $\lambda=\mu=0.5$, a batch size of $2$, and the same learning rate as our model, all of which were heuristically searched for using an empirical search space.
It should be noted that for BiFuse and \cite{zeng2020joint}, we train for the same number of epochs as our models, scaling their training regimes appropriately.
Specifically for \cite{zeng2020joint}, whose training schedule is unreported, we resort to the following heuristically searched epoch milestones: $[15, 30, 50]$, each one corresponding to a training stage transition.
It should be noted that we do not evaluate on our entire test set, but instead only on the same subset \cite{jin2020geometric} and \cite{zeng2020joint} used (the splits used in LayoutNet \cite{zou2018layoutnet}) for a fair comparison.
We observe that performance on Stanford2D3D is split between our approach, BiFuse and \cite{jin2020geometric}.
While our ELD model performs better in the RMSE and relative metrics, as well as the stricter accuracy ($\delta^1$), the implicitly connected layout and depth model offers better performance at near depths as indicated by the RMSLE metric and is less frequently very erroneous as shown by its increased performance in the $\delta^2$ and $\delta^3$ accuracies.
At the same time, BiFuse offers the better performance in farther depths as indicated by the squared metrics, whereas our model offers the better performance for the stricter relative metrics, but is also the most balanced one.
But this is mostly the case for Stanford2D3D, as the performance gap in these metrics closes considerably when using the high quality layout boundaries available in the Kujiale dataset, indicating that this relative performance deviation may be associated with the quality of the layout information.
It should also be noted that the Kujiale dataset is a synthetic one and only contains cuboid rooms.
The same applies to the Stanford2D3D real-world dataset, whose layout annotations are cuboid too, sometimes violating the observed scene, a fact that introduces as a contradiction between the estimated geometry and layout.

Figure~\ref{fig:m3d} presents a set of qualitative results on Matterport3D's test set, which comprise unseen buildings.
Apart from the predicted depth and normal maps, which showcase how the model captures each scene's main structures adequately, the predicted and annotated weak layout cues are also illustrated on the color images.
Moreover, Figure~\ref{fig:m3d_3d} shows the 3D reconstructions of a set of Matterport3D test scenes derived from the estimated depth, after converted into a point cloud.
The resulting meshes capture the structural planes efficiently and can reconstruct complex room topologies and dominant structural elements like beds, sofas and kitchen bars from a single monocular panorama image.

\subsection{Ablation Study}
\label{sec:ablation}
\begin{table*}[!htbp]
\centering
\caption{Layout ablation results on the Stanford2D3D, Kujiale and Matterport3D datasets.}
\label{tab:ablation}
\begin{tabular}{@{}lcccccccc@{}}
\toprule
 &
   &
  AbsRel $\downarrow$ &
  SqRel $\downarrow$ &
  RMSE $\downarrow$ &
  RMSLE $\downarrow$ &
  $\delta^1 \uparrow$ &
  $\delta^2 \uparrow$ &
  $\delta^3 \uparrow$ \\ \midrule
\multirow{3}{*}{Stanford2D3D} &
  ELD &
  \first{0.1077} &
  \second{0.0600} &
  \second{0.3923} &
  \first{0.0724} &
  \first{87.54\%} &
  \first{96.92\%} &
  \first{99.07\%} \\
 & w/o connection & 0.1175          & 0.0647          & 0.4134          & 0.0789          & 84.63\%          & 96.72\%         & \second{99.01\%} \\
 &
  Depth only &
  \second{0.1101} &
  \first{0.0592} &
  \first{0.3910} &
  \second{0.0736} &
  \second{86.71\%} &
  \second{96.89\%} &
  98.96\% \\ \midrule
\multirow{3}{*}{Kujiale} &
  ELD &
  \first{0.0833} &
  \first{0.0403} &
  \first{0.2085} &
  \first{0.0577} &
  \first{93.98\%} &
  \second{98.05\%} &
  \second{98.96\%} \\
 & w/o connection & 0.0882          & 0.0461          & \second{0.2105} & \second{0.0580} & 93.41\%          & \first{98.12\%} & \first{99.00\%}  \\
 & Depth only     & \second{0.0856} & \second{0.0406} & 0.2138          & 0.0977          & \second{93.46\%} & 98.00\%         & 98.93\%          \\ \midrule
\multirow{3}{*}{Matterport3D} &
  ELD &
  \first{0.1136} &
  \first{0.0707} &
  \first{0.4066} &
  \first{0.0743} &
  \first{88.46\%} &
  \first{97.10\%} &
  \first{98.91\%} \\
 &
  w/o connection &
  \second{0.1185} &
  \second{0.0751} &
  0.4166 &
  \second{0.0752} &
  \second{87.39\%} &
  \second{96.89\%} &
  \second{98.88\%} \\
 & Depth only     & 0.1227          & 0.0766          & \second{0.4110} & 0.0780          & 86.47\%          & 96.78\%         & \second{98.88\%} \\ \bottomrule
\end{tabular}
\end{table*}
\begin{table*}[!htbp]
\centering
\caption{Loss ablation results on the Stanford2D3D and Kujiale datasets. Bold red and orange denote the best and second-best performance respectively, while the arrows next to each metric show the direction of better performance.}
\label{tab:loss_ablation}
\begin{tabular}{lcccccccc}
\hline
 &
   &
  AbsRel $\downarrow$ &
  SqRel $\downarrow$ &
  RMSE $\downarrow$ &
  RMSLE $\downarrow$ &
  $\delta^1 \uparrow$ &
  $\delta^2 \uparrow$ &
  $\delta^3 \uparrow$ \\ \hline
\multicolumn{1}{c}{\multirow{3}{*}{Stanford2D3D}} &
  ELD (Ours) &
  \textbf{\first{0.1077}} &
  \textbf{\first{0.0600}} &
  \textbf{\first{0.3923}} &
  \textbf{\first{0.0724}} &
  \textbf{\first{87.54\%}} &
  \textbf{\first{96.92\%}} &
  \second{99.07\%} \\
\multicolumn{1}{c}{} &
  w/o $\mathcal{L}_S$ &
  \second{0.1113} &
  \textbf{\first{0.0600}} &
  \second{0.3941} &
  \second{0.0747} &
  \second{85.92\%} &
  96.69\% &
  \textbf{\first{99.11\%}} \\
\multicolumn{1}{c}{} &
  w/o $\mathcal{L}_V$ &
  0.1173 &
  \second{0.0628} &
  0.3997 &
  0.0761 &
  85.59\% &
  \second{96.74\%} &
  \second{99.07\%} \\ \hline
\multirow{3}{*}{Kujiale} &
  ELD (Ours) &
  \textbf{\first{0.0833}} &
  \textbf{\first{0.0403}} &
  \textbf{\first{0.2085}} &
  \textbf{\first{0.0577}} &
  \textbf{\first{93.98\%}} &
  \textbf{\first{98.05\%}} &
  \textbf{\first{98.96\%}} \\
 &
  w/o $\mathcal{L}_S$ &
  \second{0.0916} &
  0.0705 &
  \second{0.2308} &
  0.0805 &
  \second{92.97\%} &
  \second{97.89\%} &
  \second{98.93\%} \\
 &
  w/o $\mathcal{L}_V$ &
  0.1114 &
  \second{0.0572} &
  0.2390 &
  \second{0.0739} &
  90.51\% &
  96.95\% &
  98.74\% \\ \hline
\end{tabular}
\end{table*}

We additionally perform an ablation study of the explicit layout and depth connections as presented in Section~\ref{sec:explicit}, comparing it to an implicit connection and a baseline pure depth estimation model.
For the former, we adapt the model to also predict the bottom layout, instead of reconstructing it using the predicted depth, and we remove the layout-based attention mechanism, while for the latter we only keep the depth estimating decoder.

Table~\ref{tab:ablation} presents the results on the Matterport3D and Stanford2D3D parts of our dataset with weak layout annotations, and on the Kujiale dataset with high quality layout annotations.
Overall, we find that ELD boosts performance compared to the simpler baselines.
Interestingly, the disconnected dual task model does not always result in performance improvements over the baseline depth.
While there are no high level implicit interactions like those used in \cite{jin2020geometric}, it would be expected that due to the task complementarity, the two tasks would benefit each other when jointly learned with a shared encoder.
To investigate this further, in Table~\ref{tab:layout} we present results for the layout boundary indicator of Eq.\eqref{eq:layout_indicator} compared to a baseline model with only the layout decoder (adapted to predict both boundaries) and our model trained without any connection between the two decoding branches (neither attention, nor reconstruction).
The best performing models are selected based on the depth performance indicator of Eq.\eqref{eq:depth_indicator}.
The explicit connection model (ELD) not only offers better performance for depth estimation, but also better approximates the layout boundary.
Furthermore, simple two branch joint training does not necessarily help both tasks as indicated by both these results, and those presented for the depth estimation performance.
Our findings hint that such explicit connections help the model reach a consensus in both tasks simultaneously, compared to disjoint training that leads to a changing task bias during training.

Finally, we also ablate our loss functions by removing the surface and virtual normal losses and training the corresponding models on both Stanford2D3D and Kujiale.
The results are presented in Table~\ref{tab:loss_ablation}, which shows that both terms contribute to increased performance, with the surface term offering a higher boost to the relative metrics.

\begin{table}[!htbp]
\centering
\caption{Layout performance indicators for the ablated models across the different datasets.}
\label{tab:layout}
\begin{tabular}{@{}lccc@{}}
\toprule
               & $I_L \downarrow$ & $I_L \downarrow$ & $I_L \downarrow$ \\ \midrule
               & Stanford2D3D     & Kujiale          & Matterport3D     \\
ELD            & \first{8.94}     & \first{12.78}    & \first{15.52}    \\
w/o connection & 11.54            & \second{12.94}   & \second{15.87}   \\
Layout only    & \second{10.05}   & 13.92            & 17.03            \\ \bottomrule
\end{tabular}
\end{table}

\subsection{Data Study}
\label{sec:data_study}
\begin{table*}[!htbp]
\centering
\caption{Results on the Matterport3D dataset using our rendered 360V data and the original Matterport3D perspective data stitched into panoramas, evaluated only on the valid stitched regions.}
\label{tab:m3d_stitched}
\begin{tabular}{@{}lcccccccc@{}}
\toprule
 &  & AbsRel $\downarrow$ & SqRel $\downarrow$ & RMSE $\downarrow$ & RMSLE $\downarrow$ & $\delta^1 \uparrow$ & $\delta^2 \uparrow$ & $\delta^3 \uparrow$ \\ \midrule
\multirow{2}{*}{Matterport3D (360V)}     & ELD (Depth Only)             & 0.1274 & 0.0905 & 0.4629 & 0.0826 & 85.99\% & 96.44\% & 98.67\% \\
                                         & BiFuse \cite{wang2020bifuse} & 0.1289 & 0.1031 & 0.4744 & 0.0787 & 85.93\% & 96.53\% & 98.66\% \\ \midrule
\multirow{2}{*}{Matterport3D (stitched)} & ELD (Depth Only)             & 0.1290 & 0.1084 & 0.5176 & 0.0841 & 84.90\% & 95.17\% & 97.98\% \\
                                         & BiFuse \cite{wang2020bifuse} & 0.1320 & 0.1061 & 0.5154 & 0.0836 & 83.89\% & 95.05\% & 98.04\% \\ \bottomrule
\end{tabular}
\end{table*}
\begin{figure*}[!htbp]
\captionsetup[subfigure]{position=top,labelformat=empty}
    \centering
    \begin{subfigure}[b]{0.19\linewidth}\includegraphics[width=\hsize]{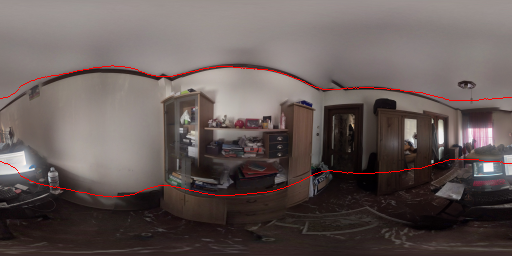}\end{subfigure}
    \hfill
    \begin{subfigure}[b]{0.19\linewidth}\includegraphics[width=\hsize]{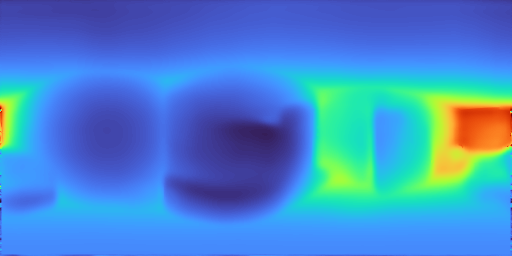}\end{subfigure}
    \hfill
    \hfill
    \begin{subfigure}[b]{0.19\linewidth}\includegraphics[width=\hsize]{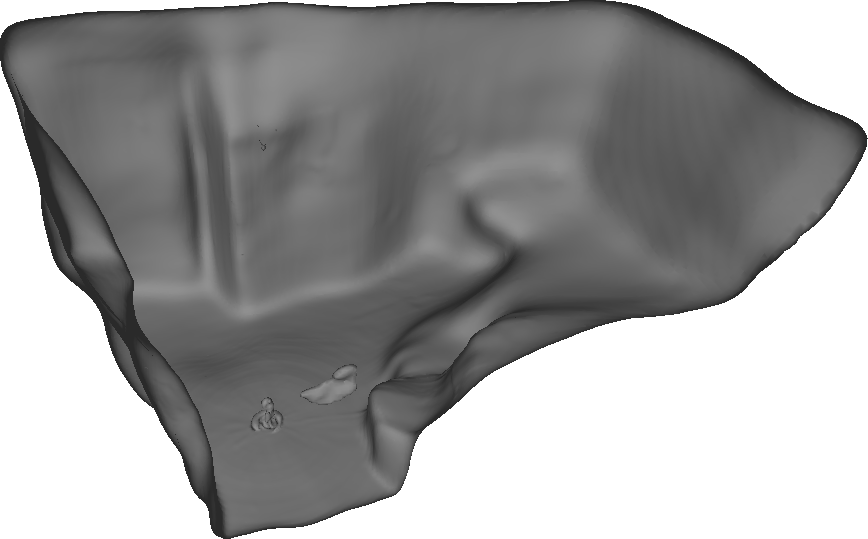}\end{subfigure}
    \hfill
    \begin{subfigure}[b]{0.19\linewidth}\includegraphics[width=\hsize]{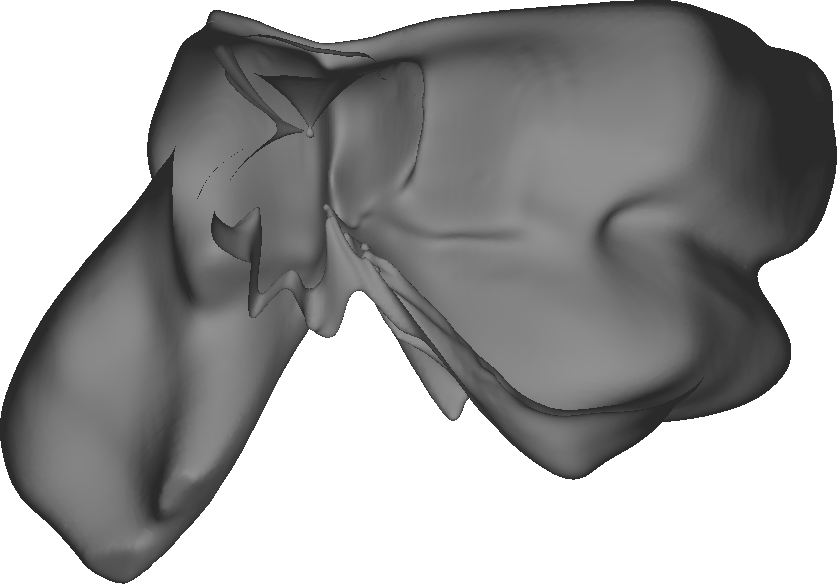}\end{subfigure}
    \hfill
    \begin{subfigure}[b]{0.19\linewidth}\includegraphics[width=\hsize]{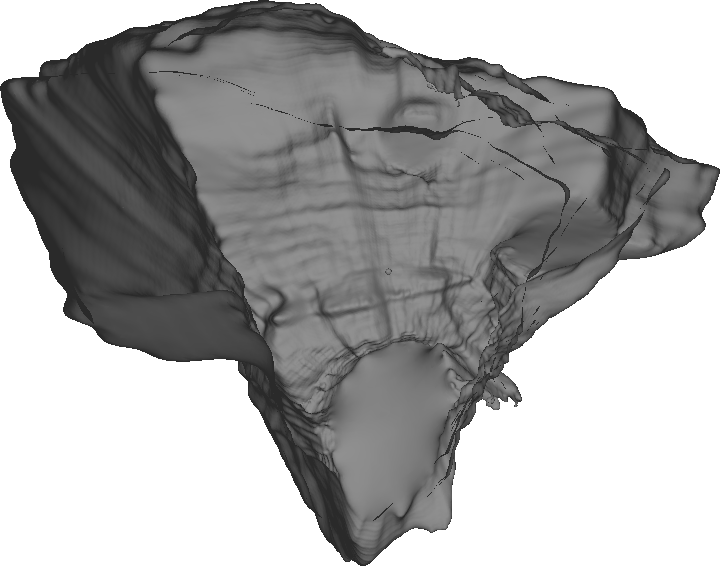}\end{subfigure}
    \hfill

    \begin{subfigure}[b]{0.19\linewidth}\includegraphics[width=\hsize]{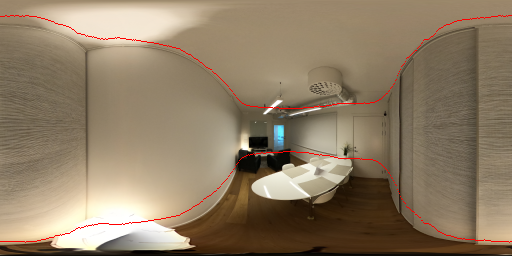}\end{subfigure}
    \hfill
    \begin{subfigure}[b]{0.19\linewidth}\includegraphics[width=\hsize]{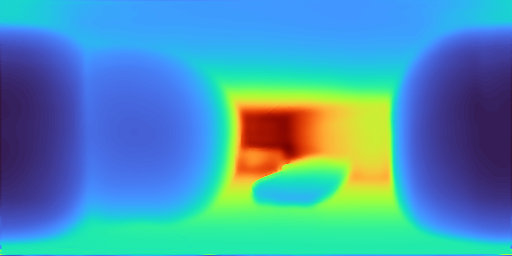}\end{subfigure}
    \hfill
    \hfill
    \begin{subfigure}[b]{0.19\linewidth}\includegraphics[width=\hsize,height=0.6\hsize]{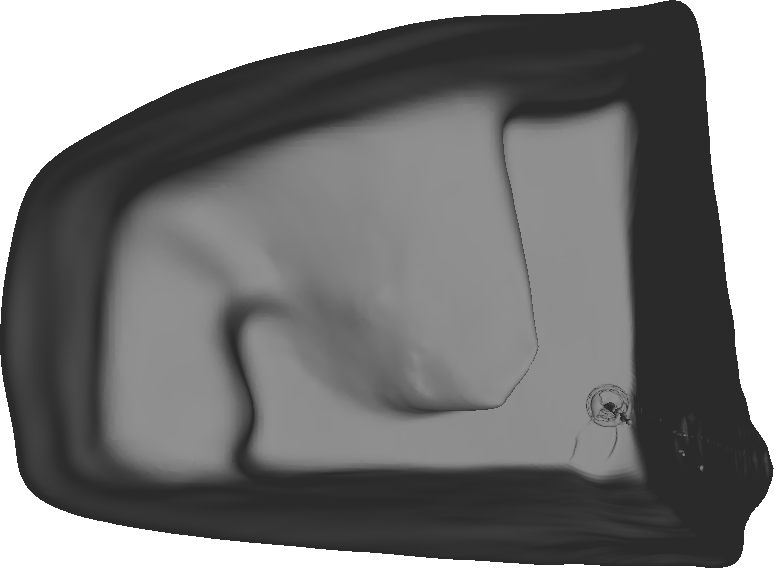}\end{subfigure}
    \hfill
    \begin{subfigure}[b]{0.19\linewidth}\includegraphics[width=\hsize,height=0.6\hsize]{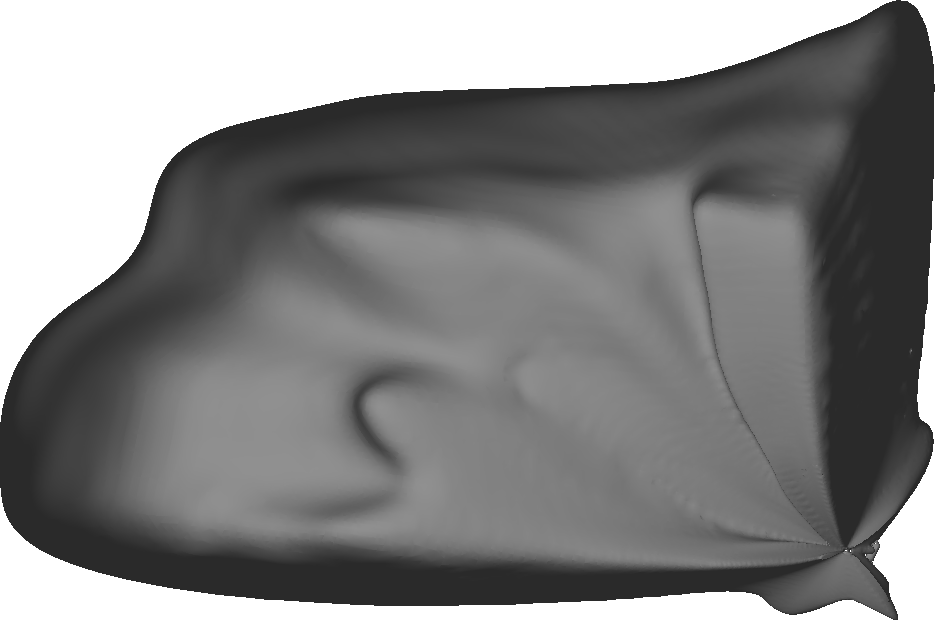}\end{subfigure}
    \hfill
    \begin{subfigure}[b]{0.19\linewidth}\includegraphics[width=\hsize,height=0.6\hsize]{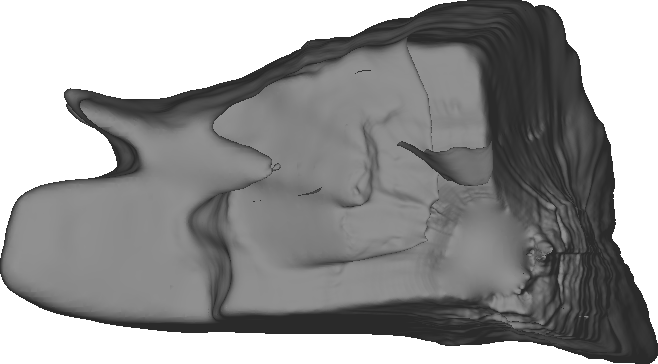}\end{subfigure}
    \hfill

    \begin{subfigure}[b]{0.19\linewidth}\includegraphics[width=\hsize]{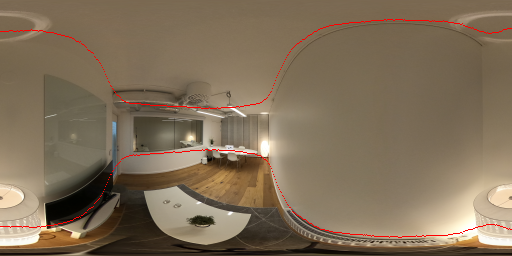}\end{subfigure}
    \hfill
    \begin{subfigure}[b]{0.19\linewidth}\includegraphics[width=\hsize]{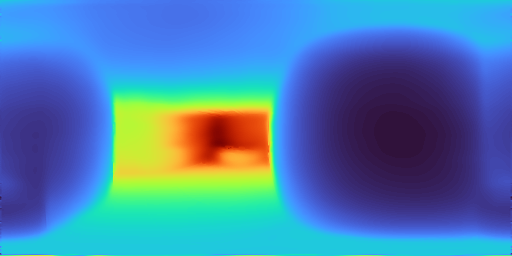}\end{subfigure}
    \hfill
    \hfill
    \begin{subfigure}[b]{0.19\linewidth}\includegraphics[width=\hsize,height=0.6\hsize]{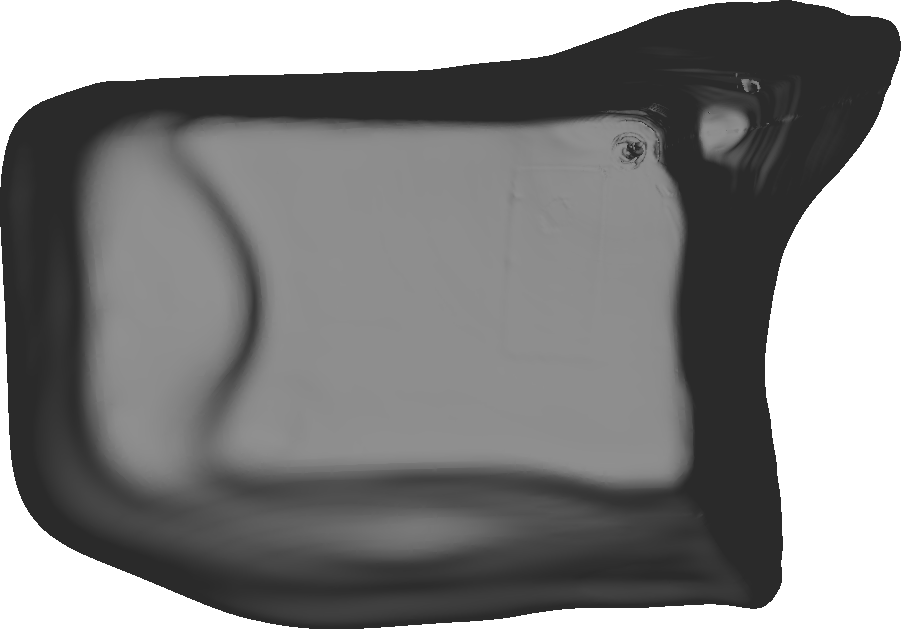}\end{subfigure}
    \hfill
    \begin{subfigure}[b]{0.19\linewidth}\includegraphics[width=\hsize,height=0.6\hsize]{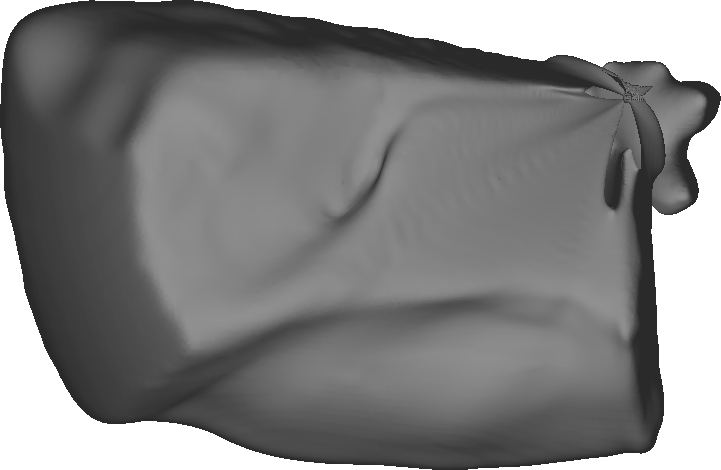}\end{subfigure}
    \hfill
    \begin{subfigure}[b]{0.19\linewidth}\includegraphics[width=\hsize,height=0.6\hsize]{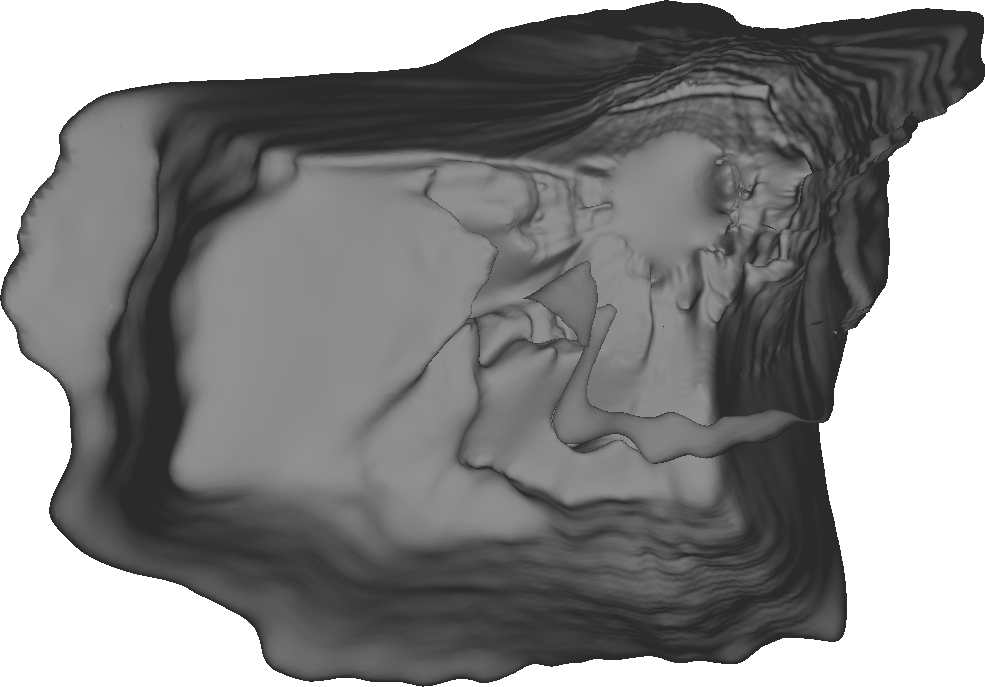}\end{subfigure}
    \hfill
    \rule[-1ex]{16cm}{0.5pt}\\

    \begin{subfigure}[b]{0.19\linewidth}\includegraphics[width=\hsize]{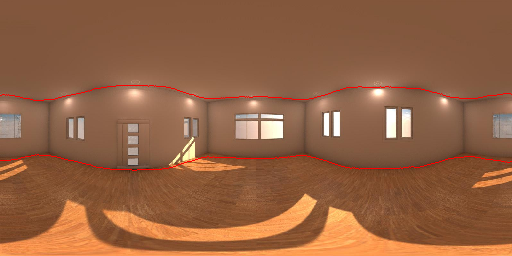}\end{subfigure}
    \hfill
    \begin{subfigure}[b]{0.19\linewidth}\includegraphics[width=\hsize]{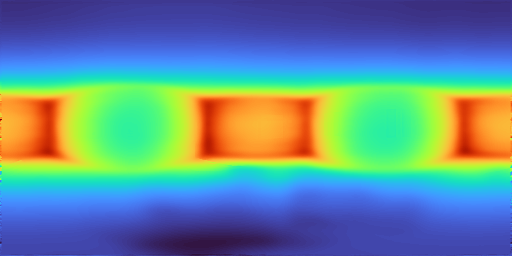}\end{subfigure}
    \hfill
    \hfill
    \begin{subfigure}[b]{0.19\linewidth}\includegraphics[width=\hsize,height=0.6\hsize]{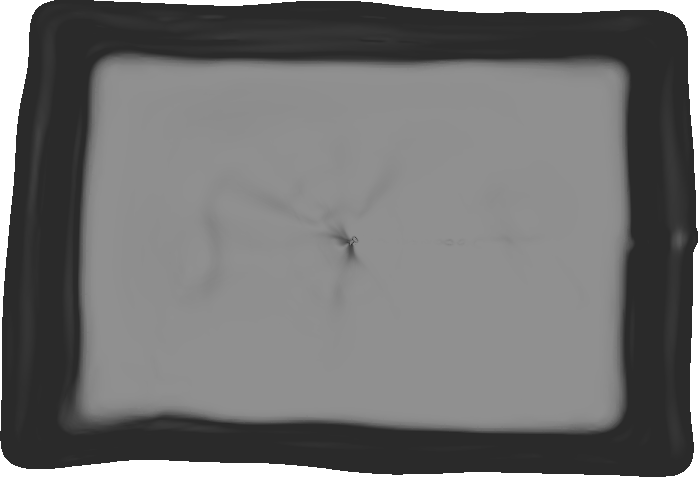}\end{subfigure}
    \hfill
    \begin{subfigure}[b]{0.19\linewidth}\includegraphics[width=\hsize,height=0.6\hsize]{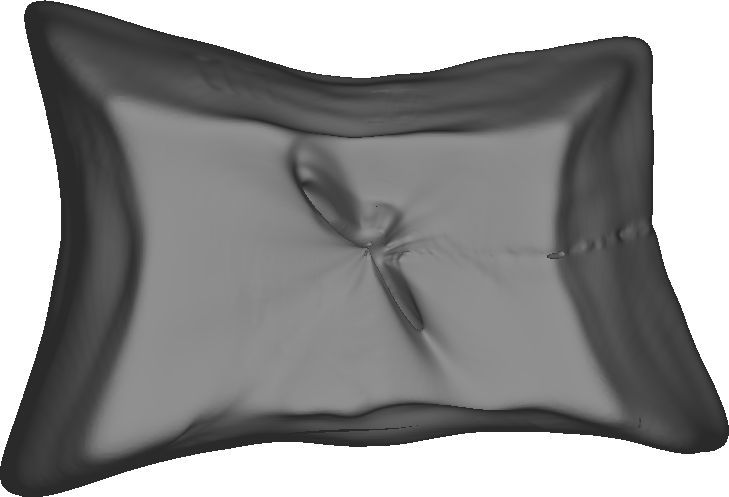}\end{subfigure}
    \hfill
    \begin{subfigure}[b]{0.19\linewidth}\includegraphics[width=\hsize,height=0.6\hsize]{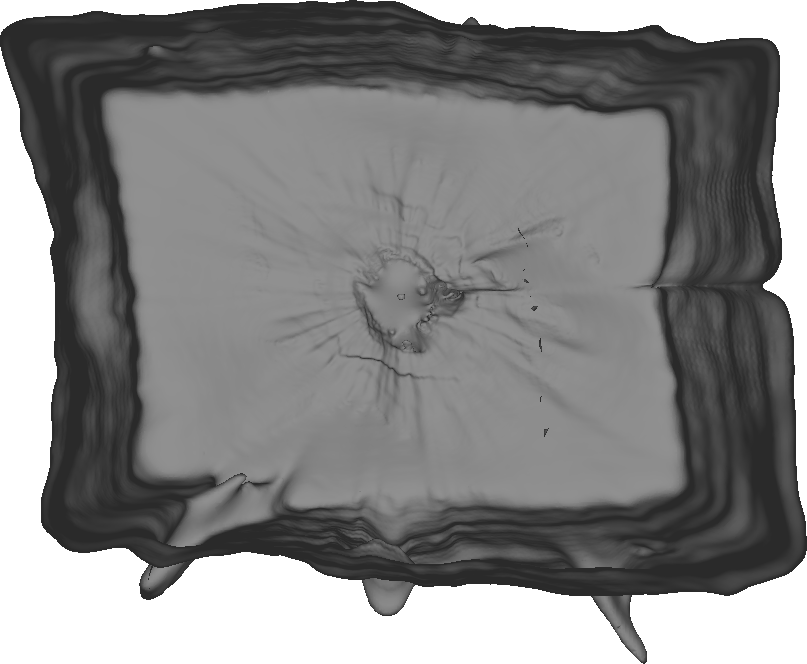}\end{subfigure}
    \hfill

    \begin{subfigure}[b]{0.19\linewidth}\includegraphics[width=\hsize]{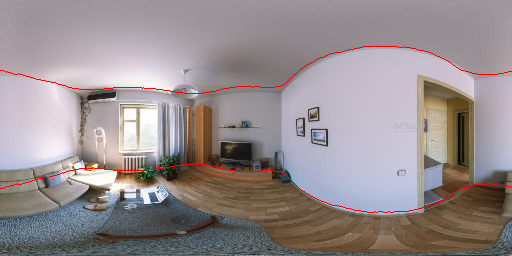}\end{subfigure}
    \hfill
    \begin{subfigure}[b]{0.19\linewidth}\includegraphics[width=\hsize]{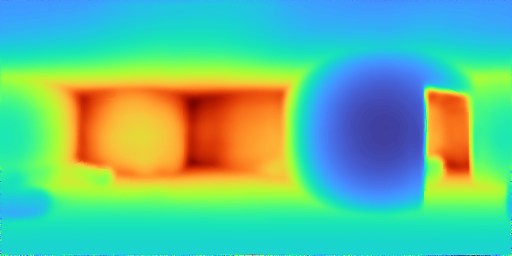}\end{subfigure}
    \hfill
    \hfill
    \begin{subfigure}[b]{0.19\linewidth}\includegraphics[width=\hsize,height=0.6\hsize]{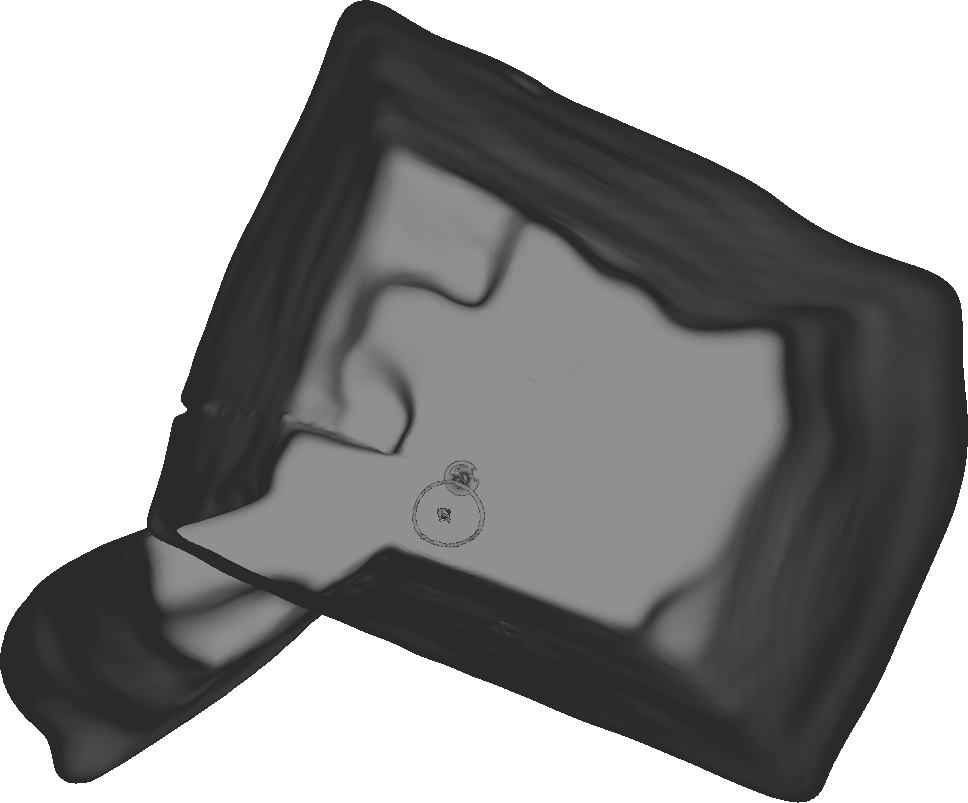}\end{subfigure}
    \hfill
    \begin{subfigure}[b]{0.19\linewidth}\includegraphics[width=\hsize]{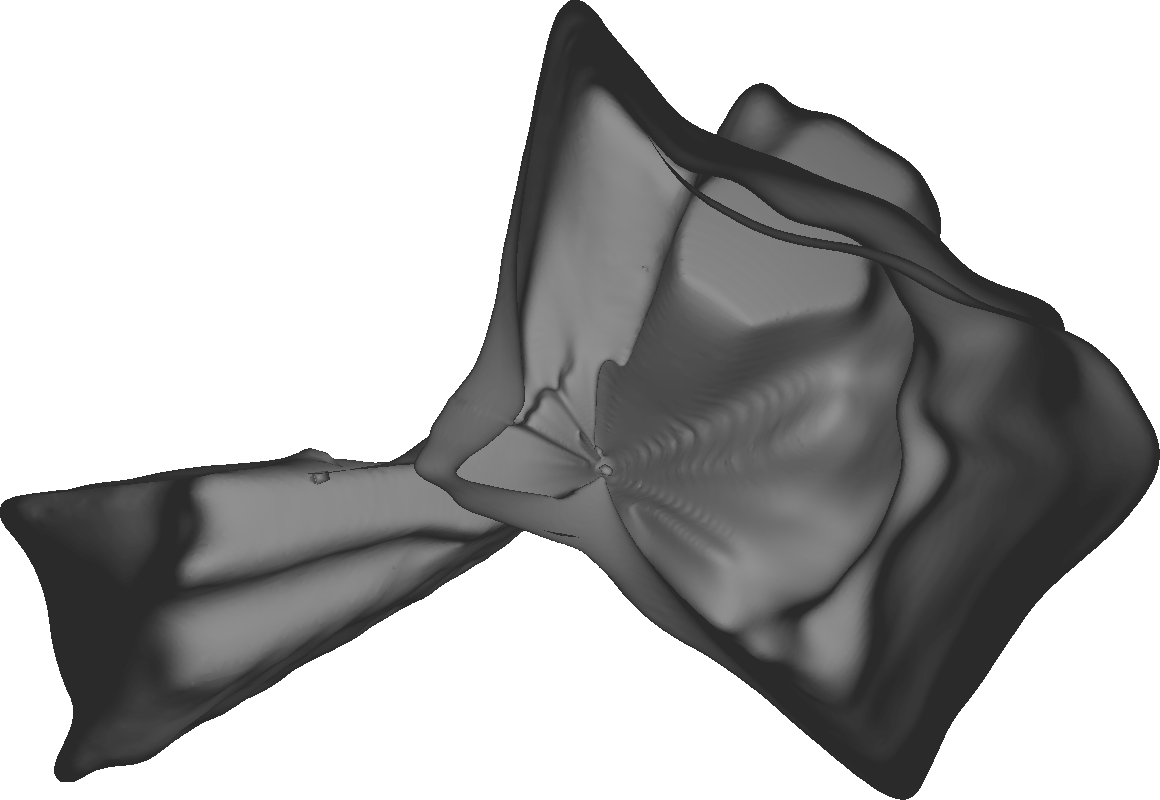}\end{subfigure}
    \hfill
    \begin{subfigure}[b]{0.19\linewidth}\includegraphics[width=\hsize]{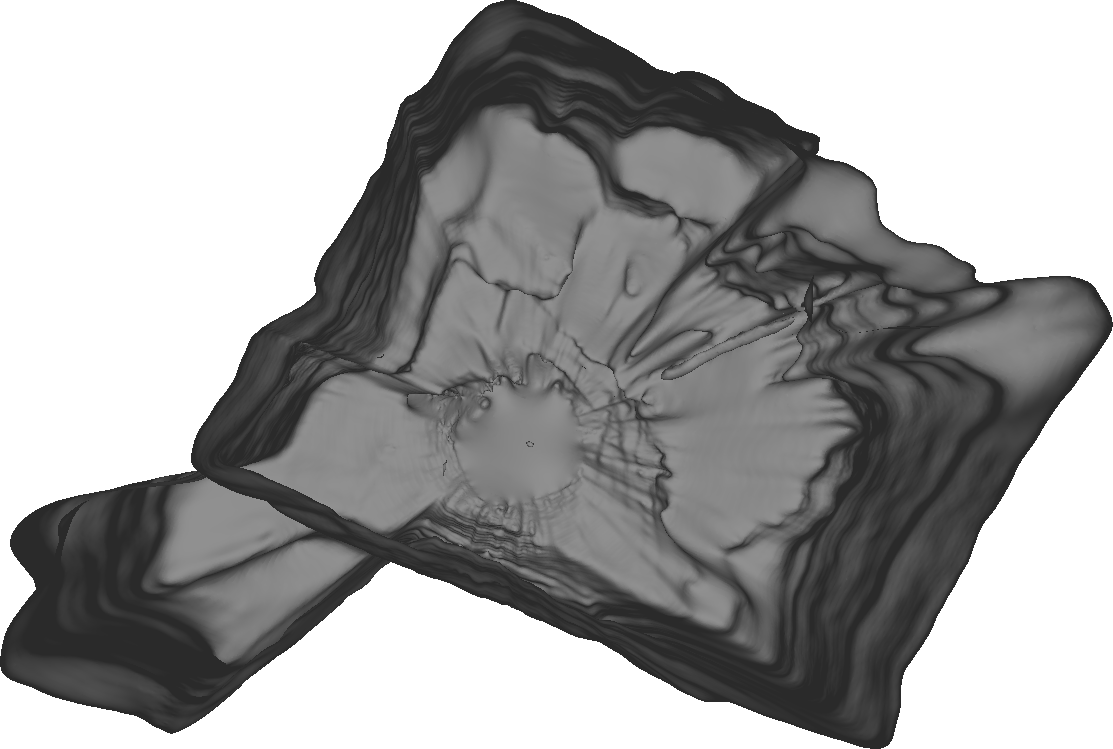}\end{subfigure}
    \hfill

    \begin{subfigure}[b]{0.19\linewidth}\includegraphics[width=\hsize]{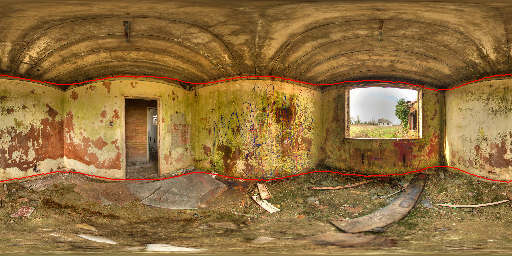}\end{subfigure}
    \hfill
    \begin{subfigure}[b]{0.19\linewidth}\includegraphics[width=\hsize]{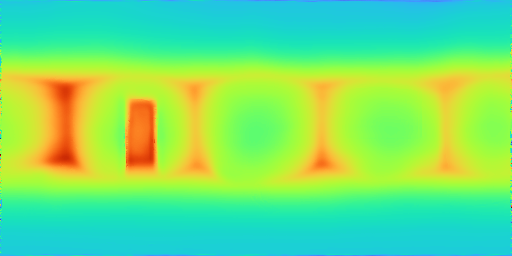}\end{subfigure}
    \hfill
    \hfill
    \begin{subfigure}[b]{0.19\linewidth}\includegraphics[width=\hsize,height=0.6\hsize]{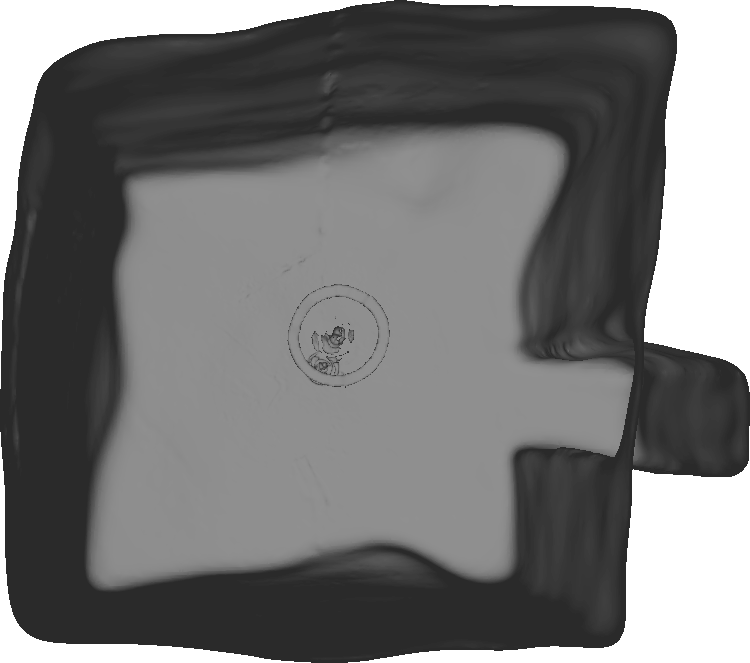}\end{subfigure}
    \hfill
    \begin{subfigure}[b]{0.19\linewidth}\includegraphics[width=\hsize]{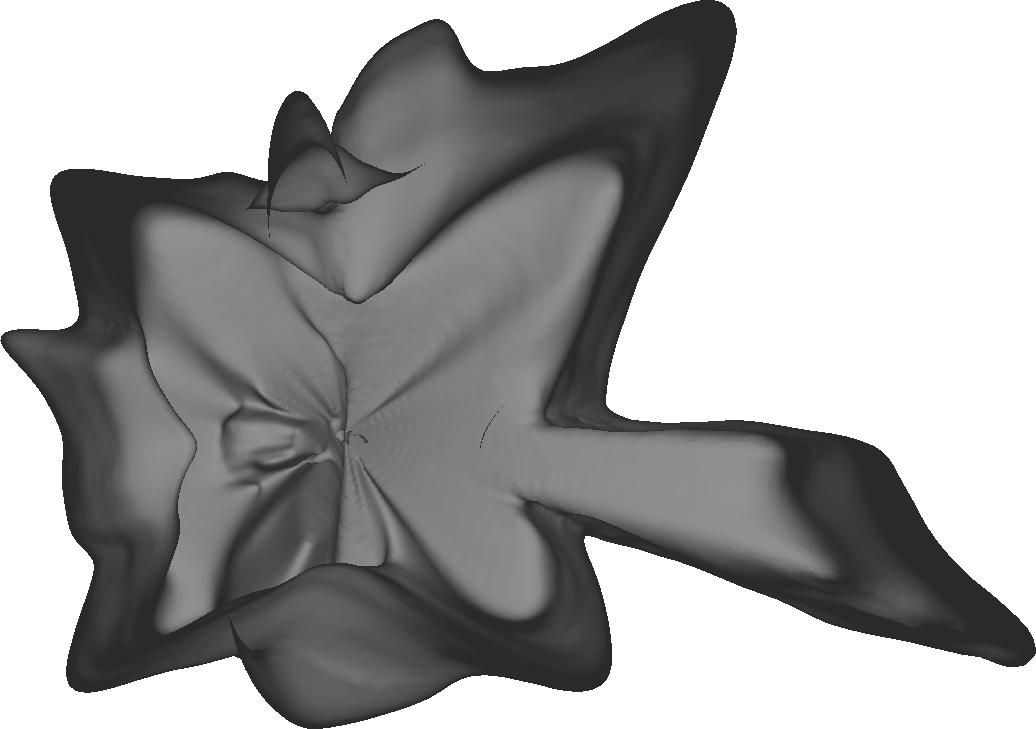}\end{subfigure}
    \hfill
    \begin{subfigure}[b]{0.19\linewidth}\includegraphics[width=\hsize]{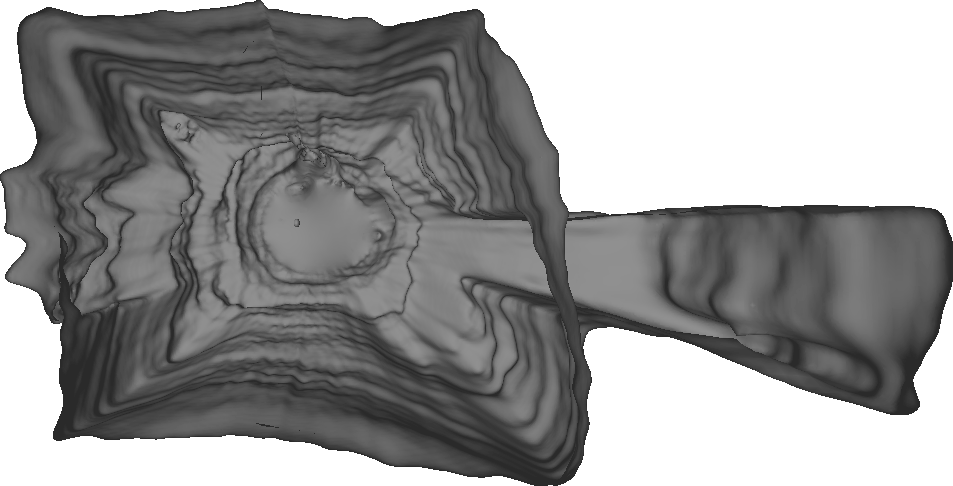}\end{subfigure}
    \hfill

    \begin{subfigure}[b]{0.19\linewidth}\includegraphics[width=\hsize]{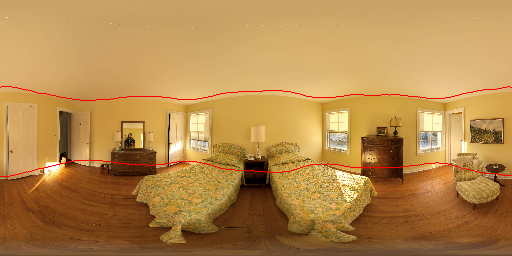}\end{subfigure}
    \hfill
    \begin{subfigure}[b]{0.19\linewidth}\includegraphics[width=\hsize]{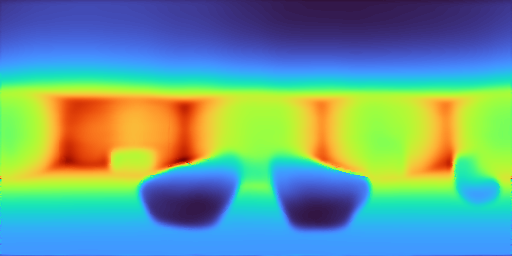}\end{subfigure}
    \hfill
    \hfill
    \begin{subfigure}[b]{0.19\linewidth}\includegraphics[width=\hsize,height=0.6\hsize]{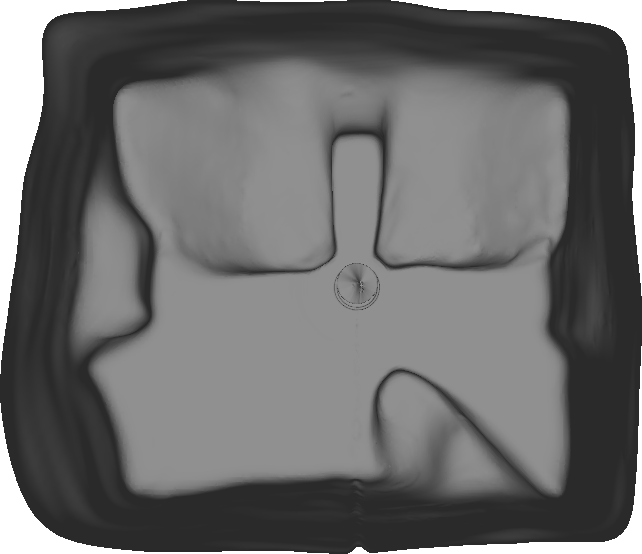}\end{subfigure}
    \hfill
    \begin{subfigure}[b]{0.19\linewidth}\includegraphics[width=\hsize,height=0.6\hsize]{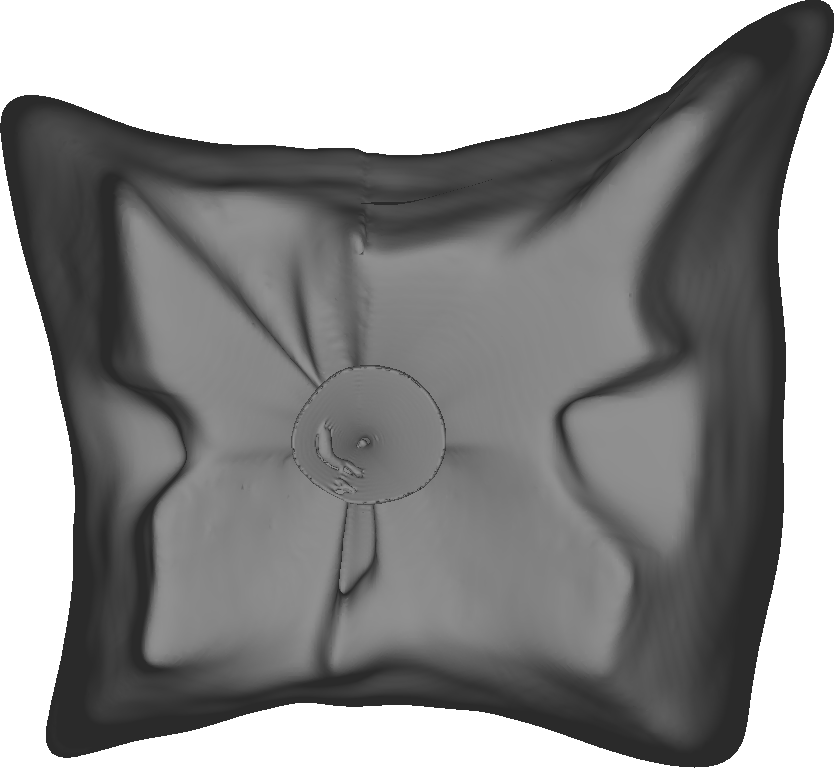}\end{subfigure}
    \hfill
    \begin{subfigure}[b]{0.19\linewidth}\includegraphics[width=\hsize,height=0.6\hsize]{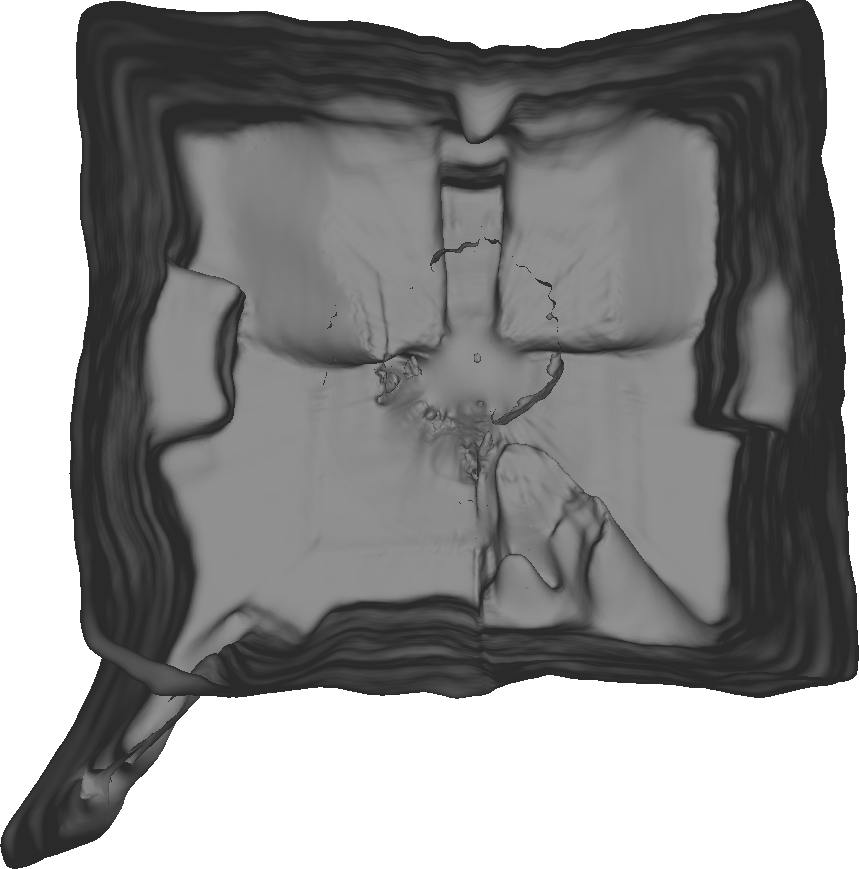}\end{subfigure}
    \hfill

    \begin{subfigure}[b]{0.19\linewidth}\includegraphics[width=\hsize]{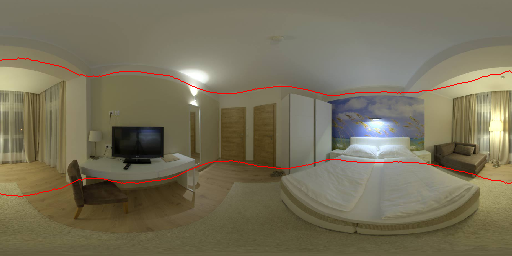}\end{subfigure}
    \hfill
    \begin{subfigure}[b]{0.19\linewidth}\includegraphics[width=\hsize]{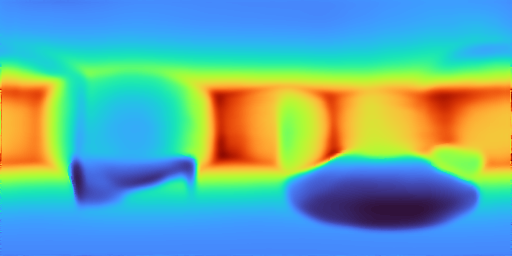}\end{subfigure}
    \hfill
    \hfill
    \begin{subfigure}[b]{0.19\linewidth}\includegraphics[width=\hsize,height=0.6\hsize]{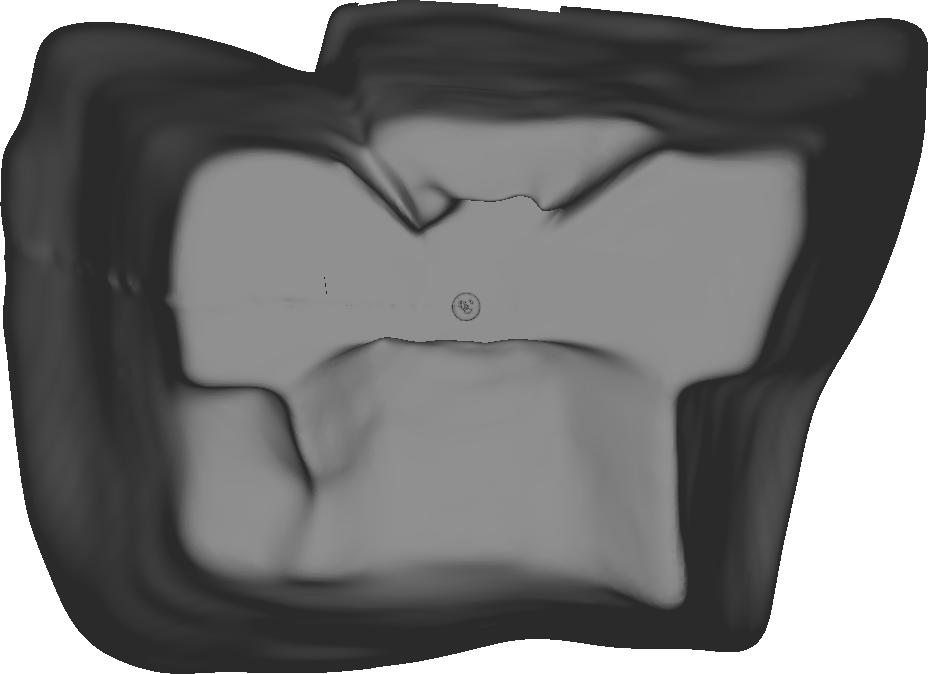}\end{subfigure}
    \hfill
    \begin{subfigure}[b]{0.19\linewidth}\includegraphics[width=\hsize,height=0.6\hsize]{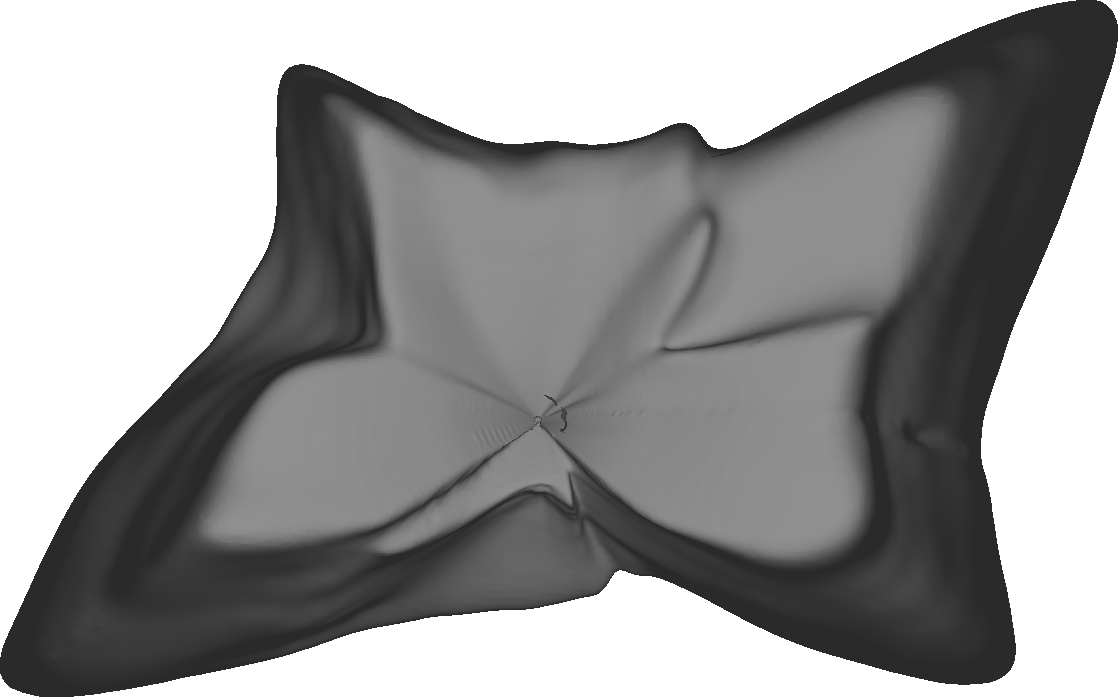}\end{subfigure}
    \hfill
    \begin{subfigure}[b]{0.19\linewidth}\includegraphics[width=\hsize,height=0.6\hsize]{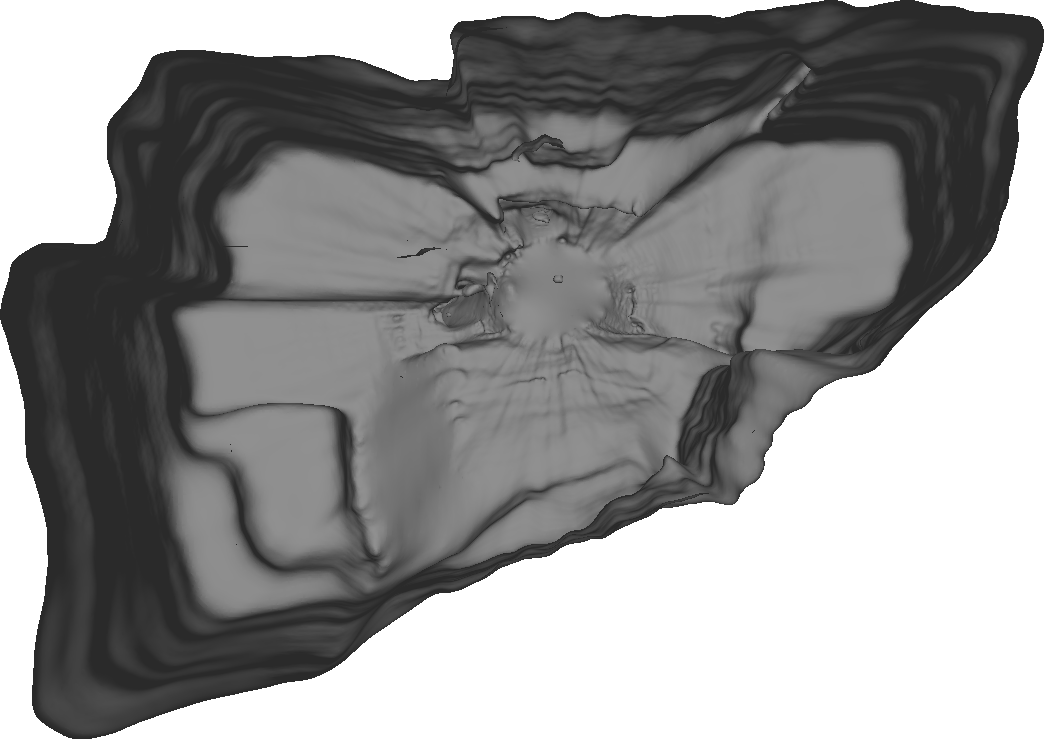}\end{subfigure}
    \hfill

    \begin{subfigure}[b]{0.19\linewidth}\includegraphics[width=\hsize]{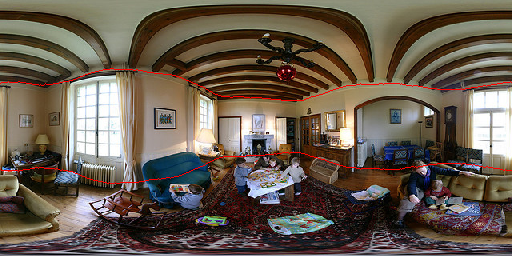}\end{subfigure}
    \hfill
    \begin{subfigure}[b]{0.19\linewidth}\includegraphics[width=\hsize]{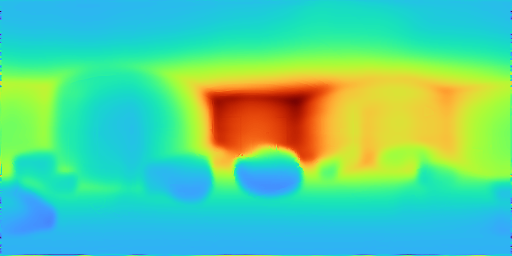}\end{subfigure}
    \hfill
    \hfill
    \begin{subfigure}[b]{0.19\linewidth}\includegraphics[width=\hsize]{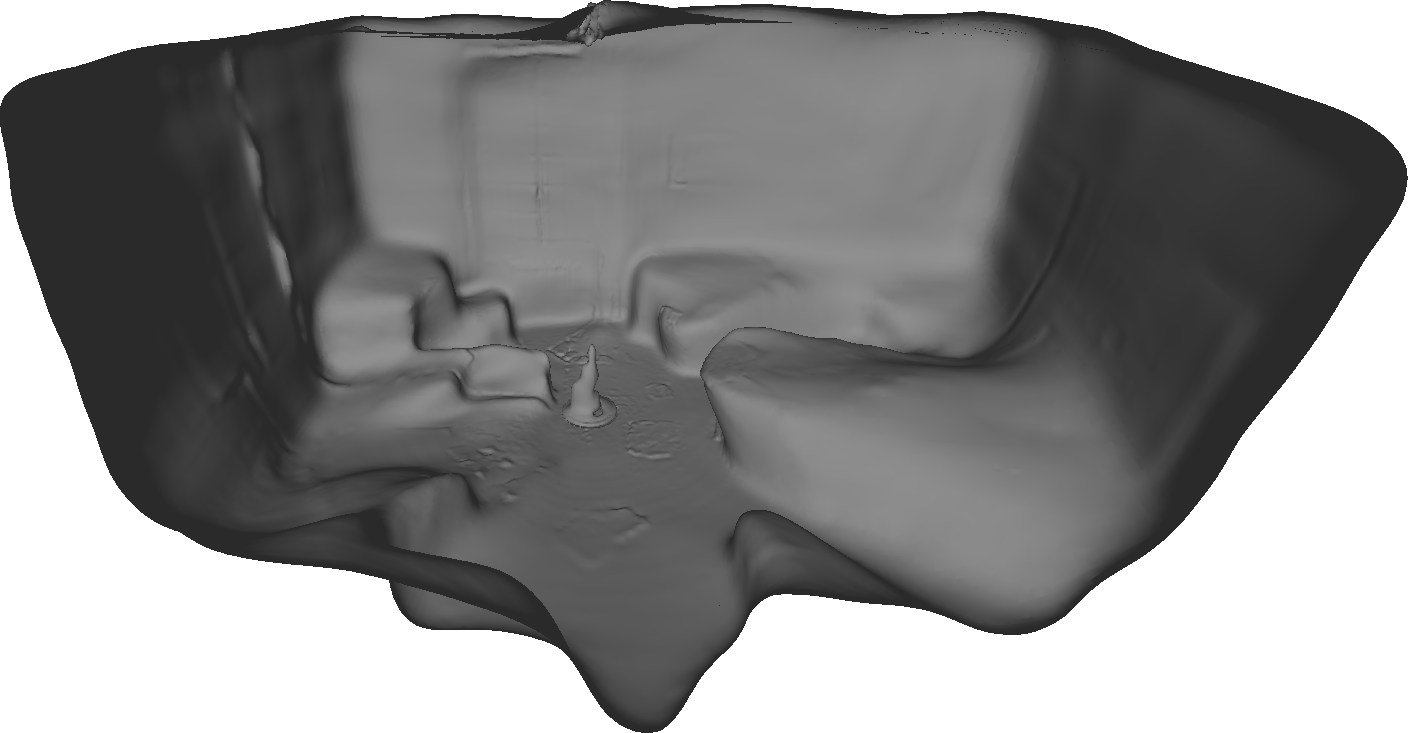}\end{subfigure}
    \hfill
    \begin{subfigure}[b]{0.19\linewidth}\includegraphics[width=\hsize,height=0.6\hsize]{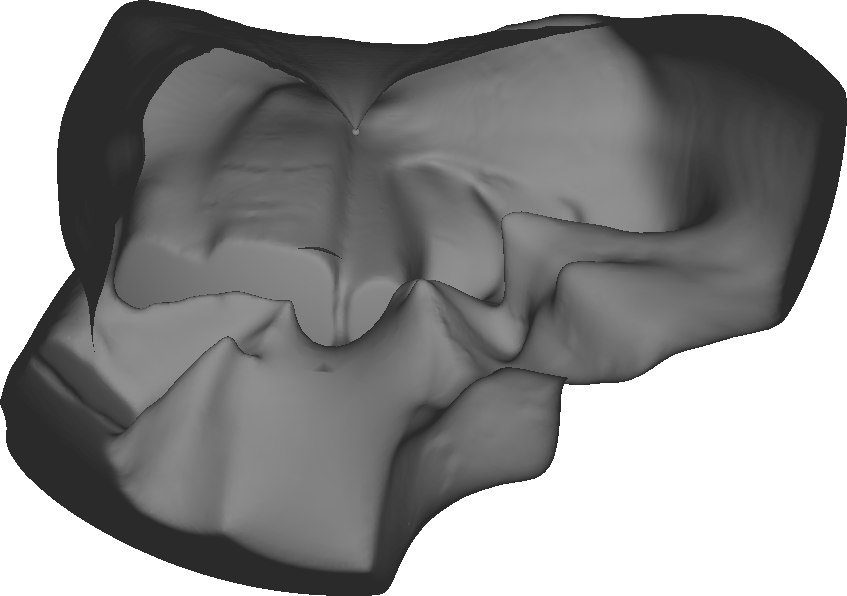}\end{subfigure}
    \hfill
    \begin{subfigure}[b]{0.19\linewidth}\includegraphics[width=\hsize]{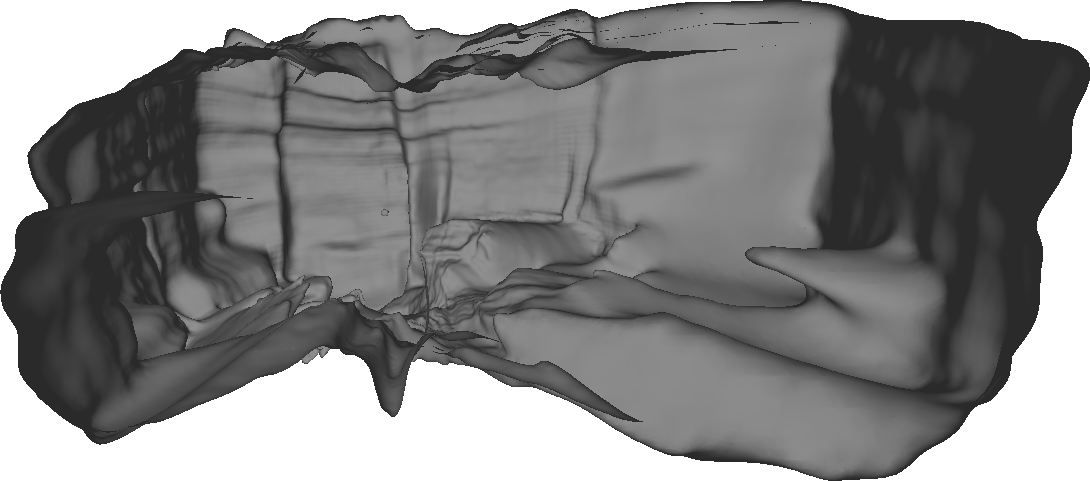}\end{subfigure}
    \hfill
    
    \caption{
        Qualitative results on the unseen, in-the-wild data. Top 3 rows are a mobile stitched panorama of a room, followed by two panoramas of an interview room acquired by a commercial \360 camera. The bottom 6 rows are samples from the SUN360 dataset. The columns present the input color image with the weak layout cue predictions of our ELD model, the corresponding depth map prediction, and the resulting mesh, followed by the mesh obtained from our ELD depth only model trained on the stitched Matterport3D data, and the publicly available BiFuse \cite{wang2020bifuse} pretrained model, again trained on the stitched Matterport3D data at double the resolution ($1024 \times 512$).
    }
    \label{fig:in-the-wild}
\end{figure*}

Our dataset is generated via synthesis from scanned 3D models, and while it is not purely synthetic as its measurements are acquired via capturing real-world scenes, its effectiveness remains to be proven.
To that end, we train our depth only model and BiFuse \cite{wang2020bifuse} using our 360V dataset and the traditional Matterport3D panorama dataset, which is created by stitching the perspective color and depth views of the Matterport camera.
All trains are conducted on the same resolution, and when testing the models we mask out the invalid areas of the stitched dataset in ours as well.
The results are presented in Table~\ref{tab:m3d_stitched} and we observe that there are no significant differences in performance, albeit the stitched dataset presents with worse metrics.
This indicates that performance on the train set does not transfer well on the test set when using the stitched panoramas, even though the same scenes are used.
It should be noted that the color camera domain is slightly different, as our scenes are the result of reconstructed data (\textit{i.e.}~processed when constructing the texture maps), while for the original raw camera data, only the stereo depth estimations are the result of a computational process that introduces noise.
To further investigate, we present a set of qualitative results for in-the-wild panoramas in Figure~\ref{fig:in-the-wild} for our ELD model trained using 360V, the depth only model trained on the stitched Matterport3D panoramas, and the publicly available\footnote{\href{https://github.com/Yeh-yu-hsuan/BiFuse}{https://github.com/Yeh-yu-hsuan/BiFuse}} pretrained BiFuse model trained on a higher resolution Matterport3D stitched panorama dataset.
Interestingly, neither model trained on the stitched data offers the robustness to in-the-wild data that our scanned domain dataset offers.
While the BiFuse meshes seemingly capture details, which is reasonable given their higher resolution inputs, in some cases the results are of low quality.
What is more interesting, is that our model integrating layout information during training, produces higher quality scene structures in all cases, especially compared to our depth only model.
The results also demonstrate our dataset's capacity to generalize to real-world scenes, and also indicate that the stereo artifacts presented at the bottom of Figure~\ref{fig:semantic} (\textit{e.g.}~mirrors, counterfactual depths in relation to the color inputs) hurt learning performance.

\section{Conclusion}
\label{sec:discussion}
This work has introduced a holistic dataset for geometric scene understanding using \360 panoramas.
It can be used for stereo-vision tasks \cite{laga2020survey}, multi-task learning \cite{9336293}, or pure geometric or semantic labelling tasks.
Apart from the high quality dense pixel-level annotations, it also provides weakly annotated layout cues which were automatically annotated using the semantic label, normal and depth maps.
To overcome the challenges associated with them, the bottom layout was reconstructed in a geometrically derived formulation following the Manhattan assumption.
Under this formulation, our work shows that the two complementary tasks of layout and depth estimation can be explicitly coupled, offering increased depth estimation performance compared to implicit or nonexistent coupling.
In addition, we also derive a layout-based attention scheme and design a depth estimation model around this dual task concept.
Our experiments demonstrate increased depth estimation performance, even when using weakly annotated layout data.

We believe that our work can open up new research directions for joint layout and depth, with larger scale datasets without relying on manual annotation, or exploiting simpler and quicker to collect annotations (\textit{e.g.}~scribbles).
This has the potential to transform traditional monocular \360 cameras into indoor 3D scanners, with works that focus on stitching disjoint scans like \cite{pintore2019automatic,pintore2018recovering} enabling the building-scale 3D reconstruction and modelling of interior scenes. 
Closing, one limitation of our approach is that the guidance of the layout cues can sometimes mislead the model and infer walls instead of cavities, as the annotations only greedily extract the first structural semantic edge, while more could follow when considering extruding or interior scene structures.

\section*{Acknowledgements}
This work was supported by the European Commission H2020 funded project ATLANTIS (\href{http://atlantis-ar.eu/}{http://atlantis-ar.eu/}) [GA 951900]

{\small
\bibliographystyle{ieee_fullname}
\bibliography{egbib}
}

\end{document}